\documentclass[10pt,twocolumn,letterpaper]{article}

\usepackage[pagenumbers]{cvpr} 

\definecolor{cvprblue}{rgb}{0.21,0.49,0.74}
\usepackage[pagebackref,breaklinks,colorlinks,allcolors=cvprblue]{hyperref}

\usepackage{multirow}
\usepackage{array}
\usepackage{booktabs} 
\usepackage{caption}
\definecolor{myred}{rgb}{0.996,0.578,0.574}
\definecolor{myyellow}{rgb}{0.988,0.961,0.898}
\definecolor{mylightred}{rgb}{0.992,0.887,0.883}
\definecolor{mylightblue}{rgb}{0.686, 0.933, 0.933}  
\usepackage[table,xcdraw]{xcolor}
\definecolor{lightgreen}{rgb}{0.56, 1.0, 0.56} 
\definecolor{lightcoral}{rgb}{1.0, 0.73, 0.73} 
\usepackage{makecell}  
\usepackage{longtable}
\usepackage{afterpage}

\def\paperID{11442} 
\def\confName{CVPR}
\def\confYear{2026}

\title{OralGPT-Omni: A Versatile Dental Multimodal Large Language Model}

\author{%
    Jing Hao\textsuperscript{1}$^\ast$$^\spadesuit$\quad
    Yuci Liang\textsuperscript{2}$^\ast$\quad
    Lizhuo Lin\textsuperscript{1}$^\ast$\quad
    Yuxuan Fan\textsuperscript{3}\quad
    Wenkai Zhou\textsuperscript{1}\quad 
    Kaixin Guo\textsuperscript{1}\quad \\
    Zanting Ye\textsuperscript{4}\quad 
    Yanpeng Sun\textsuperscript{5}\quad
    Xinyu Zhang\textsuperscript{6}\quad 
    Yanqi Yang\textsuperscript{1}\quad 
    Qiankun Li\textsuperscript{7}\quad \\
    Hao Tang\textsuperscript{8}\quad
    James Kit-Hon Tsoi\textsuperscript{1}\quad
    Linlin Shen\textsuperscript{9}$^\dagger$\quad 
    Kuo Feng Hung\textsuperscript{1}$^\dagger$\quad \\
    \textsuperscript{1}Faculty of Dentistry, The University of Hong Kong\quad \\
    \textsuperscript{2}College of Computer Science and Software Engineering, Shenzhen University\quad \\
    \textsuperscript{3}The Hong Kong University of Science and Technology (GZ)\quad \\
    \textsuperscript{4}School of Biomedical Engineering, Southern Medical University\quad \\
    \textsuperscript{5}Singapore University of Technology and Design\quad 
    \textsuperscript{6}University of Auckland\quad \\
    \textsuperscript{7}University of Science and Technology of China\quad \\
    \textsuperscript{8}School of Computer Science, Peking University\quad \\
    \textsuperscript{9}College of Artificial Intelligence, Shenzhen University
}

\begin{document}
\maketitle
\addtocontents{toc}{\protect\setcounter{tocdepth}{0}} 

\begingroup
\renewcommand\thefootnote{}
\footnotetext{$^\ast$ Equal Contribution.}
\footnotetext{$^\spadesuit$ Project Leader.}
\footnotetext{$^\dagger$ Corresponding Authors: hungkfg@hku.hk, llshen@szu.edu.cn}
\addtocounter{footnote}{-2} 
\endgroup

\begin{abstract}

Multimodal Large Language Models (MLLMs) have exhibited immense potential across numerous medical specialties; yet,
 dentistry remains underexplored, in part due to limited domain-specific data, scarce dental expert annotations, insufficient modality-specific modeling, and challenges in reliability.
In this paper, we present OralGPT-Omni, the first dental-specialized MLLM designed for comprehensive and trustworthy analysis across diverse dental imaging modalities and clinical tasks.
To explicitly capture dentists’ diagnostic reasoning, we construct TRACE-CoT, a clinically grounded chain-of-thought dataset that mirrors dental radiologists’ decision-making processes.
This reasoning supervision, combined with our proposed four-stage training paradigm, substantially strengthens the model’s capacity for dental image understanding and analysis. 
In parallel, we introduce MMOral-Uni, the first unified multimodal benchmark for dental image analysis. It comprises 2,809 open-ended question–answer pairs spanning five modalities and five tasks, offering a comprehensive evaluation suite to date for MLLMs in digital dentistry. 
OralGPT-Omni achieves an overall score of 51.84 on the MMOral-Uni benchmark and 45.31 on the MMOral-OPG benchmark, dramatically outperforming the scores of GPT-5. Our work promotes intelligent dentistry and paves the way for future advances in dental image analysis. All code, benchmark, and models will be made publicly available.

\end{abstract}    
\section{Introduction}
\label{sec:intro}



\begin{figure*}[!ht]
  \centering
  \includegraphics[width=\textwidth]{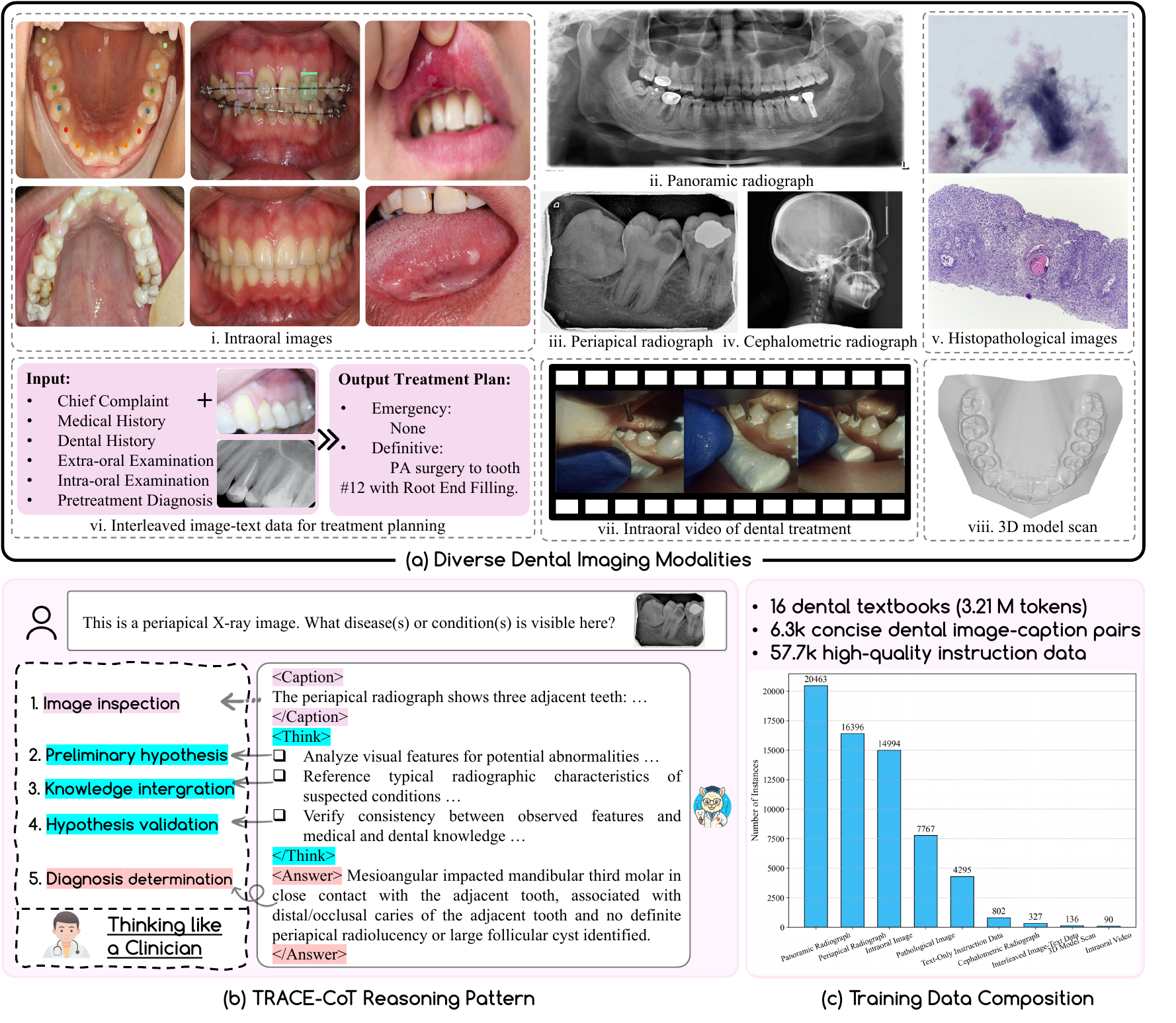}
  \vspace{-0.8cm}
  \caption{Overview of diverse dental-specialized corpus. (a) Eight types of widely used dental imaging modalities. (b) Introduction of our proposed TRACE-CoT reasoning pattern that enhances the reliability of MLLM's response. (c) The composition of the training corpus for OralGPT-Omni. The bar chart shows the distribution of various dental modalities. }
  \vspace{-0.4cm}
  \label{fig:fig_1_overview}
\end{figure*}
Multimodal Large Language Models (MLLMs) have exhibited remarkable capabilities in open-world visual understanding and reasoning in natural domains~\cite{Qwen-VL,abdin2024phi,bai2025qwen25vl,li2024llavaonevision,Mistral,xiaomi2025mimo,llava16,hao2024fullanno,sun2024descriptive}, and they also demonstrate immense potential across medical specialties, including dermatology~\cite{SkinGPT}, ophthalmology~\cite{li2025eyecaregpt}, chest radiology~\cite{khan2025chestgpt}, pathology~\cite{wsillava}, and pediatrics~\cite{yang2024pediatricsgpt}. 
However, recent studies conclude that existing MLLMs still face notable limitations in dentistry, including insufficient consistency, completeness, and clarity of generated outputs, as well as the observation of hallucinated responses, which prevents their application in real clinical applications~\cite{liu2025performance1,liu2025performance2}. Due to the lack of deep modeling of dentistry-specific expertise and modality-specific features, existing general-purpose and medical-purpose MLLMs struggle to provide reliable support for complex dental imaging analysis and highly specialized clinical requirements. Although some preliminary studies~\cite{meng2025dentvlm,zhang2025dental} have explored the
application of MLLMs to dental imaging,
their performance remains far from satisfactory.
These limitations largely stem from the substantial heterogeneity across dental imaging modalities, the intrinsic complexity of clinical diagnostic workflows, and the lack of transparency and reliability in model responses.
Progress in this domain is further constrained by the scarcity and inconsistent quality of dental imaging datasets, which result from strict privacy concerns, limited data sharing, and the high cost of expert annotation~\cite{haoreview,hao2024tmamba,hao2024semi,hao2024semit}.
At the same time, explainable decision-making is indispensable in the medical field, as clinicians and patients must understand not only the final diagnostic conclusion but also the reasoning process that leads to it. Yet, this critical aspect has been largely overlooked in existing research.

\begin{figure*}[!h]
  \centering
  \includegraphics[width=\textwidth]{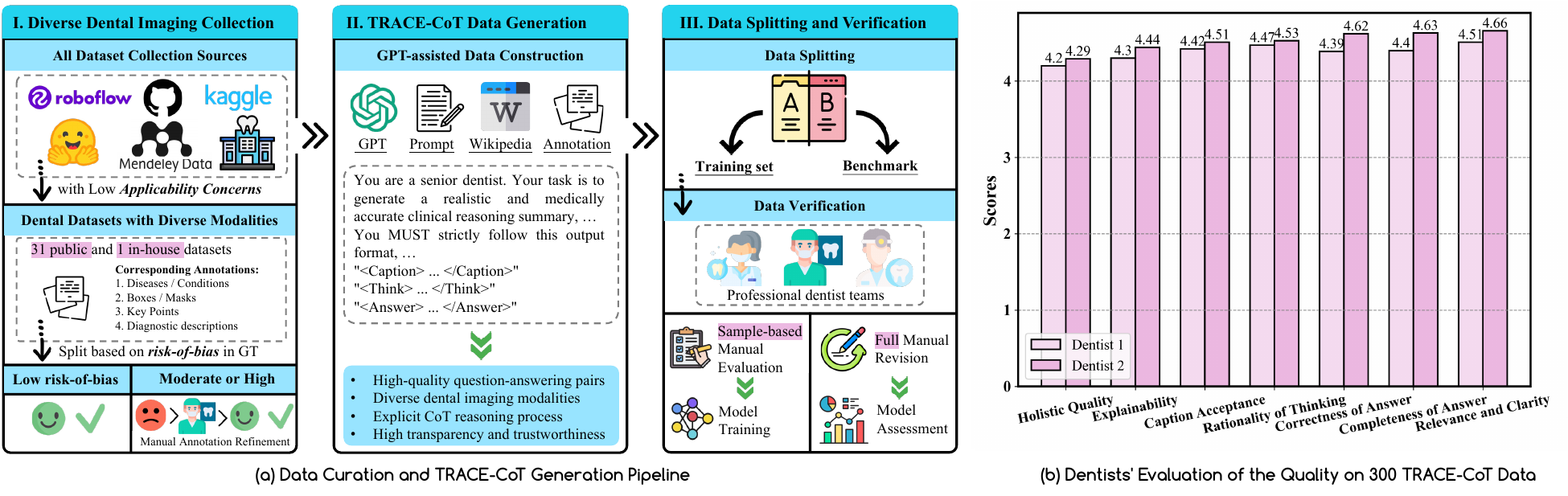}
  \vspace{-0.5cm}
  \caption{(a) The dental imaging data curation and TRACE-CoT data generation pipeline. It involves curating diverse imaging modalities from public datasets and dental hospitals. TRACE-CoT data is then generated using GPT, Wikipedia, and various annotations. Finally, the data is split into a training set and a benchmark, with professional dentists assessing the training samples and a thorough manual correction conducted on the benchmark. (b) Results from two dentists evaluating the quality of 300 TRACE-CoT data from the training set.
  }
  \vspace{-0.2cm}
  \label{fig:fig_2_data_pipeline}
\end{figure*}

To bridge these gaps in digital and intelligent dentistry, we aim to develop a dental-specialized MLLM capable of robust and comprehensive multimodal imaging analysis.
To this end, we introduce OralGPT-Omni, a versatile MLLM that facilitates comprehensive analysis across eight dental imaging modalities and five dental-related tasks, as illustrated in Figure~\ref{fig:fig_1_overview}(a).
Importantly, for the diagnosis of abnormalities, OralGPT-Omni does more than produce final answers. It reveals its diagnostic process by producing explicit chain-of-thought (CoT) rationales that mirror real clinical workflows. This capability substantially improves the model’s transparency and interpretability.
To build the OralGPT-Omni, we systematically prepare the recipe around three key aspects: diverse dental imaging curation, TRACE-CoT reasoning data construction, and a four-stage training strategy. 
First, we curate a comprehensive multimodal dental imaging dataset by aggregating data from 31 public sources and one dental hospital. Next, we design the TRACE-CoT (\textbf{T}ransparent \textbf{R}adiologic \textbf{A}nalysis with \textbf{C}linical \textbf{E}vidence), a reasoning pattern that mirrors radiologists’ diagnostic decision-making, which is demonstrated in Figure~\ref{fig:fig_1_overview}(b). 
Each CoT instance explicitly outlines the intermediate reasoning processes involved in the diagnosis of abnormalities, encompassing detailed descriptions of visual appearances, the rationale behind diagnostic hypotheses, the integration and verification of domain-specific knowledge, and a final evidence-informed diagnosis. 
Lastly, leveraging our meticulously curated large-scale, multimodal, dental-specific dataset, we adopt a four-stage training paradigm that
enhances OralGPT-Omni's integration of visual comprehension, explicit reasoning, and the ability to follow complex instructions.

Currently, there is only one public dental-related benchmark, MMOral-OPG, for panoramic X-ray analysis~\cite{mmoral}. The absence of benchmarks that include various dental imaging modalities hampers the systematic evaluation of MLLMs in dentistry. To fill this gap and guide future optimization, we present MMOral-Uni, the first unified benchmark dedicated to dental multimodal imaging analysis. It comprises 2,809 open-ended question–answer pairs that emulate realistic user interactions. All pairs are validated and refined by two experienced dentists to ensure clinical correctness. The benchmark spans five modalities and five tasks, enabling a comprehensive and rigorous evaluation of MLLMs in the dental domain.

We assess the performance of OralGPT-Omni on the MMOral-OPG~\cite{mmoral} and MMOral-Uni benchmarks to thoroughly evaluate its capability for dental applications. 
OralGPT-Omni achieves an impressive overall score of 51.84 on the Moral-Omni benchmark and 45.31 on the MMOral-OPG benchmark, significantly surpassing the scores of GPT-5. 
It also markedly outperforms existing medical MLLMs, emphasizing its unique contributions to the dental field. 
A clinical validity assessment conducted by a radiologist with over ten years of experience on three leading MLLMs indicates that our OralGPT-Omni demonstrates outstanding accuracy and potential clinical utility.
Overall, our contributions are summarized as follows:
\begin{itemize}
    \item We propose OralGPT-Omni, the first dental-specialized MLLM for comprehensive dental imaging analysis across diverse imaging modalities and tasks. 
    \item We present MMOral-Uni, the first unified benchmark for dental multimodal imaging analysis, covering five imaging modalities and five tasks, with 2,809 open-ended VQA pairs that together offer a comprehensive evaluation suite for existing MLLMs.
    \item Extensive experiments demonstrate that OralGPT-Omni delivers superior performance across diverse dental imaging modalities, achieving the superior overall scores on both the MMOral-OPG and Moral-Omni benchmarks, highlighting its potential in digital dentistry. 
\end{itemize}

\begin{figure*}[!ht]
  \centering
  \includegraphics[width=\textwidth]{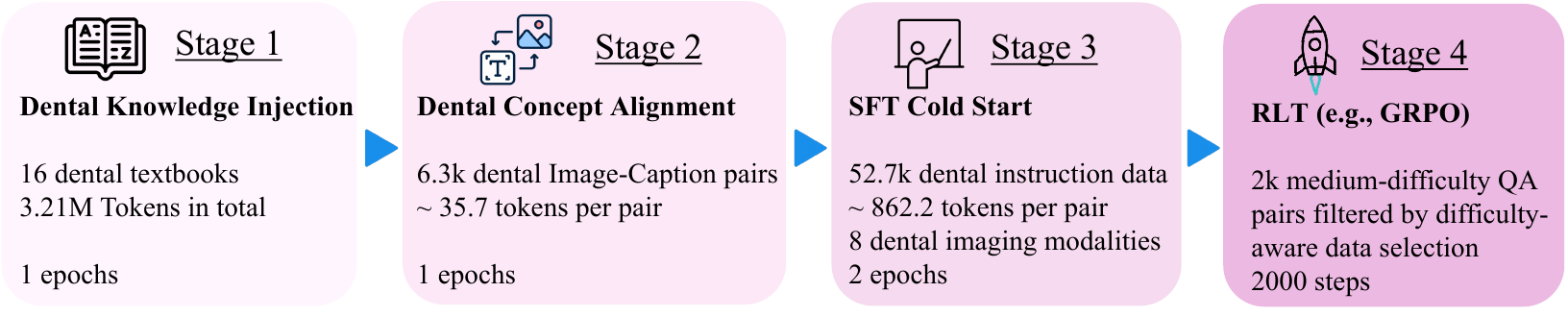}
  \vspace{-0.6cm}
  \caption{There are four stages for training OralGPT-Omni, and only the first stage is training in the single modality.}
  \label{fig:fig_3_training_stragegy}
\end{figure*}
\section{OralGPT-Omni}
\subsection{Training Corpus Construction}
We meticulously curate dental-specific multimodal datasets, including text-only corpora and imaging datasets, sourced from trusted public data platforms and dental hospitals. For the text corpora, we compile materials from undergraduate dental textbooks, dialogue data from dental community forums, and QA pairs on oral diseases and diagnoses derived from clinical guidelines. For imaging datasets, we prioritize image quality and label reliability, incorporating 31 public datasets with low applicability concerns, as defined by an authoritative systematic review~\cite{haoreview}. The dataset having ``low'' applicability concern indicates it reports both ethical approval as well as licensing requirements for its reuse. Additionally, we include one expert-annotated dataset sourced from a dental hospital in Hong Kong. A comprehensive list of all collected datasets is provided in the \textit{Appendix}. 
To further ensure the accuracy and reliability of annotations in the public datasets, we conduct manual post-processing guided by the {risk-of-bias} rating assigned in the systematic review~\cite{haoreview}. This rating quantifies the reliability of the ground truth and is widely used in diagnostic imaging research~\cite{whiting2011quadas}. For datasets with moderate or high {risk-of-bias} ratings, professional dentists manually inspect and refine the annotations, correcting any identified errors. By contrast, datasets with a low {risk-of-bias} rating do not undergo further manual validation due to high confidence in their annotation quality.

The curated training dataset comprises approximately 3.21 million text tokens, 59,658 images, and 90 videos. It supports five primary tasks: abnormality diagnosis, CVM stage prediction, recommended treatment planning, video understanding, and tooth localization and counting. The dataset encompasses eight dental imaging modalities, as illustrated in Figure~\ref{fig:fig_1_overview}(c). It includes intraoral photographs and videos, panoramic, periapical, and cephalometric radiographs, pathology images, 3D model scans, and interleaved image-text data. In addition, it provides broad demographic coverage, with samples from at least ten countries. It also offers rich structured annotations, such as abnormality labels, bounding boxes, segmentation masks, keypoints, and diagnostic descriptions. Building on this high-quality, heterogeneous resource, we convert the original annotations into chain-of-thought reasoning data, as detailed in Sec.~\ref{sec:Chain-of-Thought Data Generation}.

\subsection{TRACE-CoT Data Generation}
\label{sec:Chain-of-Thought Data Generation}
Unlike black-box predictions~\cite{aljohani2025comprehensive}, CoT explicitly reveals the intermediate reasoning steps, thereby enhancing the transparency and trustworthiness of MLLMs. However, two mainstream approaches for generating reasoning chains, CoT prompting strategies~\cite{liu2025fleming,sun2025reasonmed,lai2025medr1,ng2025x} and human-verified annotation~\cite{ding2025building}, exhibit notable limitations. The former relies heavily on the reasoning capability of the base model, while the latter is difficult to scale up due to intensive expert involvement. 

To overcome these challenges, we propose TRACE-CoT (\textbf{T}ransparent \textbf{R}adiologic \textbf{A}nalysis with \textbf{C}linical \textbf{E}vidence), a reasoning pattern that mirrors the diagnostic decision-making process of radiologists. This process is clinically coherent and closely aligns with routine radiologic practice, involving image inspection, hypothesis generation, reference to medical expertise, and verification to reach a final diagnosis. The reasoning chains proceed as follows:

\noindent
\textbf{(1) Image inspection:} conduct a thorough examination of the image, carefully describing salient structures, visual appearances, and notable patterns.

\noindent
\textbf{(2) Hypothesis generation:} propose plausible abnormalities based on the observed features.

\noindent
\textbf{(3) Medical expertise reference:} refer to authoritative clinical guidelines and well-established knowledge regarding the suspected abnormalities and their characteristic imaging signatures.

\noindent
\textbf{(4) Feature-based verification:} compare the observed image features against knowledge-based standards to identify and resolve inconsistencies.

\noindent
\textbf{(5) Evidence-informed conclusion:} aggregate the accumulated evidence and finalize the diagnostic findings.

We construct the TRACE-CoT data for abnormality diagnosis across three widely used clinical dental imaging modalities: intraoral images, periapical radiographs, and pathology images. The pipeline can be found in Figure~\ref{fig:fig_2_data_pipeline}(a). We first prompt GPT-5-mini to generate detailed descriptions of visual appearances and patterns, then we treat the sparse annotations for each image as initial diagnostic hypotheses. Guided by these hypotheses, we retrieve characteristic imaging patterns, expressed in natural language, from authoritative clinical guidelines and Wikipedia. Next, we instruct GPT-5-mini to organize these elements into complete five-step reasoning chains, and we ultimately generate 36,777 TRACE-CoT reasoning chains. Because abnormality taxonomies differ across datasets, we design dataset-specific prompts tailored to each taxonomy to ensure the quality of the chains; the full prompts are provided in the \textit{Appendix.}  To validate the quality of the generated TRACE-CoT data, we invited two dentists to evaluate 300 instances from seven perspectives, and the results are shown in Figure~\ref{fig:fig_2_data_pipeline}(b), demonstrating its high quality and reliability. Clinically grounded reasoning chains make MLLMs more transparent, interpretable, and reliable. They also mitigate the black-box nature of MLLMs, bolster clinician trust, and facilitate adoption in high‑stakes medical settings.

\subsection{Model \& Training Strategy}
OralGPT-Omni is built upon the Qwen2.5-VL-7B model~\cite{bai2025qwen25vl} due to its superior generalization and instruction-following capability. 
To further adapt the model’s capacity to dental scenarios, we employ a four-stage training strategy that progressively strengthens its multimodal understanding and reasoning abilities, as shown in Figure~\ref{fig:fig_3_training_stragegy}. 
In the first stage, we perform dental knowledge injection using the corpus composed of 16 professional dental textbooks. This stage aims to inject and consolidate fundamental dental knowledge into the model, during which only the language model is updated. In the second stage, to guide the initial alignment between dental concepts and visual representations, we employ 6,318 dental image–caption pairs extracted from dental textbooks to optimize only the vision–language projector. In the third stage, we conduct supervised fine-tuning (SFT) on the entire OralGPT-Omni architecture
to further enhance its instruction-following, multimodal comprehension, and explicit reasoning capabilities. The fine-tuning process uses 52,725 high-quality dental instruction pairs, comprising 31,777 CoT reasoning pairs and additional pairs without explicit reasoning patterns. The training corpus spans eight imaging modalities as well as text-only dialogue data, comprehensively covering a variety of dental imaging and diagnostic scenarios. 

\begin{figure}[!t]
  \centering
  \includegraphics[width=0.48\textwidth]{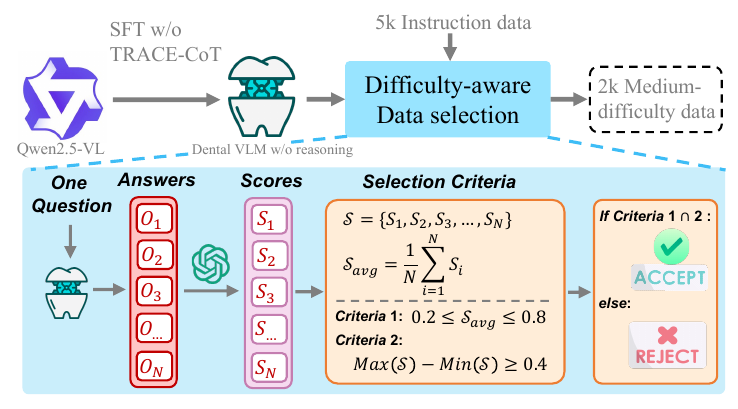}
  \vspace{-0.5cm}
  \caption{The difficulty-aware data selection strategy for RLT.}
  \label{fig:data_selection}
\end{figure}

In the final stage, we apply reinforcement learning tuning (RLT) within the Group Relative Policy Optimization (GRPO)~\cite{grpo} framework to further incentivize OralGPT-Omni’s reasoning ability. We introduce two components: a difficulty-aware data selection strategy and a TRACE-based reward closely aligned with the TRACE-CoT reasoning pattern. For data selection, we assess the difficulty of each instance relative to the preceding SFT model without the TRACE-CoT pattern and retain only medium-difficulty cases, as shown in Figure~\ref{fig:data_selection}. This is because extremely easy or extremely hard instances provide limited learning signals ~\cite{yu2025dapo,zha2025visiong1}. The rationale for selecting a model without the TRACE-CoT pattern is that we anticipate RLT can further enhance the model's performance and generalization by stimulating TRACE-CoT reasoning patterns in these medium-difficulty cases. Concretely, we perform $N$ rollouts per instance, compute the score $\mathcal{S}=\{\mathcal{S}_1,\mathcal{S}_2,...,\mathcal{S}_N\}$, and retain only those that meet two criteria: (1) $0.2\leq\mathcal{S}_{avg}\leq0.8$, (2) $Max(\mathcal{S})-Min(\mathcal{S})\geq0.4$. We set $N=5$ and select 2,000 medium-difficulty cases from 5,000 instruction data.
We also introduce a TRACE-based reward $\mathcal{R}_{\text{trace}}$ that comprehensively evaluates reasoning quality with LLM-based judge models. This reward is integrated into RLT to guide models toward producing higher-quality and more reliable reasoning paths across three aspects, including factual knowledge soundness, logical coherence, and answer consistency. Following conventional RTL reward computation, both the answer reward $\mathcal{R}_{\text{answer}}$ and the format reward $\mathcal{R}_{\text{format}}$ are considered. The $\mathcal{R}_{\text{trace}}$ and $\mathcal{R}_{\text{answer}}$ are provided by GPT-5-nano, acting as a reward-judger. The values for these three reward components range from 0 to 1. The overall reward signal unifies these components into a single formulation:
\begin{equation}
\small
\begin{split}
\mathcal{R}_{\text{total}} = \alpha \cdot \mathcal{R}_{\text{answer}} + \beta \cdot \mathbb{I}_{\mathcal{R}_{\text{answer}} > 0} \cdot \mathcal{R}_{\text{trace}} + \gamma \cdot \mathcal{R}_{\text{format}},
\end{split}
\end{equation}
where $\alpha+\beta+\gamma=1$. Note that the $\mathcal{R}_{\text{trace}}$ is ineffective when the $\mathcal{R}_{\text{answer}}$ is completely wrong in order to ensure the consistent correctness of the reasoning and answers. More details of TRACE-based reward and RLT are illustrated in the \textit{Appendix.}

\begin{figure*}[!ht]
  \centering
  \includegraphics[width=\textwidth]{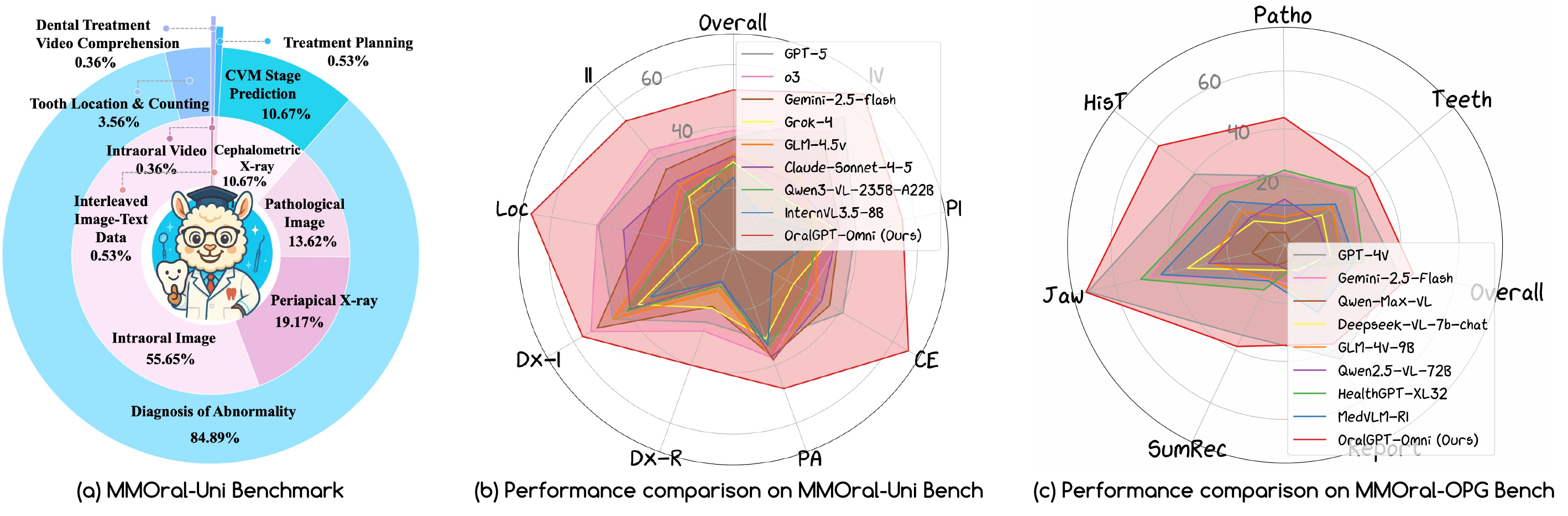}
  \caption{(a) The distribution of the MMOral-Uni benchmark, spanning five dental imaging modalities and covering five tasks. (b) Performance comparison on the MMOral-Uni benchmark. (c) Performance comparison on the MMOral-OPG benchmark. 
  }
  \label{fig:fig_4_performance}
\end{figure*}

\section{MMOral-Uni Benchmark}

\subsection{Benchmark Composition}
We introduce MMOral-Uni, the first unified benchmark for dental multimodal imaging analysis, spanning five modalities and five tasks. It includes 2,809 open-ended QA pairs to simulate realistic user interactions. The modalities cover intraoral photographs, periapical and cephalometric radiographs, pathological images, intraoral videos, and interleaved image–text inputs. The tasks include abnormality diagnosis, cervical vertebral maturation (CVM) stage prediction, treatment planning, tooth localization and counting, and dental treatment video comprehension. The combination of diverse imaging modalities and task types allows for a thorough evaluation of MLLMs in the dental field. The distribution of modalities and tasks is shown in Fig.~\ref{fig:fig_4_performance}(a). 

To ensure the quality and reliability of the benchmark, we implement strict data selection and extensive manual validation.  All images in the MMOral-uni benchmark are sourced from publicly available datasets that had been assessed as low applicability concern~\cite{haoreview}. We design the questions to align with specific task types and anticipated user intents. For example, the question is ``This is an intraoral photograph of the oral cavity. Identify any oral diseases present.'' for abnormality diagnosis, and ``What is the estimated CVM stage based on this cephalometric radiograph?'' for CVM staging. Answers are generated by converting sparse annotations into clear textual descriptions with assistance from GPT-5-mini, after which two experienced dentists validated and refined all QA pairs to further ensure correctness. More examples can be found in the \textit{Appendix}. The MMOral-Uni benchmark will be released publicly to facilitate comprehensive assessment of current MLLMs and to inform future research in multimodal AI for digital and intelligent dentistry.

\subsection{Evaluation Metrics}
Following well-established benchmarks~\cite{mmoral,yu2023mmvetv1,yu2024mmvetv2}, we meticulously design a few-shot prompt and use GPT-5-mini to conduct the open-ended evaluation. The prompt incorporates five in-context examples with free-form answers, covering fully correct, partially correct, and incorrect cases. GPT-5-mini assigns a score ranging from 0 to 1 for each sample based on the input question, ground truth, and model prediction. We report the evaluation scores for each modality as well as the overall performance. 
Comprehensive ablation experiments have validated the feasibility and stability of using LLMs as judges; the full few-shot prompt and further details are provided in the \textit{Appendix}. We have integrated the MMOral-Uni evaluation into the VLMEvalKit framework~\cite{duan2024vlmevalkit} to facilitate subsequent capability assessments of newly developed MLLMs.
\section{Experiments}

\begin{table*}[h!]
\centering
\caption{Results on the MMOral-Uni for various existing LVLMs across five dental imaging modalities. The abbreviations are defined as follows: II - Intraoral Image, PA - Periapical Radiograph, CE - Cephalometric Radiograph, PI - Pathological Image, TP - Treatment Planning, IV - Intraoral Video. The best-performing model in each category is highlighted \textbf{in-bold}, while the second-best is \underline{underlined}. }
\vspace{-5pt}
\label{tab:main}
\small 
\renewcommand{\arraystretch}{0.9}
\setlength{\tabcolsep}{4pt} 
\begin{tabular}{
    >{\raggedright\arraybackslash}p{4.3cm}  
    *{9}{>{\centering\arraybackslash}p{1.15cm}} 
}
\toprule
\multirow{2}{*}{Model} 
& \multicolumn{3}{c}{II} 
& \multirow{2}{*}{PA} 
& \multirow{2}{*}{CE} 
& \multirow{2}{*}{PI} 
& \multirow{2}{*}{TP}
& \multirow{2}{*}{IV}
& \multirow{2}{*}{Overall} \\
\cmidrule(lr){2-4}
& Loc & Dx-I & Dx-R &  &  &  &  &  &  \\
\hline
\rowcolor{mylightblue} \multicolumn{10}{l}{\textit{Proprietary LVLMs}} \\ 
\hline
GPT-5~\cite{gpt5}  & 44.60 & 45.24 & 25.16 & 31.43 & \underline{41.27} & \underline{40.52} & \textbf{80.67} & 56.00 & 36.42 \\
o3~\cite{gpto3} & \underline{45.00} & \underline{53.49} & \underline{28.19} & 37.48 & 28.60 & 37.52 & 75.33 & 50.00 & \underline{38.70} \\
Grok-4~\cite{grok4}	& 11.80 & 35.71 & 20.27 & 30.89 & 22.60 & 34.73 & 72.67 & 23.00 & 28.47 \\
Claude-Sonnet-4-5-20250929~\cite{Claude} & 36.30 & 38.85 & 11.28 & 38.03 & 33.33 & 34.52 & \underline{79.33} & 29.00 & 30.32 \\
Doubao-1.5-vision-pro-32k~\cite{doubao} & 26.90 & 43.33 & 16.97 & \underline{40.48} & 18.20 & 28.49 & 68.00 & 45.00 & 30.67 \\
Gemini-2.5-flash~\cite{team2023gemini} & 31.00 & 51.10 & 19.70 & 37.01 & 36.30 & 35.09 & 73.33 & 46.00 & 35.72 \\
GLM-4.5v~\cite{v2507glm}	& 21.70 & 45.10 & 14.37 & 38.37 & 31.00 & 27.60 & 70.00 & 33.00 & 31.05 \\
\hline
\rowcolor{mylightblue} \multicolumn{10}{l}{\textit{Open-Source LVLMs}} \\ 
\hline
Qwen2.5-VL-7B~\cite{bai2025qwen25vl} & 11.20 & 25.40 & 23.84 & 19.72 & 13.70 & 30.73 & 36.67 & 12.00 & 22.88 \\
Qwen3-VL-8B-Instruct~\cite{yang2025qwen3} & 14.30 & 27.17 & 16.95 & 33.54 & 4.97 & 30.68 & 54.00 & 18.00 & 23.45 \\
Qwen3-VL-235B-A22B~\cite{yang2025qwen3} & 18.70 & 40.00 & 12.53 & 33.62 & 26.70 & 28.75 & 56.00 & 23.00 & 27.83 \\
GLM-4.1V-9B-Thinking~\cite{v2507glm}  & 15.50 & 37.78 & 18.14 & 37.57 & 13.53 & 27.13 & 62.00 & 43.00 & 27.86 \\
InternVL3.5-8B~\cite{wang2025internvl3} & 10.10 & 30.85 & 11.02 & 32.95 & 14.80 & 28.67 & 64.67 & 14.00 & 23.39 \\
LLaVA-v1.6-Mistral-7B~\cite{llava16} & 4.80 & 20.57 & 14.03 & 11.61 & 2.50 & 12.06 & 34.67 & 18.00 & 13.54 \\
LLaVA-OneVision~\cite{li2024llavaonevision} & 7.50 & 23.13 & 17.59 & 14.34 & 0.00 & 11.49 & 40.67 & 5.00 & 15.39 \\
MiMo-VL-7B~\cite{xiaomi2025mimo} & 25.40 & 36.28 & 22.51 & 34.62 & 18.70 & 31.91 & 62.00 & 43.00 & 29.62 \\
Phi-4-multimodal-instruct~\cite{abdin2024phi} & 3.70 & 18.32 & 7.86 & 18.31 & 1.67 & 9.63 & 34.67 & 7.00 & 12.12 \\
Mistral-Small-3.1-24B~\cite{Mistral} & 16.30 & 28.14 & 18.77 & 29.15 & 11.50 & 27.02 & 62.00 & 28.00 & 23.69 \\
R-4B~\cite{yang2025r4b} & 3.80 & 32.29 & 19.16 & 28.96 & 16.60 & 27.89 & 50.67 & 19.00 & 24.94 \\
Ovis2.5-9B~\cite{lu2025ovis2} & 15.40 & 37.47 & 19.76 & 36.70 & 18.13 & 34.33 & 66.00 & \underline{61.00} & 29.60 \\
\hline
\rowcolor{mylightblue} \multicolumn{10}{l}{\textit{Medical Specific LVLMs}} \\ 
\hline
LLaVA-Med~\cite{li2023llavamed} & 5.40 & 3.99 & 5.81 & 25.31 & 1.50 & 16.08 & 28.67 & 14.00 & 10.16 \\
HuatuoGPT-Vision-7B~\cite{chen2024huatuogpt} & 11.90 & 30.19 & 23.10 & 25.79 & 19.57 & 28.02 & 41.33 & 19.00 & 25.41 \\
Lingshu-7B ~\cite{xu2025lingshu}  & 12.00 & 30.58 & 25.77 & 27.48 & 20.50 & 30.94 & 48.00  & 20.00 & 27.08 \\
MedVLM-R1~\cite{pan2025medvlm} & 10.10 & 23.88 & 14.30 & 16.25 & 0.00 & 21.28 & 24.00  & 26.00 & 16.50 \\
Med-R1-Diagnosis ~\cite{lai2025medr1} & 6.30 & 22.93 & 8.91 & 20.13 & 2.00 & 19.01 & 16.00 &  22.00 & 15.29 \\
Chiron-o1-8B ~\cite{sunchiron} & 4.60 & 20.94 & 17.37 & 34.95 & 2.03 & 31.20 & 42.00 & 23.00 & 21.61 \\
MedGemma-27B ~\cite{sellergren2025medgemma} & 11.90 & 20.44 & 17.09 & 21.65 & 22.17 & 32.69 & 68.67 & 11.00 & 21.56 \\
HealthGPT~\cite{lin2025healthgpt} & 18.00 & 20.43 & 8.84 & 19.44 & 7.50 & 31.02 & 54.67 & 14.00 & 17.32 \\

OralGPT-Omni (Ours) & \textbf{66.80} & \textbf{56.60} & \textbf{39.99} & \textbf{48.11} & \textbf{65.90} & \textbf{56.01} & 47.33 & \textbf{65.00} & \textbf{51.84} \\
\bottomrule
\end{tabular}
\end{table*}

\subsection{Experimental Setup}
\noindent
\textbf{Benchmarked MLLMs.} We systematically evaluate OralGPT-Omni on the MMOral-Uni benchmark to provide a comprehensive assessment of its performance on multimodal dental imaging analysis. We conduct evaluations of 27 representative MLLMs: 7 proprietary systems accessed via API~\cite{gpt5,gpto3,grok4,Claude,doubao,team2023gemini,glm2024chatglm}, 12 general-purpose models~\cite{Qwen-VL,yang2025qwen3,bai2025qwen25vl,v2507glm,wang2025internvl3,llava16,li2024llavaonevision,xiaomi2025mimo,abdin2024phi,Mistral,yang2025r4b,lu2025ovis2}, and 8 medical MLLMs~\cite{chen2024huatuogpt,pan2025medvlm,lai2025medr1,li2023llavamed,xu2025lingshu,sunchiron,lin2025healthgpt}.
We also evaluate OralGPT-Omni on the MMOral-OPG benchmark~\cite{mmoral} to assess its capability for panoramic X-ray analysis.

\noindent
\textbf{Implementation Details.}
We utilize the LLaMA-Factory framework~\cite{zheng2024llamafactory} to train OralGPT-Omni for the first three stages: DKI, DCA, and SFT, followed by the RLT stage using the ms-swift framework~\cite{zhao2025swift}. OralGPT-Omni is initialized with the Qwen2.5-VL-7B-Instruct pre-trained model~\cite{bai2025qwen25vl} due to its exceptional instruction-following capabilities. The training of OralGPT-Omni is conducted on 2 $\times$ NVIDIA A100 80G GPUs over approximately 90 hours. For the RLT stage, the GRPO policy generates six candidate rationales per sample, with a sampling temperature of $\tau = 0.8$. For further details on the hyperparameters employed in model training, please refer to the \textit{Appendix.}

\subsection{Comparisons on MMOral-Uni Benchmark}
The performance of various MLLMs on the MMOral-Uni benchmark is summarized in Table~\ref{tab:main} and Fig.~\ref{fig:fig_4_performance}(b). In general, our OralGPT-Omni achieves the highest overall score of 51.84, significantly surpassing GPT-5's score of 15.42, attesting to its strong capabilities in multimodal imaging analysis within the dental domain. However, it is noteworthy that OralGPT-Omni underperforms compared to proprietary MLLMs in the recommended treatment planning task. We attribute this discrepancy to differences in task objectives. 
The treatment planning task requires the formulation of subsequent strategies based on a final diagnosis. This necessitates a deeper understanding of medical expertise related to surgical procedures and postoperative recovery, which is an area where proprietary MLLMs particularly excel.
In contrast, OralGPT-Omni is primarily specialized in dental imaging analysis and abnormality diagnosis, with treatment planning data comprising only 0.006\% of the whole training dataset. This limitation contributes to its lower performance in this specific task. Nevertheless, effective treatment planning depends critically on accurate diagnostic findings, underscoring the importance of OralGPT-Omni. Furthermore, we observe that existing medical MLLMs exhibit no substantial advantage over general MLLMs in the field of dentistry, further emphasizing the unique contribution of OralGPT-Omni in multimodal dental image analysis and reasoning.

\subsection{Comparisons on MMOral-OPG Benchmark}
\begin{table}[t]
\centering
\caption{Performance on the MMOral-OPG benchmark for various LVLMs on open-ended VQA tasks. The best-performing model is highlighted \textbf{in-bold}, while the second-best is \underline{underlined}.
\vspace{-0.2cm}
}
\label{tab:MMOral-OPG}
\small
\renewcommand{\arraystretch}{0.8}
\resizebox{0.48\textwidth}{!}{%
\begin{tabular}{@{}l|ccccccc}
\toprule
\multirow{2}{*}{\textbf{Model}} 
  & \multicolumn{7}{c}{\textbf{Open‐ended VQA}} \\
\cline{2-8} 
\rule{0pt}{2.5ex} 
  & \textbf{Teeth} & \textbf{Patho} & \textbf{His} & \textbf{Jaw} & \textbf{Summ} & \textbf{Report} & \textbf{Overall} \\
\hline
\rowcolor{mylightblue} \multicolumn{8}{l}{\textit{General-purpose LVLMs}} \\ 
\hline
GPT-5~\cite{gpt5}  & \textbf{39.77} & 29.32 & \underline{44.05} & \textbf{78.56} & \textbf{40.12} & 28.20 & \underline{42.42} \\
GPT-4V~\cite{hurst2024gpt4v}  & {31.46} & 23.79 & 39.51 & 69.81 & {34.29} & \textbf{43.70} & {39.38} \\
Gemini-2.5-Flash~\cite{team2023gemini} & 28.04 & 24.77 & 31.90 & 47.81 & 12.98 & 16.70 & 27.84 \\
Qwen-Max-VL~\cite{Qwen-VL} & 2.10 & 4.47 & 7.06 & 11.62 & 7.98 & 5.50 & 5.29 \\
Deepseek-VL-7b-chat~\cite{lu2024deepseek} & 16.48 & 7.50 & 13.44 & 34.56 & 9.52 & 9.60 & 15.95  \\
GLM-4V-9B~\cite{glm2024chatglm}  & 20.94 & 9.70 & 18.77 & 26.62 & 12.74 & 21.30 & 19.74 \\
Qwen2.5-VL-72B~\cite{bai2025qwen25vl}  & 13.90 & 15.83 & 15.40 & 27.12 & 7.38 & 11.50 & 15.38  \\ \hline
\rowcolor{mylightblue} \multicolumn{8}{l}{\textit{Medical Specific LVLMs}} \\ \hline
LLaVA-Med~\cite{li2023llavamed}  & 0.91 & 1.52 & 0.00 & 0.00 & 0.00 & 24.50 & 4.76  \\
HealthGPT-XL32~\cite{lin2025healthgpt}  & 30.64 & 25.83 & 27.98 & 51.12 & 17.02 & 8.00 & 27.80 \\
MedVLM-R1~\cite{pan2025medvlm}  & 22.42 & 13.71 & 24.42 & 43.88 & 13.57 & 25.80 & 24.70 \\
MedDr~\cite{he2024meddr}  & 22.99 & \underline{32.58} & 29.57 & 52.44 & 20.95 & 8.70 & 26.20  \\
OralGPT-Omni (Ours)  & \underline{37.26} & \textbf{43.94} & \textbf{55.34} & \underline{70.50} & \underline{38.57} & \underline{37.90} & \textbf{45.31}  \\

\bottomrule
\end{tabular}%
}
\vspace{-0.3cm}
\end{table}

We evaluate OralGPT-Omni on the MMOral-OPG benchmark~\cite{mmoral}, specifically focusing on open-ended VQA for panoramic X-ray analysis. We assess performance across six clinically grounded dimensions, as shown in Table~\ref{tab:MMOral-OPG} and Fig.~\ref{fig:fig_4_performance}(c). OralGPT-Omni outperforms existing medical MLLMs, achieving an overall score of 45.31. Its scores on the the ``Teeth'', ``Patho'', ``His'', ``Jaw'', and ``Summ'' dimensions notably exceed those of GPT-4V, underscoring its strength in panoramic X-ray interpretation. By contrast, its report generation performance lags behind GPT-4V. We hypothesize that this gap arises from the intrinsic complexity of panoramic radiographs, which contain dense anatomical structures and multi-dimensional conditions that OralGPT-Omni may not fully capture when generating comprehensive reports.

\begin{figure*}[!t]
  \centering
  \includegraphics[width=\textwidth]{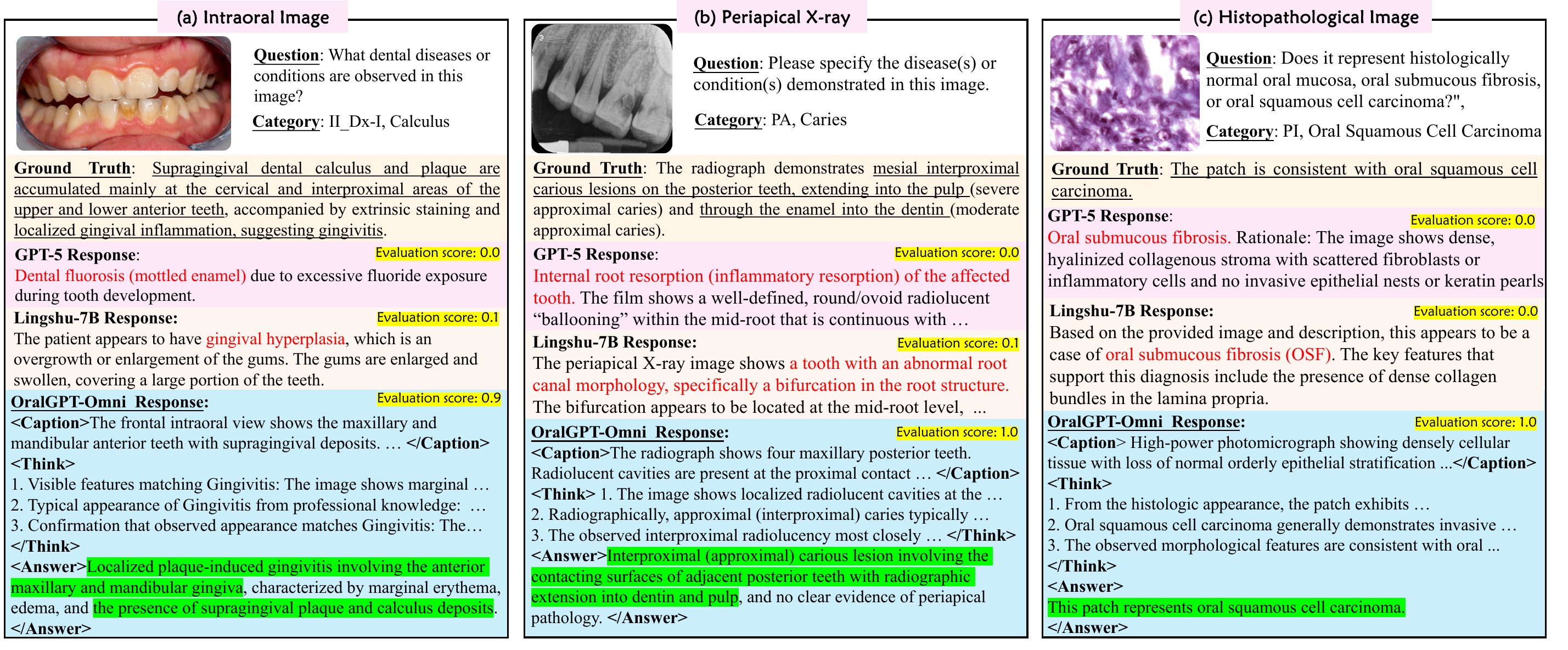}
  \vspace{-0.7cm}
  \caption{Three modalities of case studies on the MMOral-Uni benchmark are presented, with correct responses highlighted in \colorbox{green}{\textcolor{black}{green}} and obvious incorrect responses highlighted in \textcolor{red}{red}.
  }
  \vspace{-0.2cm}
  \label{fig:fig_6_case_vis}
\end{figure*}

\begin{table*}[h!]
\centering
\small 
\caption{Ablation experiments of the four-stage training strategy and the TRACE-CoT reasoning pattern.}
\vspace{-6pt}
\label{tab:ablation}
\renewcommand{\arraystretch}{0.9}
\setlength{\tabcolsep}{2.5pt} 
\begin{tabular}{
    >{\centering\arraybackslash}p{0.8cm}  
    >{\raggedright\arraybackslash}p{4.3cm}  
    *{9}{>{\centering\arraybackslash}p{1.15cm}} 
}
\toprule
\multirow{2}{*}{No.} & \multirow{2}{*}{Model} 
& \multicolumn{3}{c}{II} 
& \multirow{2}{*}{PA} 
& \multirow{2}{*}{CE} 
& \multirow{2}{*}{PI} 
& \multirow{2}{*}{TP}
& \multirow{2}{*}{IV}
& \multirow{2}{*}{Overall} \\
\cmidrule(lr){3-5}
& & Loc & Dx-I & Dx-R &  &  &  &  &  &  \\
\hline 
1 & Baseline (Qwen2.5-VL-7B)~\cite{bai2025qwen25vl} & 11.20 & 25.40 & 23.84 & 19.72 & 13.70 & 30.73 & 36.67 & 12.00 & 22.88 \\
\hline
2 & +1st stage (DKI) & 12.50 & 24.43 & 22.96 & 21.32 & 14.37 & 36.63 & 38.67 & 18.00 & 23.66 \\
3 & +2nd stage (DCA) & 11.90 & 25.44 & 23.22 & 24.38 & 17.70 & 29.19 & 50.00 & 31.00 & 24.00 \\
4 & +3rd stage (SFT) & 64.40 & 52.61 & 38.02 & 43.97 & 63.30 & 53.89 & 35.33 & 33.00 & 48.67 \\
5 & +4th stage (RLT) & 66.80 & 56.60 & 39.99 & 48.11 & 65.90 & 56.01 & 47.33 & 65.00 & 51.84 \\
\hline
6 & SFT wo/ TRACT-CoT & 60.60 & 44.26 & 33.73 & 48.00 & 58.93 & 44.67 & 36.00 & 31.00 & 44.31 \\
7 & SFT w/ TRACT-CoT & 64.40 & 52.61 & 38.02 & 43.97 & 63.30 & 53.89 & 35.33 & 33.00 & 48.67 \\
- & $\Delta$ $\uparrow$ & \cellcolor{lightgreen}{+3.80} & \cellcolor{lightgreen}{+8.35} & \cellcolor{lightgreen}{+4.29} & \cellcolor{lightcoral}{-4.03} & \cellcolor{lightgreen}{+4.37} & \cellcolor{lightgreen}{+9.22} & \cellcolor{lightcoral}{-0.67} & \cellcolor{lightgreen}{+2.00} & \cellcolor{lightgreen}{+4.36} \\
\bottomrule
\end{tabular}
\vspace{-0.4cm}
\end{table*}
\subsection{Ablation and In-Depth Study}
\noindent
\textbf{Effect of Four-Stage Training Strategy.}
We conduct ablation studies on MMOral-Uni to evaluate the effectiveness of the four-stage training strategy, with results presented in Table~\ref{tab:ablation}. Overall, this strategy progressively improves performance in dental image analysis. The first two stages, DKI and DCT, achieve a modest improvement over the baseline model (Qwen2.5-VL-7B). By incorporating a domain-specific dental knowledge corpus and concise descriptions of diverse dental imaging modalities, the overall score on MMOral-Uni increases from 22.88 to 24.00. The SFT stage, with carefully curated high-quality instruction data, yields a substantial improvement, raising the overall score from 24.00 to 48.67. SFT markedly strengthens instruction following and provides a solid foundation for explicit, transparent reasoning in dental image analysis, thereby preparing the model for the subsequent RLT stage. RLT further improves the overall score by 3.17 points and incentivizes stronger reasoning capabilities.

\noindent
\textbf{Effect of TRACE-CoT Reasoning Data.}
Our proposed TRACE-CoT pattern mirrors the diagnostic decision-making process of radiologists and is designed to explicitly improve the transparency and trustworthiness of MLLMs. To assess whether explicit reasoning chains can enhance diagnostic accuracy, we perform an ablation study during the SFT phase. We compare the model trained with TRACE-CoT data against the model trained without it, and the results are shown in Table~\ref{tab:ablation} (rows 6-7). Incorporating TRACE-CoT data during SFT increases the overall score by 4.36 points relative to training on answer-only data that excludes reasoning chains. Notably, in our instruction data, TRACE-CoT annotations are included only for the II-Dx-I, II-Dx-R, PA, and PI modalities, and the most significant improvements appear in II-Dx-I, II-Dx-R, and PI. These findings provide concrete evidence that TRACE-CoT data could strengthen diagnostic performance.

\noindent
\textbf{Case Study \& Clinical Feedback.}
We conduct an in-depth case study of the OralGPT-Omni model, focusing on its performance across various dental imaging modalities, as illustrated in Fig.~\ref{fig:fig_6_case_vis}. In comparison to GPT-5 and the medical-specific MLLM Lingshu-7B~\cite{xu2025lingshu}, our OralGPT-Omni demonstrates superior abnormality diagnosis capabilities in three modalities: intraoral images, periapical X-rays, and histopathological images. Additionally, it provided explicit TRACE-CoT reasoning patterns, which enhance its transparency and reliability. To further evaluate the clinical validity of OralGPT-Omni, we invite a radiologist with over ten years of experience to perform comprehensive clinical assessments of the responses from three top-performing models, the details of which can be found in the \textit{Appendix}.

\section{Conclusion}
In this paper, we introduce OralGPT-Omni, the first dental-specialized MLLM for comprehensive analysis across diverse dental imaging modalities and tasks. The success of our OralGPT-Omni relies on high-quality, large-scale, dental-specific multimodal datasets; the explicit TRACE-CoT reasoning pattern and data construction pipeline; and the four-stage training strategy. In addition, the proposed MMOral-Uni is the first unified benchmark dedicated to multimodal dental imaging analysis, which offers a comprehensive evaluation suite for digital dentistry.

Experiments show that OralGPT-Omni outperforms current state-of-the-art MLLMs on the MMOral-Uni and MMOral-OPG benchmarks by a large margin. It also provides explicit reasoning chains that explain how the final diagnosis is reached. This work paves the way for future advancements in digital and intelligent dentistry by enabling natural language interaction, multimodal dental imaging analysis, and enhanced explainability in next-generation dental artificial intelligence.

{
    \small
    \bibliographystyle{ieeenat_fullname}
    \bibliography{main}
}

\addtocontents{toc}{\protect\setcounter{tocdepth}{1}} 
\clearpage
\setcounter{section}{0}
\setcounter{page}{1}
\maketitlesupplementary

\tableofcontents


\section{Related Works}

\textbf{Medical Large Vision-Language Models.}
Medical MLLMs have shown great potential to serve as valuable assistants for clinicians, researchers, and trainees by providing an interactive natural language interface for analyzing medical images across diverse modalities. Recent efforts~\cite{chen2024huatuogpt,he2024meddr,li2024gmai,li2023llavamed} aim to develop medical general-purpose MLLMs that are capable of simultaneously analyzing and responding to images and instructions from multiple medical disciplines. 
Meanwhile, more researchers are dedicated to building discipline-specialized medical MLLMs, such as dermatology~\cite{SkinGPT}, ophthalmology~\cite{li2025eyecaregpt}, chest~\cite{khan2025chestgpt}, pathology~\cite{wsillava}, and pediatrics~\cite{yang2024pediatricsgpt}. These specialized models demonstrate superior imaging diagnostic performance within their respective fields compared to those medical general-purpose MLLMs. However, dentistry-specific MLLMs remain largely underexplored. OralGPT~\cite{mmoral}, trained on a large-scale instruction dataset, represents the first vision-language model specialized for panoramic X-ray analysis. Jia Z. \etal~\cite{zhang2025dental} finetuned the Qwen2.5-VL model using less than 2k intraoral images to achieve diagnosis of four types of oral mucosal diseases. DentVLM~\cite{meng2025dentvlm} supports basic oral disease diagnosis on three imaging modalities—panoramic, lateral, and intraoral images—but lacks the capability to provide detailed explanations. In contrast, our OralGPT-Omni offers comprehensive analysis across seven dental imaging-based modalities as well as text-only interactions. It also generates explicit chain-of-thoughts, significantly enhancing the model’s transparency and trustworthiness.

\noindent
\textbf{Medical Chain-of-Thought Reasoning.}
Existing medical MLLMs have made significant progress through supervised fine-tuning (SFT) on specific instruction datasets. However, these models~\cite{he2024meddr,li2023llavamed,li2024gmai,meng2025dentvlm,zhang2025dental} primarily generate final answers without revealing the underlying reasoning processes, due to the bias towards memorizing task-specific shortcuts rather than learning generalizable reasoning in the SFT stage. Prior studies, including MedVLM-R1~\cite{pan2025medvlm}, Med-R1~\cite{lai2025medr1}, and MED-RLVR~\cite{zhang2025medrlvr}, have attempted to enhance the reasoning abilities of medical MLLMs via reinforcement learning. Nevertheless, the quality of the generated reasoning chains heavily relies on the base policy model. To address this limitation, GMAI-VL-R1~\cite{su2025gmair1} designs a reasoning data synthesis approach that employs GPT-4o for step-by-step reasoning generation through rejection sampling. However, it still suffers from the hallucination sourced from GPT-4o. To alleviate hallucination issues, Q. Li \etal~\cite{li2025aor} constructed the AOR-Instruction reasoning dataset by incorporating explainable region-level visual content based on anatomical regions and ontologies. Similarly, X-Ray-CoT~\cite{ng2025x} created reasoning data for chest X-rays by integrating general medical knowledge and visual concept descriptions using the CoT prompting strategy. In contrast, we propose the TRACE-CoT reasoning pattern that closely mirrors the diagnostic decision-making process of radiologists, which is presented in Section~\ref{sec:Chain-of-Thought Data Generation}.

\section{Details on Training Data Construction}
\begin{figure*}[!ht]
  \centering
  \includegraphics[width=\textwidth]{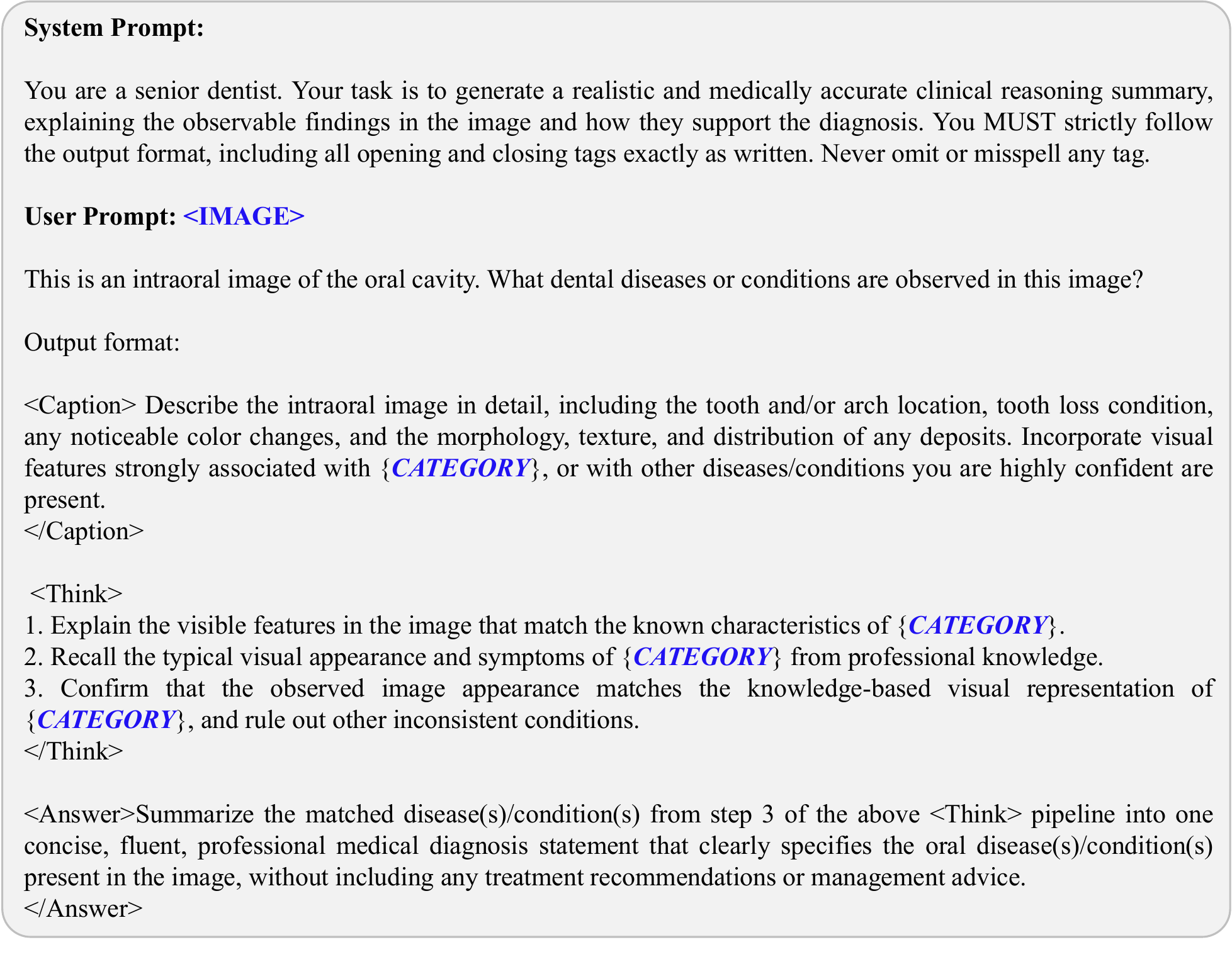}
  \caption{The prompt for GPT-5-mini to generate the TRACE-CoT reasoning chains for image-level diagnosis of intraoral images.}
  \label{fig:suppl_Prompt_II_Image-level}
\end{figure*}

\begin{figure*}[!ht]
  \centering
  \includegraphics[width=\textwidth]{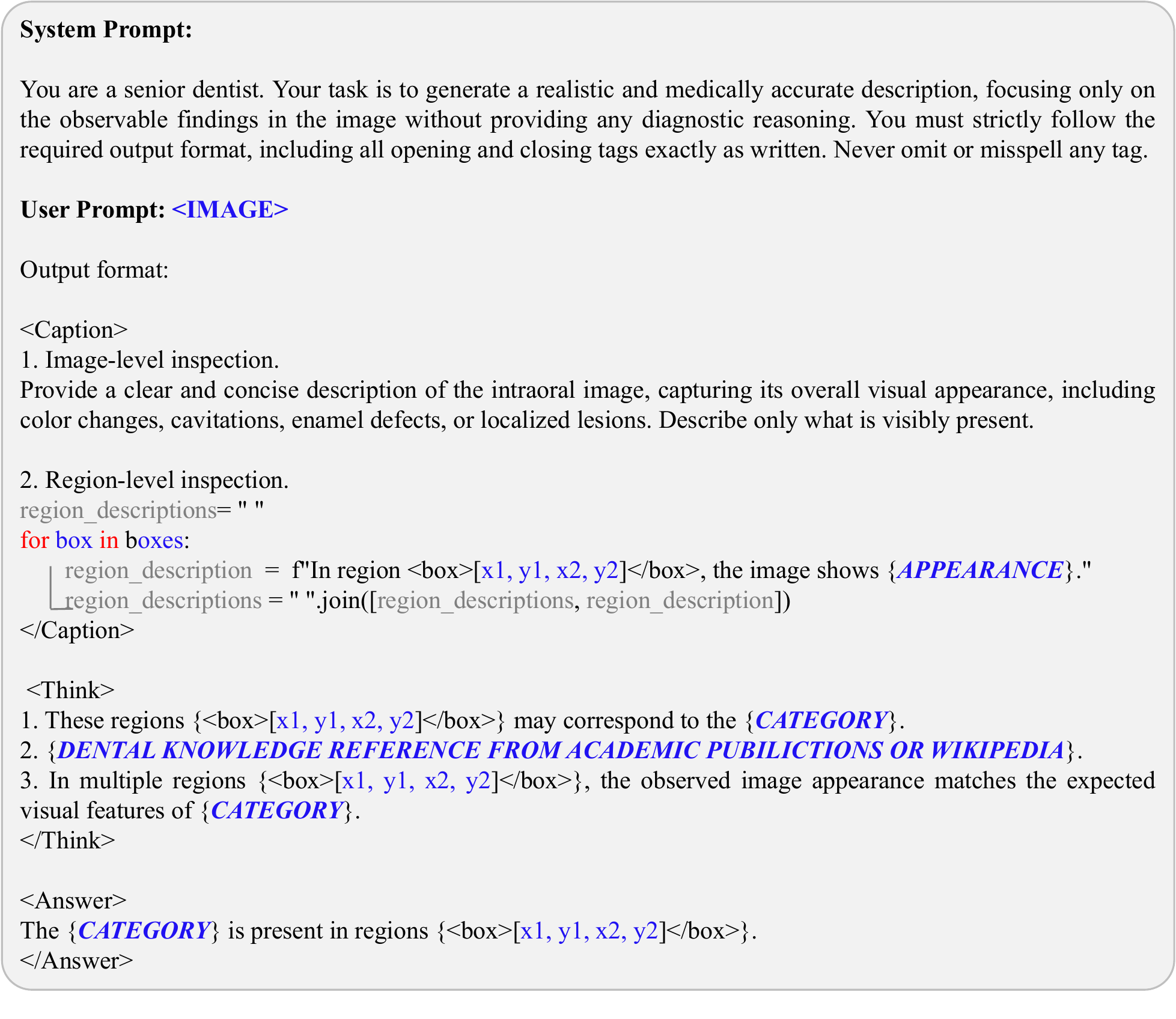}
  \caption{The prompt for GPT-5-mini to generate the TRACE-CoT reasoning chains for region-level diagnosis of intraoral images.}
  \label{fig:suppl_Prompt_II_Region-level}
\end{figure*}

\begin{figure*}[!ht]
  \centering
  \includegraphics[width=\textwidth]{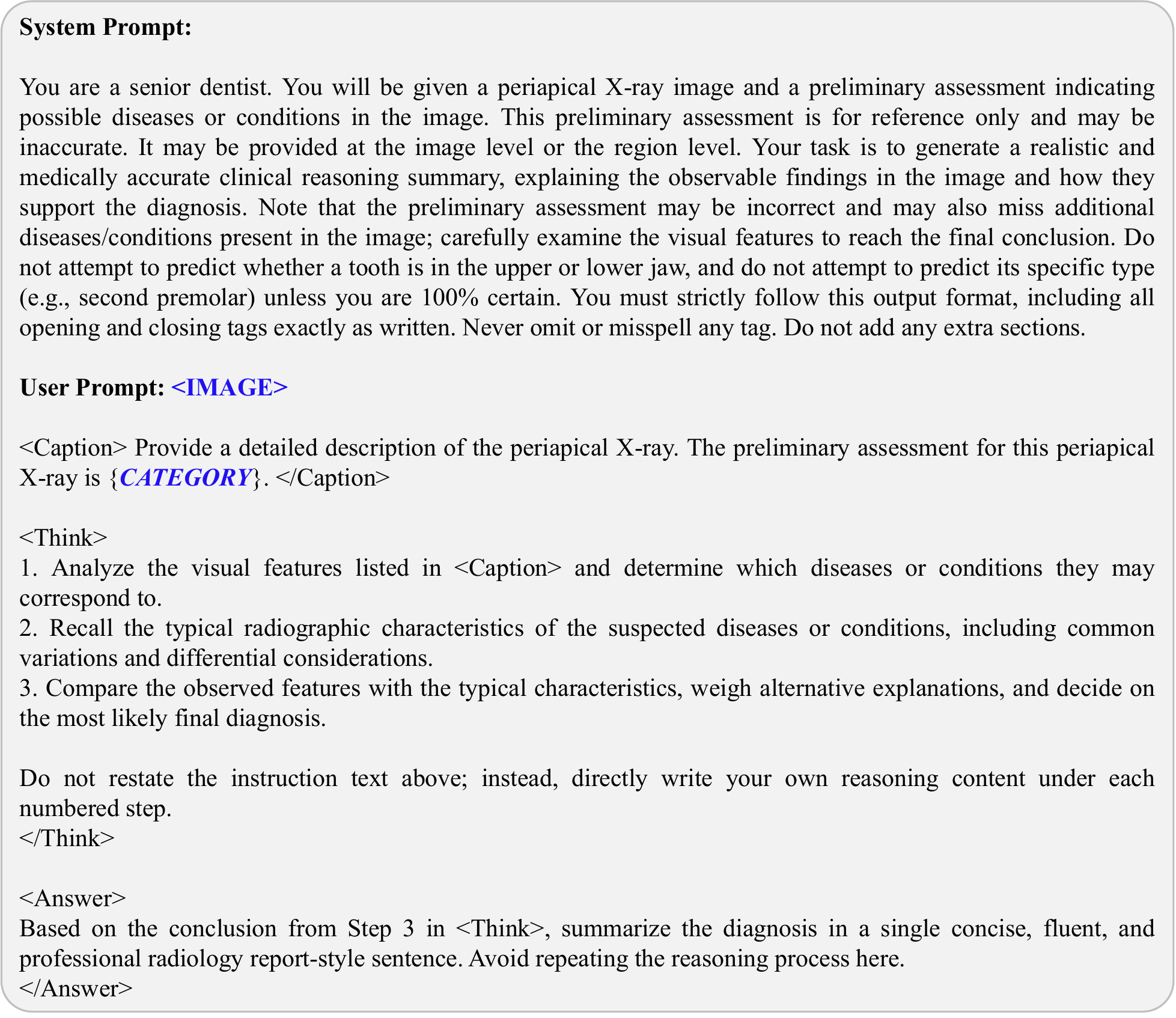}
  \caption{The prompt for GPT-5-mini to generate the TRACE-CoT reasoning chains for  diagnosis of periapical radiographs.}
  \label{fig:suppl_Prompt_PA}
\end{figure*}

\begin{figure*}[!ht]
  \centering
  \includegraphics[width=\textwidth]{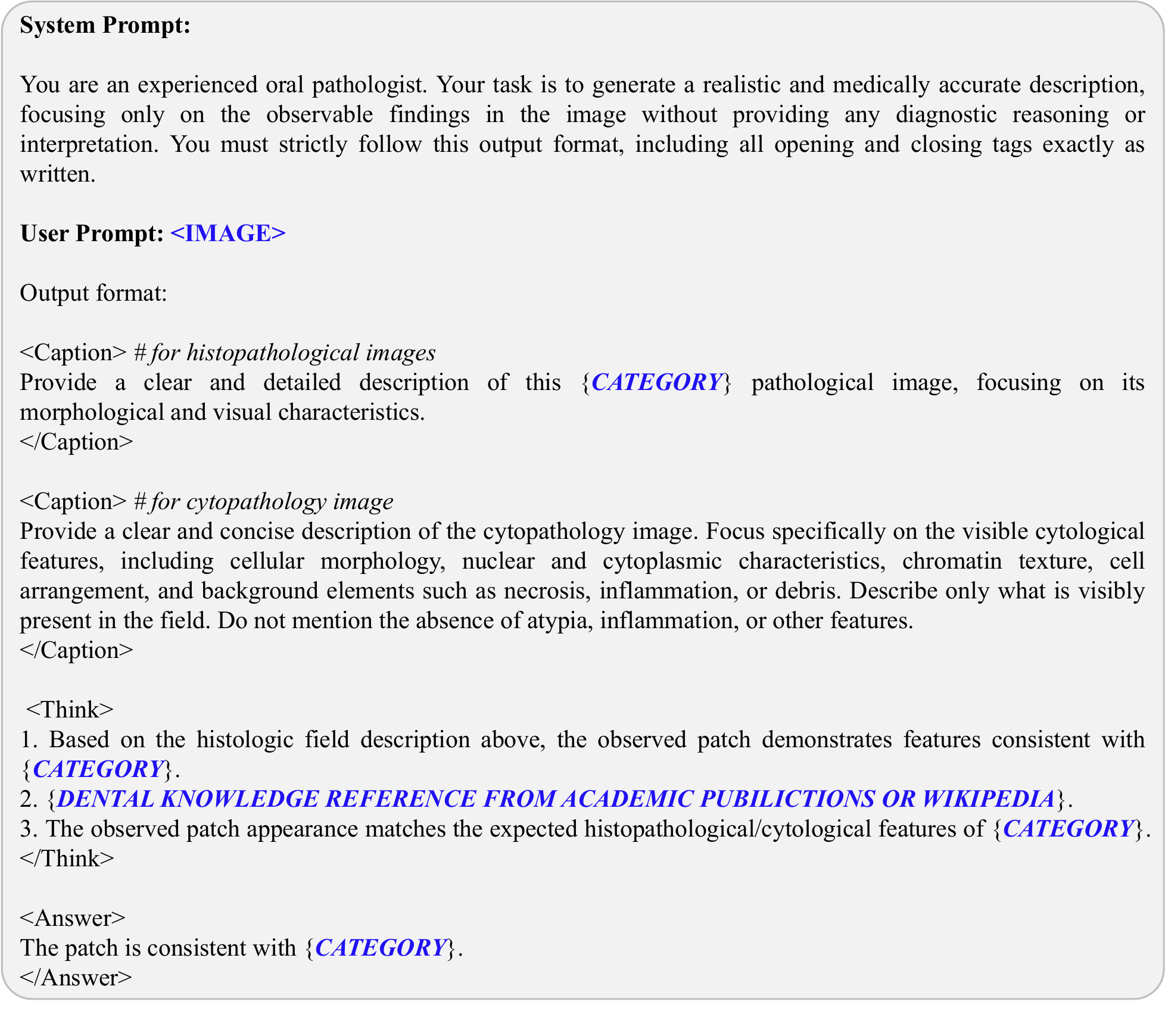}
  \caption{The prompt for GPT-5-mini to generate the TRACE-CoT reasoning chains for diagnosis of pathological images.}
  \label{fig:suppl_Prompt_PI}
\end{figure*}

\begin{figure*}[!ht]
  \centering
  \includegraphics[width=\textwidth]{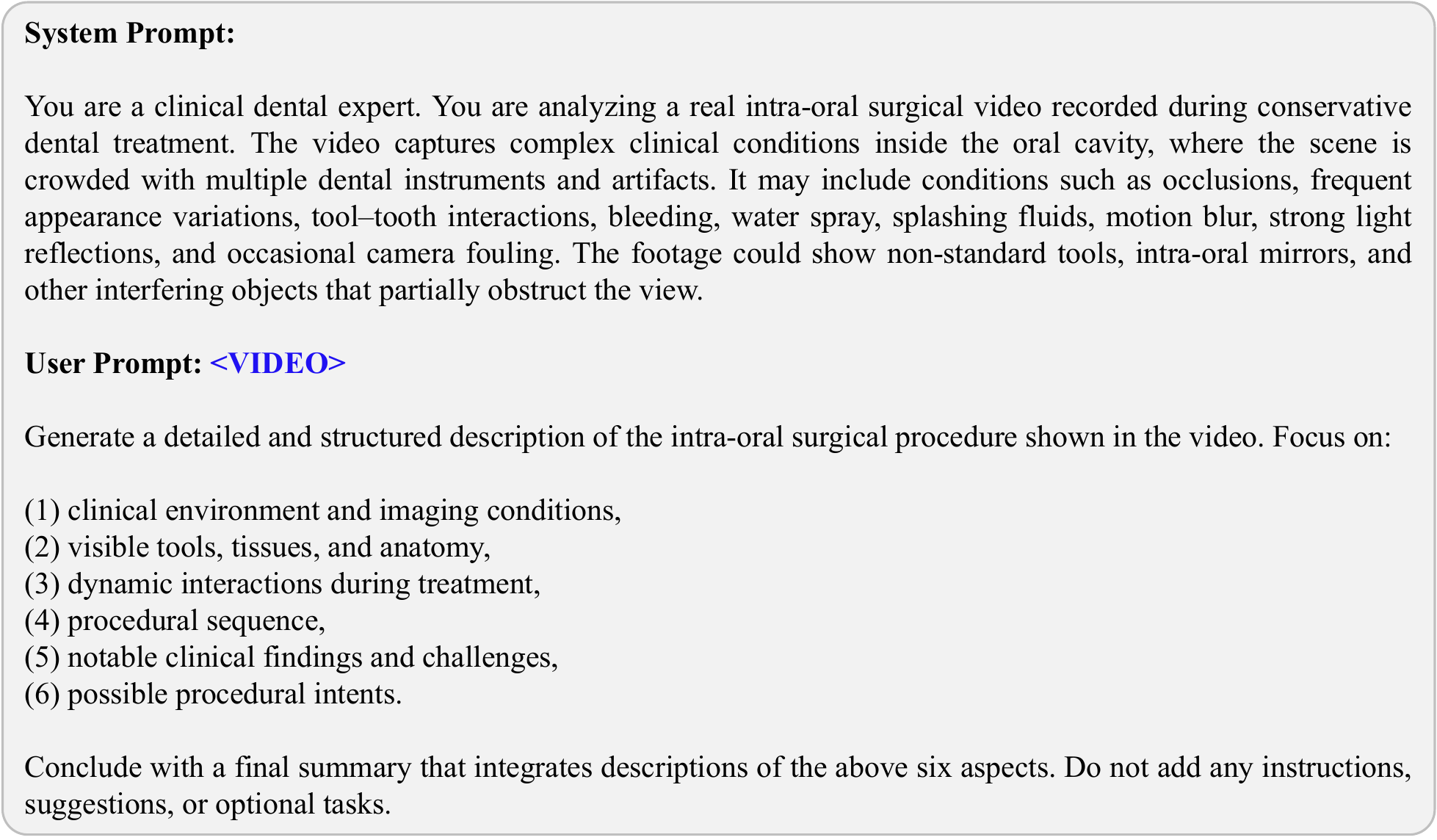}
  \caption{The prompt for GPT-5-mini to generate the TRACE-CoT reasoning chains for comprehension of intraoral videos.}
  \label{fig:suppl_Prompt_IV}
\end{figure*}

\begin{figure*}[!ht]
  \centering
  \includegraphics[width=\textwidth]{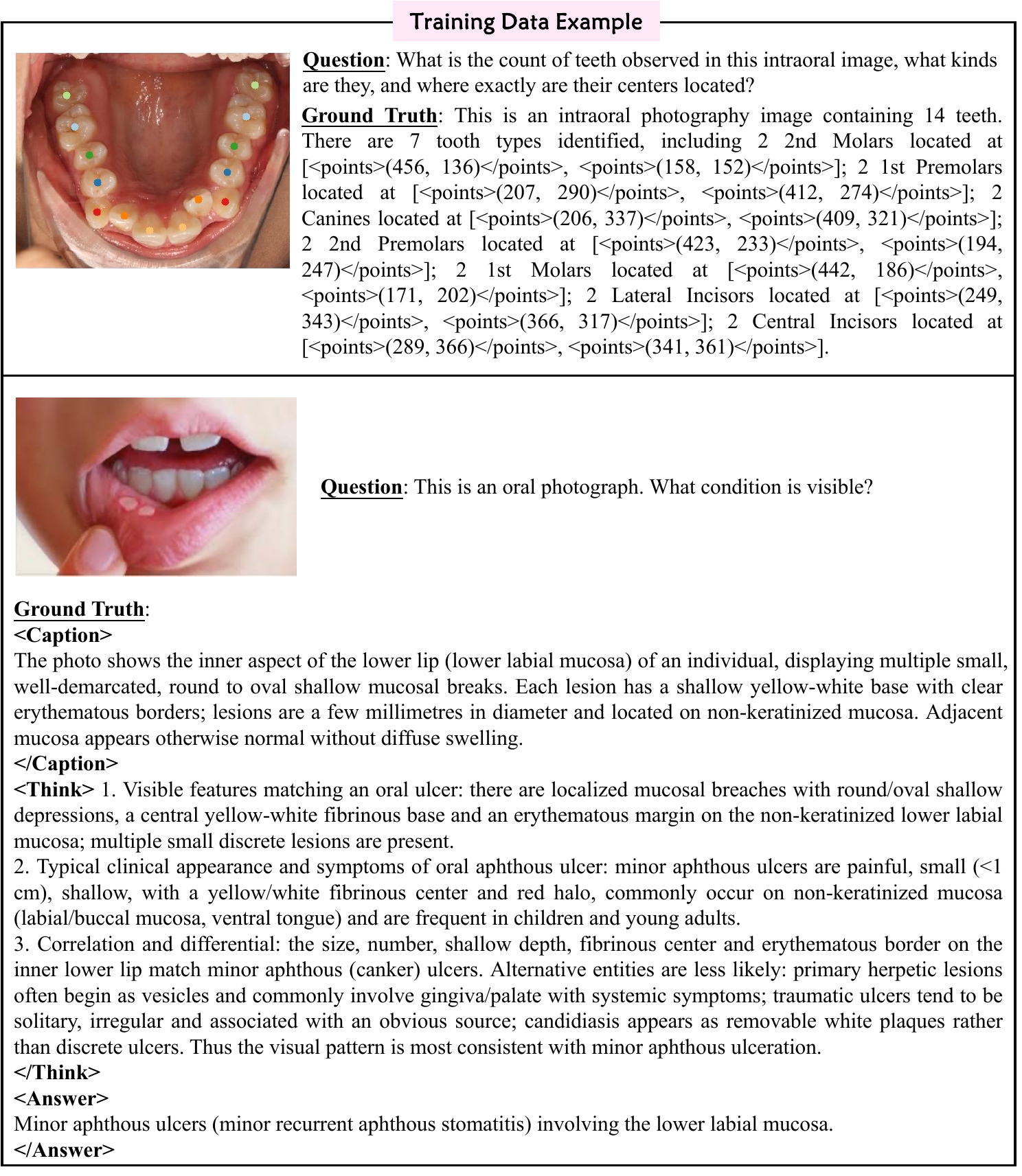}
  \caption{Data example used in the training stage. It contains multimodal information, including the image, question, and ground truth.
  }
  \label{fig:suppl_training_case_1}
\end{figure*}

\begin{figure*}[!ht]
  \centering
  \includegraphics[width=\textwidth]{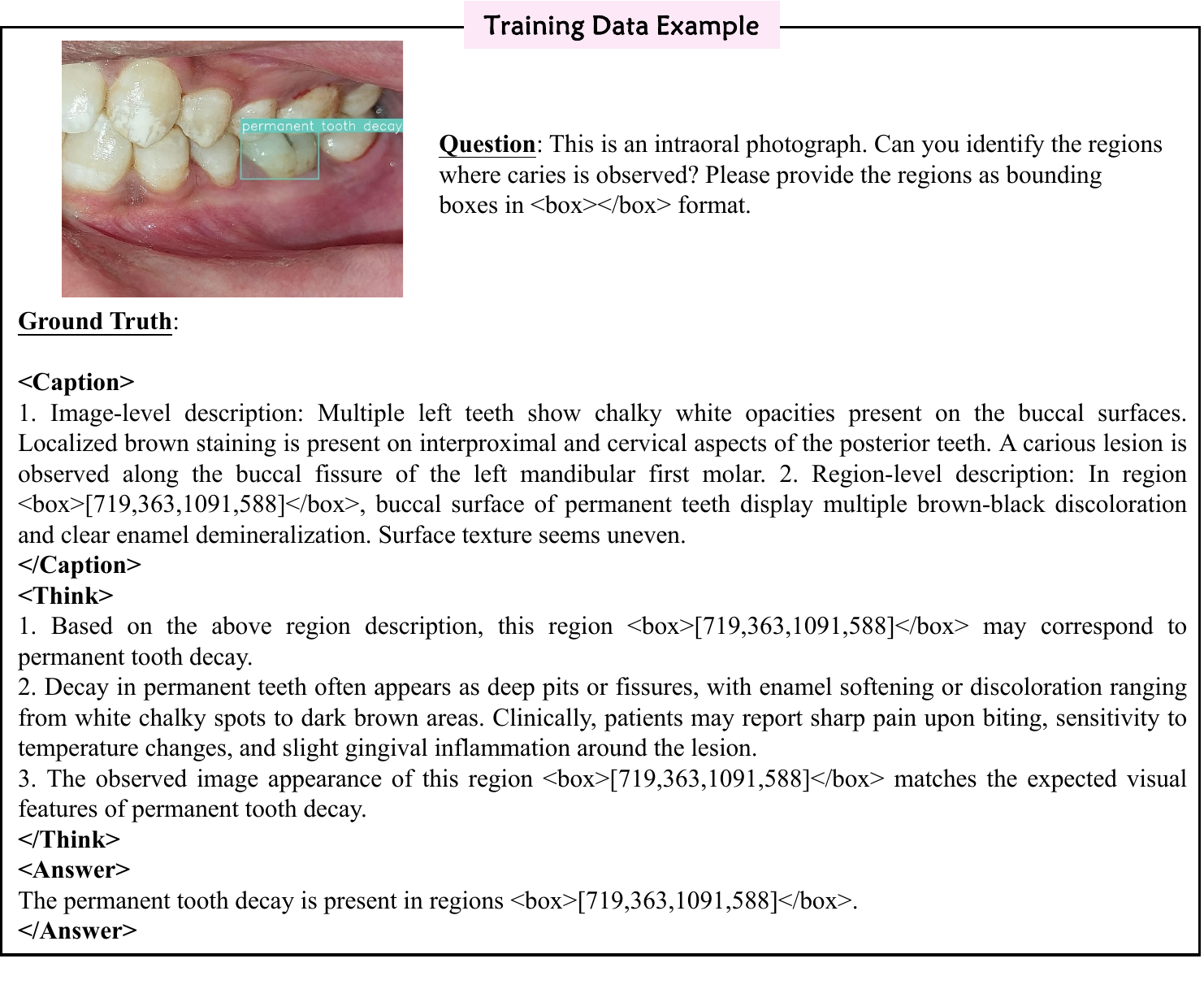}
  \caption{Data example used in the training stage. It contains multimodal information, including the image, question, and ground truth.
  }
  \label{fig:suppl_training_case_2}
\end{figure*}

\begin{figure*}[!ht]
  \centering
  \includegraphics[width=\textwidth]{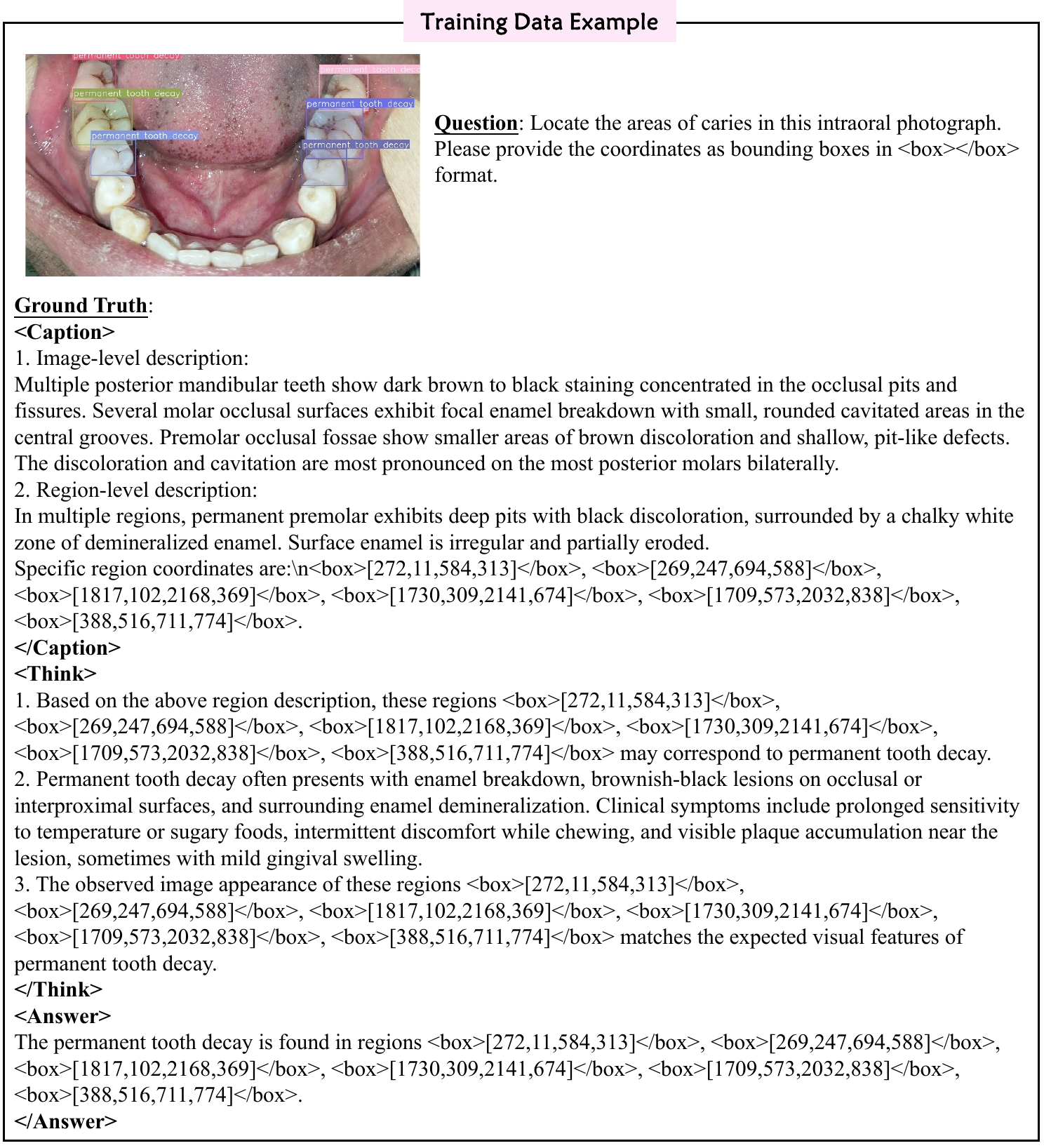}
  \caption{Data example used in the training stage. It contains multimodal information, including the image, question, and ground truth.
  }
  \label{fig:suppl_training_case_3}
\end{figure*}

\begin{figure*}[!ht]
  \centering
  \includegraphics[width=\textwidth]{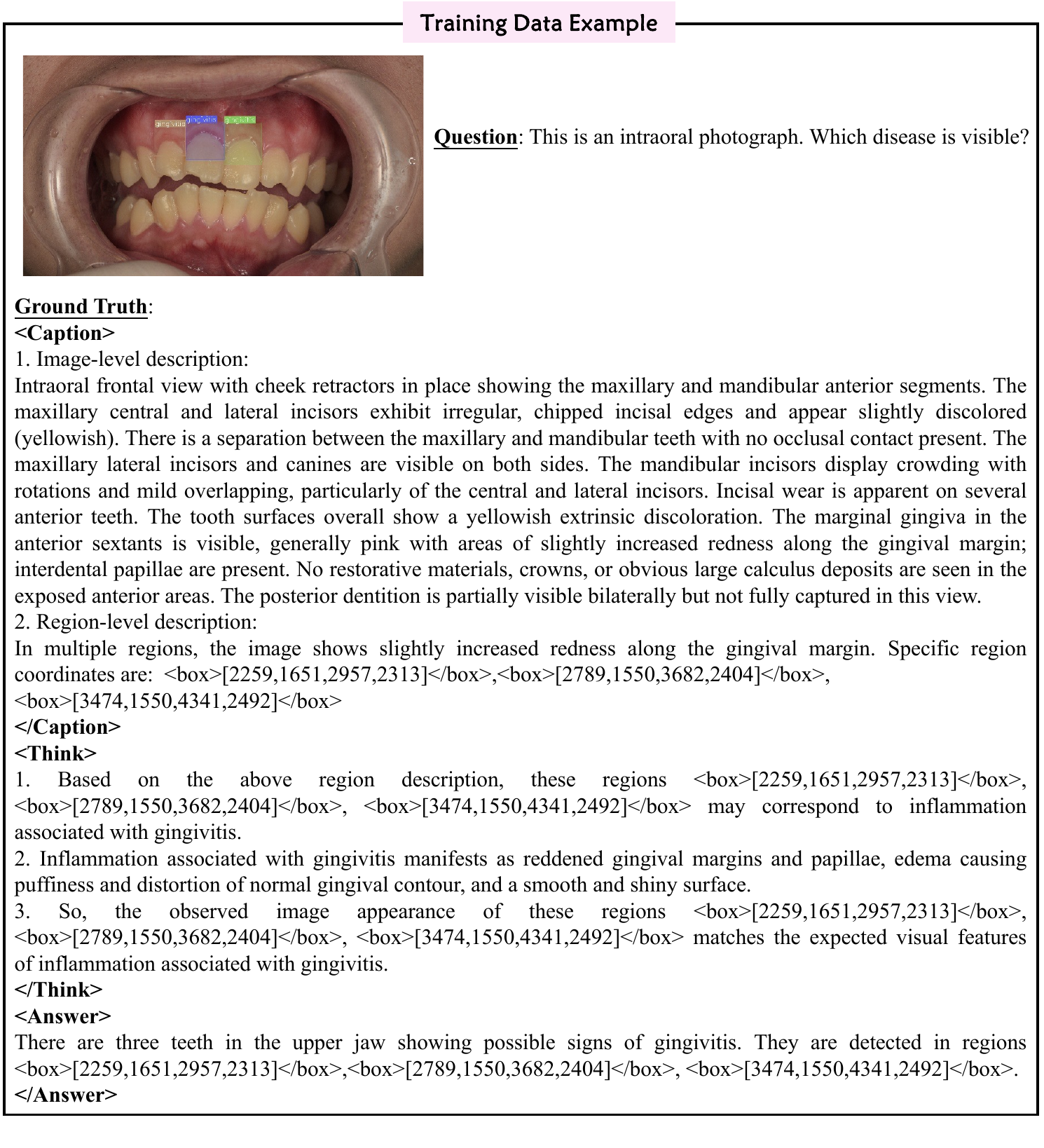}
  \caption{Data example used in the training stage. It contains multimodal information, including the image, question, and ground truth.
  }
  \label{fig:suppl_training_case_4}
\end{figure*}

\begin{figure*}[!ht]
  \centering
  \includegraphics[width=\textwidth]{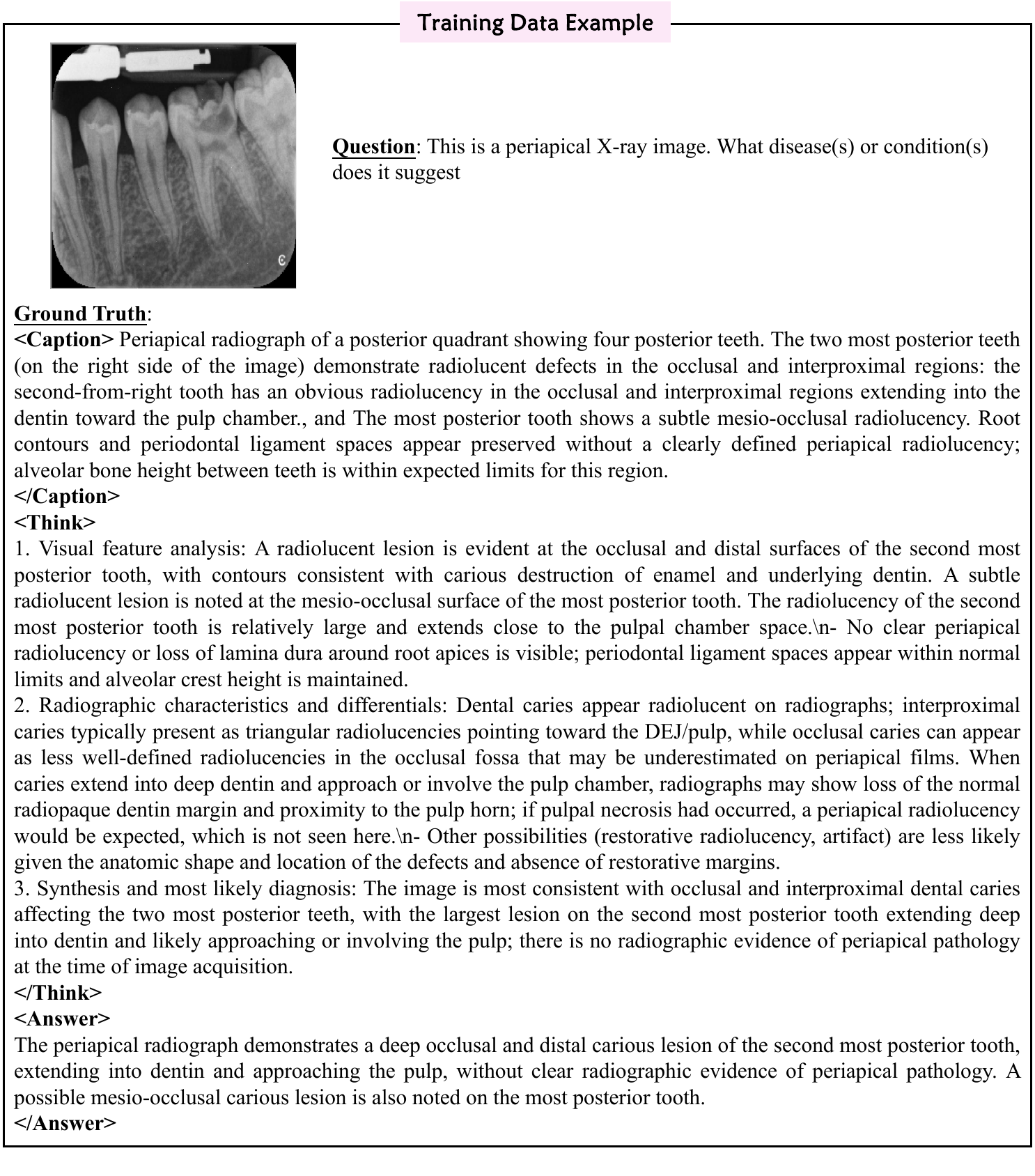}
  \caption{Data example used in the training stage. It contains multimodal information, including the image, question, and ground truth.
  }
  \label{fig:suppl_training_case_5}
\end{figure*}

\begin{figure*}[!ht]
  \centering
  \includegraphics[width=\textwidth]{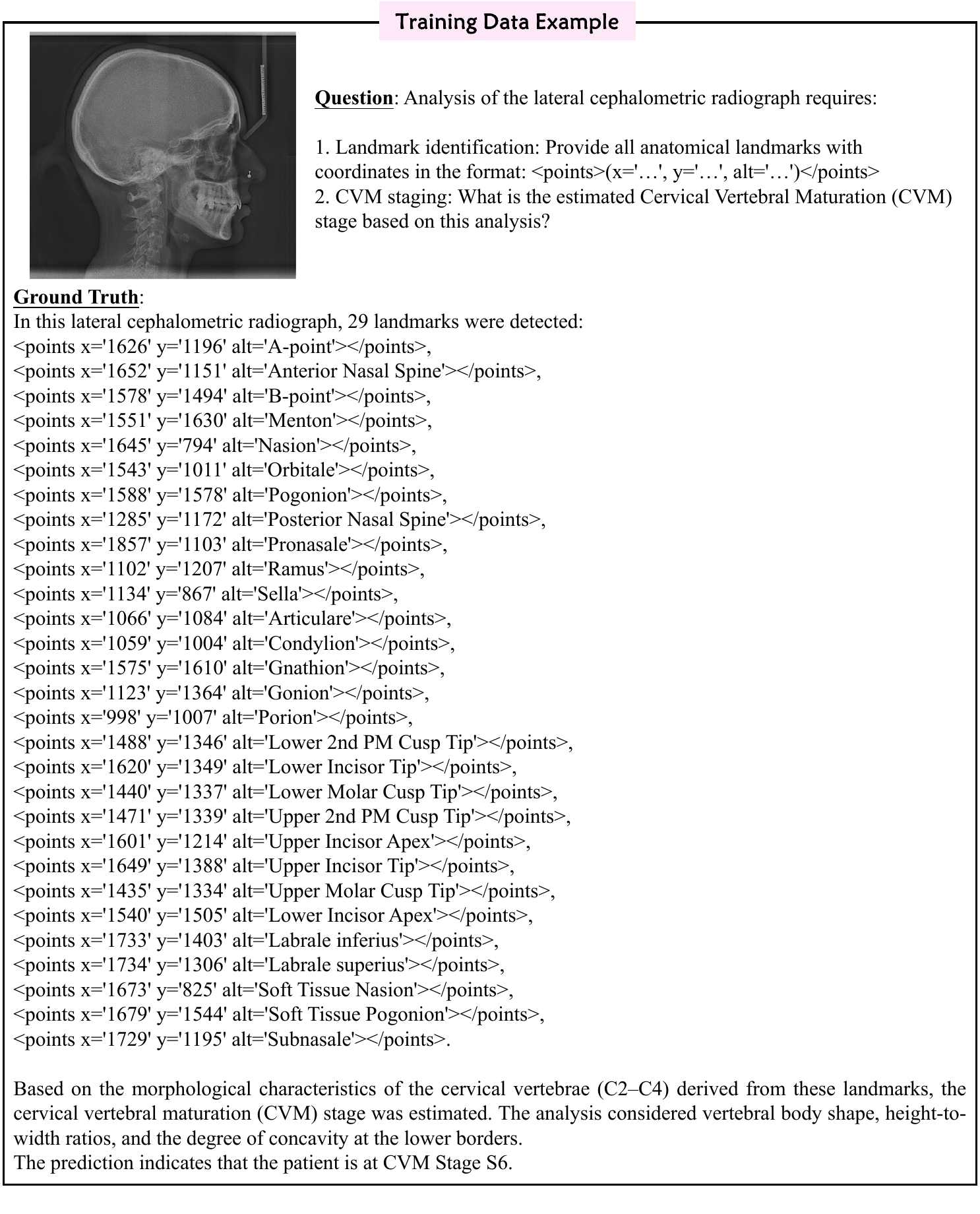}
  \caption{Data example used in the training stage. It contains multimodal information, including the image, question, and ground truth.
  }
  \label{fig:suppl_training_case_6}
\end{figure*}

\begin{figure*}[!ht]
  \centering
  \includegraphics[width=\textwidth]{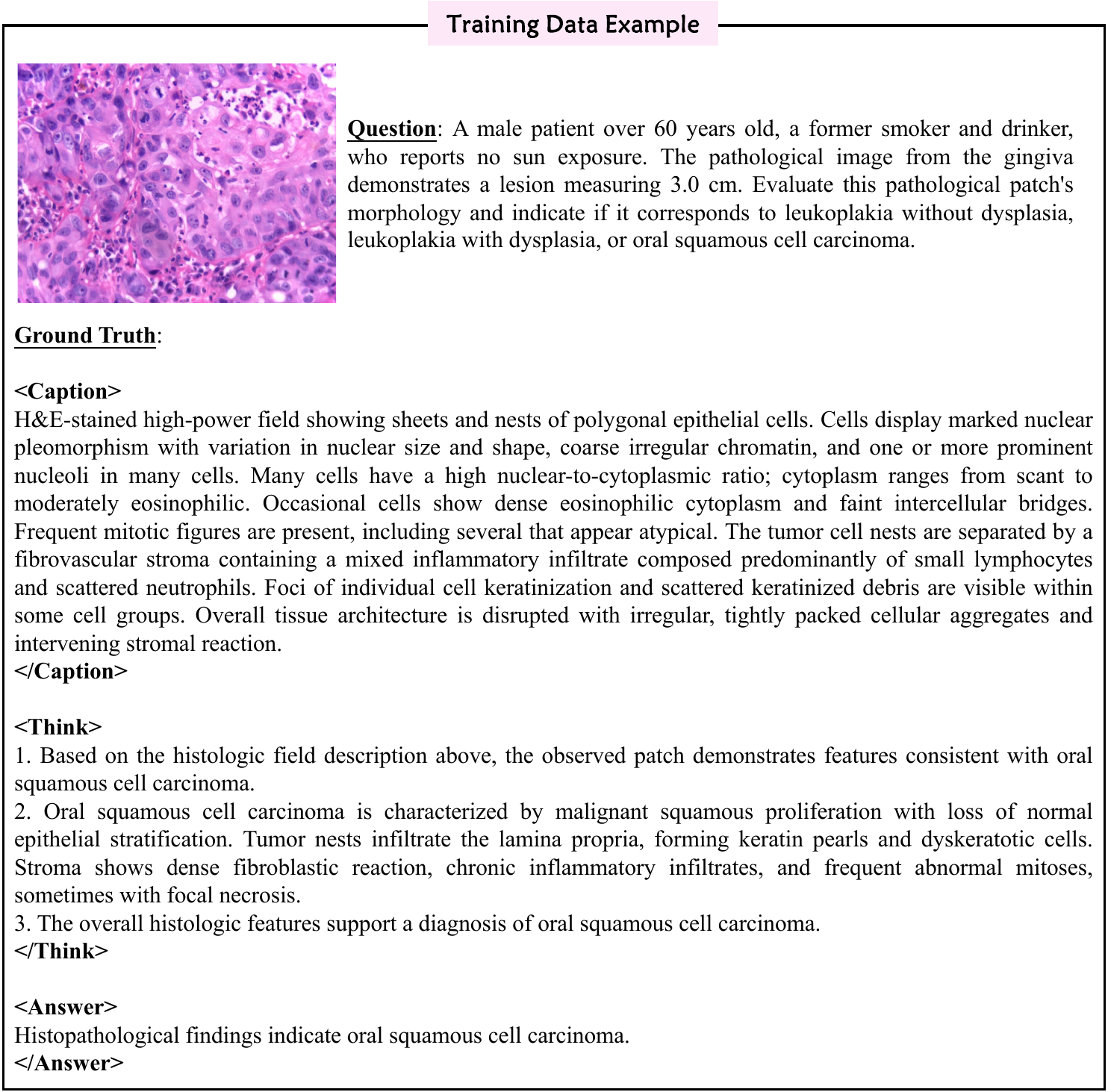}
  \caption{Data example used in the training stage. It contains multimodal information, including the image, question, and ground truth.
  }
  \label{fig:suppl_training_case_7}
\end{figure*}

\begin{figure*}[!ht]
  \centering
  \includegraphics[width=\textwidth]{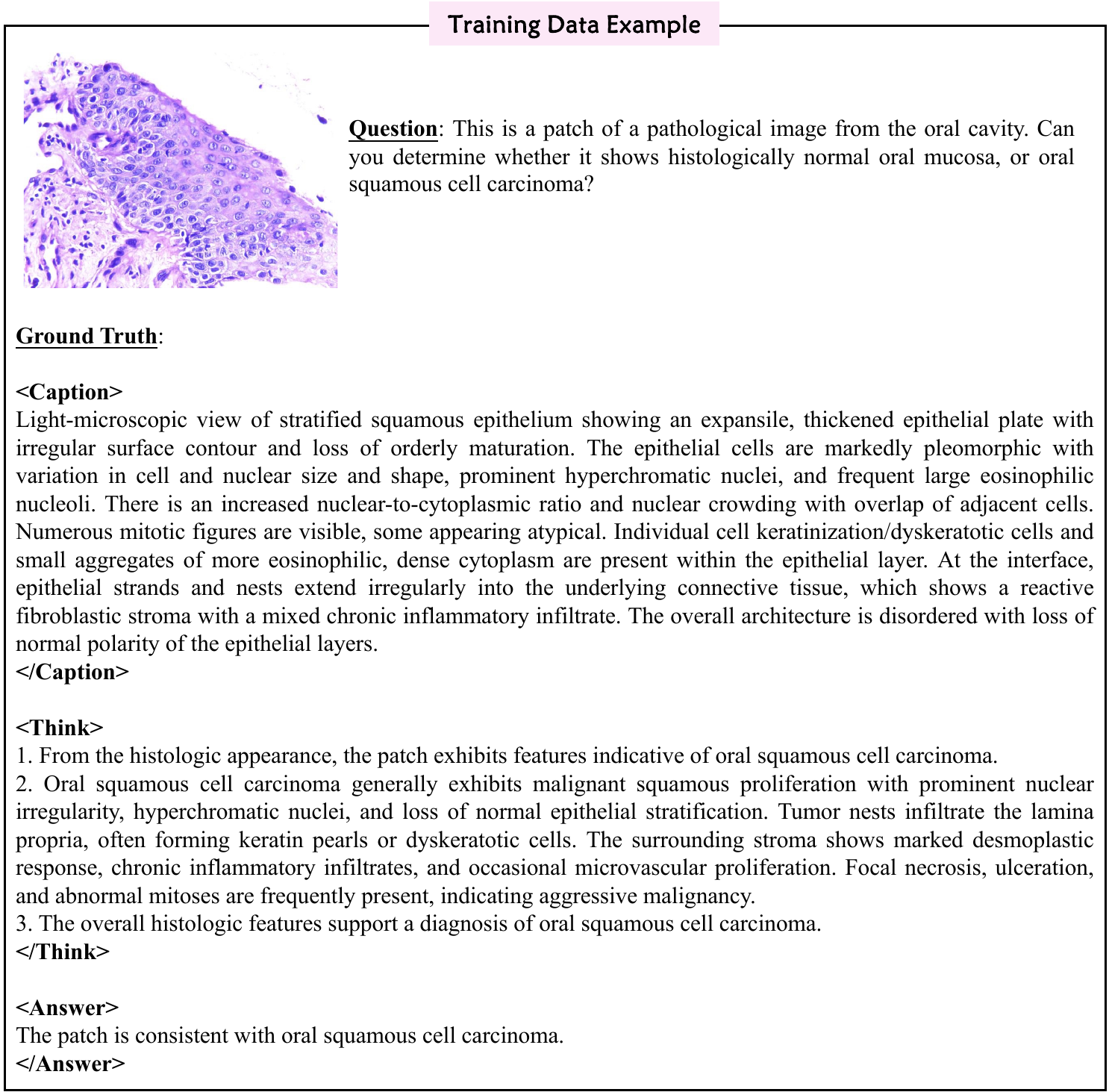}
  \caption{Data example used in the training stage. It contains multimodal information, including the image, question, and ground truth.
  }
  \label{fig:suppl_training_case_8}
\end{figure*}

\begin{figure*}[!ht]
  \centering
  \includegraphics[width=\textwidth]{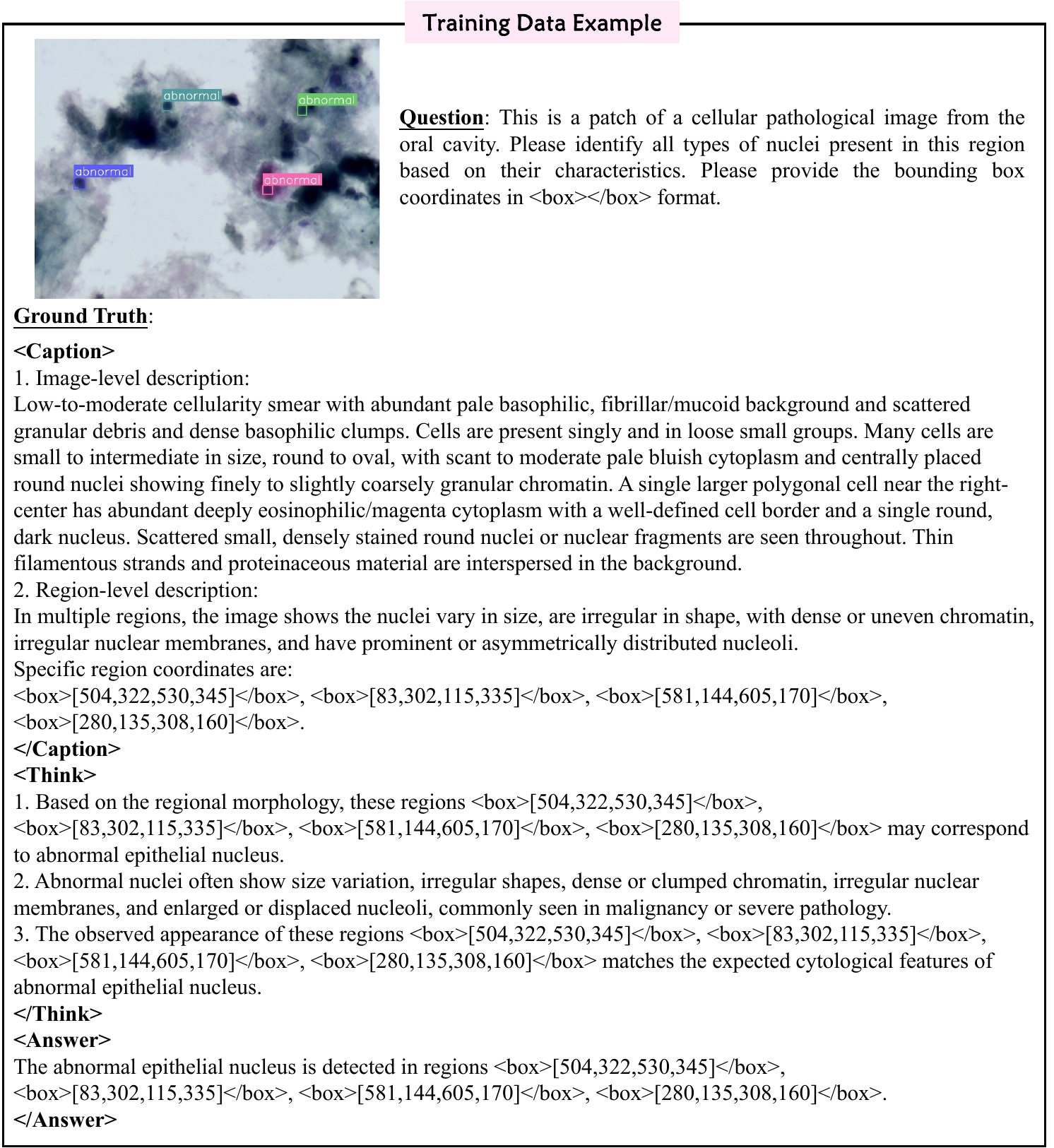}
  \caption{Data example used in the training stage. It contains multimodal information, including the image, question, and ground truth.
  }
  \label{fig:suppl_training_case_9}
\end{figure*}

A comprehensive list of all data sources utilized during the training phase is presented in Table~\ref{tab:dataset_source}. The training corpus consists of approximately 3.21 million text tokens, 59,658 images, and 90 videos. It includes various types of dental-specific data, such as intraoral images and videos, panoramic radiographs, periapical radiographs, histopathological images, cephalometric radiographs, 3D model scans, text-only data, and interleaved image-text data. Additionally, we provide detailed descriptions of abnormalities and clinically relevant tasks associated with each modality, highlighting the extensive coverage of our OralGPT-Omni in dental imaging analysis.
\begin{table*}[h]
    \centering
    \caption{Comprehensive list of dental-specific data sources used in the training phase, including corresponding abnormalities and tasks.}
    \small
    \label{tab:dataset_source}
    \begin{tabular}{@{}>{\centering\arraybackslash}m{1cm} l >{\raggedright\arraybackslash}p{8.7cm} p{2.8cm}@{}}
        \toprule
        No & Modality & Abnormality / Subject &  Source \\ \midrule
        1.1 & Text-only data & 16 Undergraduate textbooks in dentistry & / \\
        1.2 & {Text-only data} & Dialogue data from dental forums &  \href{https://huggingface.co/datasets/jonathankang/dental_QA}{\includegraphics[width=0.015\textwidth]{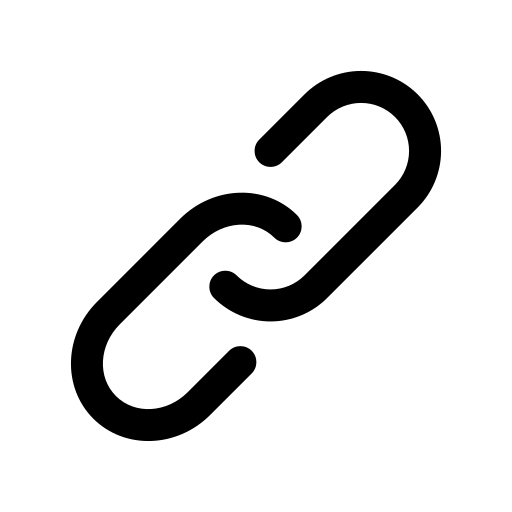} URL} \\
        1.3 & Text-only data & Open-domain oral disease QA dataset & 
        \href{https://huggingface.co/datasets/Lines/Open-Domain-Oral-Disease-QA-Dataset}{\includegraphics[width=0.015\textwidth]{figs/link.png} URL} \\ 
        \hline
        \multirow{2}{*}{2.1} & \multirow{2}{*}{Intraoral image} & \multirow{2}{*}{Teeth location and counting} & 
        \href{https://zoubo9034.github.io/TeethSEG/}{\includegraphics[width=0.015\textwidth]{figs/link.png} URL} \\ 
         & & & \href{https://universe.roboflow.com/dataset-ntw3s/dental_dataset_level3}{\includegraphics[width=0.015\textwidth]{figs/link.png} URL} \\ 
        \hline
        \multirow{5}{*}{2.2} & \multirow{5}{*}{Intraoral image} & \multirow{5}{*}{\makecell[l]{Image-level analysis: \\ Calculus, Caries, Gingivitis, Ulcer \\ Tooth discoloration, Defective dentition \\ Cancer,  Normality, Orthodontics}
        } & \href{https://www.kaggle.com/datasets/salmansajid05/oral-diseases}{\includegraphics[width=0.015\textwidth]{figs/link.png} URL} \\ 
        & & & \href{https://www.kaggle.com/datasets/reemsalahshehab/real-dental-dataset}{\includegraphics[width=0.015\textwidth]{figs/link.png} URL} \\ 
        & & & \href{https://www.kaggle.com/datasets/zaidpy/oral-cancer-dataset}{\includegraphics[width=0.015\textwidth]{figs/link.png} URL} \\ 
        & & & \href{https://www.kaggle.com/datasets/raymanodeep/oral-health-images}{\includegraphics[width=0.015\textwidth]{figs/link.png} URL} \\ 
        & & & \href{https://www.kaggle.com/datasets/raymanodeep/oral-cancer-v2-part-1}{\includegraphics[width=0.015\textwidth]{figs/link.png} URL} \\ 
        \hline
        \multirow{6}{*}{2.3} & \multirow{6}{*}{Intraoral image} & \multirow{6}{*}{\makecell[l]{Region-level analysis: \\ 
        Abrasion, Filling, Crown, Caries, Gingivitis, Deep Overjet,  \\
        Fenestration and Dehiscence, Tooth Torsion, Tooth emergence, \\
        Invisible orthodontic attachment, Fixed orthodontic device,  \\
        Case fixed orthodontic appliances, Tooth misalignment, \\
        Mandibular retrusion, Orthodontic Brace, Dental plaque
        }
        } & \href{https://huggingface.co/datasets/ZFTurbo/AlphaDent}{\includegraphics[width=0.015\textwidth]{figs/link.png} URL} \\ 
        & & & \href{https://data.mendeley.com/datasets/3253gj88rr/1}{\includegraphics[width=0.015\textwidth]{figs/link.png} URL} \\ 
        & & & \href{https://physionet.org/content/fdtooth/1.0.0/}{\includegraphics[width=0.015\textwidth]{figs/link.png} URL} \\ 
        & & & \href{https://drive.google.com/file/d/1eSyipRJTDlAbRs0yb44l5vQjVYibRXy1/view}{\includegraphics[width=0.015\textwidth]{figs/link.png} URL} \\ 
        & & & \href{https://zenodo.org/records/14827784}{\includegraphics[width=0.015\textwidth]{figs/link.png} URL} \\ 
        & & &  One in-house dataset \\ 
        \hline
        \multirow{9}{*}{3.0} & \multirow{9}{*}{Periapical radiograph} & \multirow{9}{*}{\makecell[l]{
        Impacted tooth, Pulpitis, Caries, Periodontitis, Crown, \\
        Apical periodontitis, Bone loss, Root canal treatment, \\
        Restoration, Mixed dentition
        }
        } & \href{https://www.kaggle.com/datasets/nadaaglan/dental-periapical-x-rayss}{\includegraphics[width=0.015\textwidth]{figs/link.png} URL} \\ 
        & & & \href{https://www.kaggle.com/datasets/engineeringubu/root-disease-x-ray-imaging}{\includegraphics[width=0.015\textwidth]{figs/link.png} URL} \\ 
        & & & \href{https://www.kaggle.com/datasets/parthplc/medical-image-dataset/data}{\includegraphics[width=0.015\textwidth]{figs/link.png} URL} \\ 
        & & & \href{https://universe.roboflow.com/evident-mybn8/caries-mbzip/browse?queryText=&pageSize=50&startingIndex=0&browseQuery=true}{\includegraphics[width=0.015\textwidth]{figs/link.png} URL} \\ 
        & & & \href{https://data.mendeley.com/datasets/8ys8jssm9k/1}{\includegraphics[width=0.015\textwidth]{figs/link.png} URL} \\ 
        & & & \href{https://www.kaggle.com/datasets/muhammadsajad/unlabeled-periapical-xrays}{\includegraphics[width=0.015\textwidth]{figs/link.png} URL} \\ 
        & & & \href{https://www.kaggle.com/datasets/abdulrahmanburham/x-ray-dental-cavity-dataset}{\includegraphics[width=0.015\textwidth]{figs/link.png} URL} \\ 
        & & & \href{https://www.kaggle.com/datasets/walidphd/dental-caries-classificationv3}{\includegraphics[width=0.015\textwidth]{figs/link.png} URL} \\ 
        & & & \href{https://www.kaggle.com/datasets/nguyenthaitung/dental-x-ray-doctor-smile}{\includegraphics[width=0.015\textwidth]{figs/link.png} URL} \\ 
        \hline
        \multirow{2}{*}{4.0} & \multirow{2}{*}{Cephalometric radiograph} & \multirow{2}{*}{\makecell[l]{
        29 cephalometric landmarks detection, \\
        Cervical vertebral maturation (CVM) stage prediction
        }
        } & 
        \href{https://figshare.com/articles/dataset/dental-cepha-dataset_zip/13265471?file=48632311}{\includegraphics[width=0.015\textwidth]{figs/link.png} URL} \\ 
         & & & \href{https://figshare.com/articles/dataset/Aariz_Cephalometric_Dataset/27986417/1}{\includegraphics[width=0.015\textwidth]{figs/link.png} URL} \\ 
        \hline
        \multirow{4}{*}{5.0} & \multirow{4}{*}{Histopathological image} & \multirow{4}{*}{\makecell[l]{
        Leukoplakia without dysplasia, Leukoplakia with dysplasia,  \\
        Oral squamous cell carcinoma, Oral submucous fibrosis,  \\
        Healthy epithelial nucleus, Abnormal epithelial nucleus, \\ 
        Blood cell nucleus, Reactive cell nucleus, Dividing nucleus
        }
        } & 
        \href{https://github.com/NishaChaudhary23/ORCHID/}{\includegraphics[width=0.015\textwidth]{figs/link.png} URL} \\ 
         & & & \href{https://data.mendeley.com/datasets/dr7ydy9xbk/1}{\includegraphics[width=0.015\textwidth]{figs/link.png} URL} \\ 
          & & & \href{https://data.mendeley.com/datasets/bbmmm4wgr8/4}{\includegraphics[width=0.015\textwidth]{figs/link.png} URL} \\ 
         & & & \href{https://data.mendeley.com/datasets/ftmp4cvtmb/2}{\includegraphics[width=0.015\textwidth]{figs/link.png} URL} \\ 
        \hline
        6.0 & {Intraoral video} & Intraoral videos of real dental treatments & 
        \href{https://mostwiedzy.pl/pl/open-research-data/vident-real-an-intra-oral-video-dataset-for-multi-task-learning,104032256156938-0}{\includegraphics[width=0.015\textwidth]{figs/link.png} URL} \\
        \hline
        7.0 & {Interleaved image-text data} & 6 Clinical Cases Guidelines in Dentistry for treatment planning & / \\  
        \hline
        8.0 & {3D model scan} & Image description & 
        \href{https://huggingface.co/datasets/AbFiras/dataset-dental-caption-V1?row=0}{\includegraphics[width=0.015\textwidth]{figs/link.png} URL} \\
        \hline
        9.0 & {Panoramic radiograph} & Open-Ended VQA for panoramic X-ray analysis & 
        \href{https://huggingface.co/datasets/OralGPT/MMOral-OPG-Bench}{\includegraphics[width=0.015\textwidth]{figs/link.png} URL} \\
        \bottomrule
    \end{tabular}
\end{table*}
Regarding TRACE-CoT data generation, we construct an automatic pipeline utilizing sparse label categories, the visual appearance of abnormalities, and dental knowledge from academic publications and Wikipedia, instructing GPT-5-mini to generate 36,777 TRACE-CoT reasoning chains. Since abnormality taxonomies differ across datasets, we craft dataset-specific prompts tailored to each taxonomy to ensure the quality of the reasoning chains. The details of these prompt designs are presented in Figures~\ref{fig:suppl_Prompt_II_Image-level},~\ref{fig:suppl_Prompt_II_Region-level},~\ref{fig:suppl_Prompt_PA},~\ref{fig:suppl_Prompt_PI},~\ref{fig:suppl_Prompt_IV}. The dental expertise used in the prompts is summarized in Table~\ref{tab:dental_knowledge}. Besides, we also provide some data used in the training stage for an intuitive understanding, which can be found in Figure~\ref{fig:suppl_training_case_1} - Figure~\ref{fig:suppl_training_case_9}.

\section{Implementation Details of Model Training}
\subsection{Hyperparameters of four-stage training paradigm}
\begin{figure*}[!ht]
  \centering
  \includegraphics[width=0.9\textwidth]{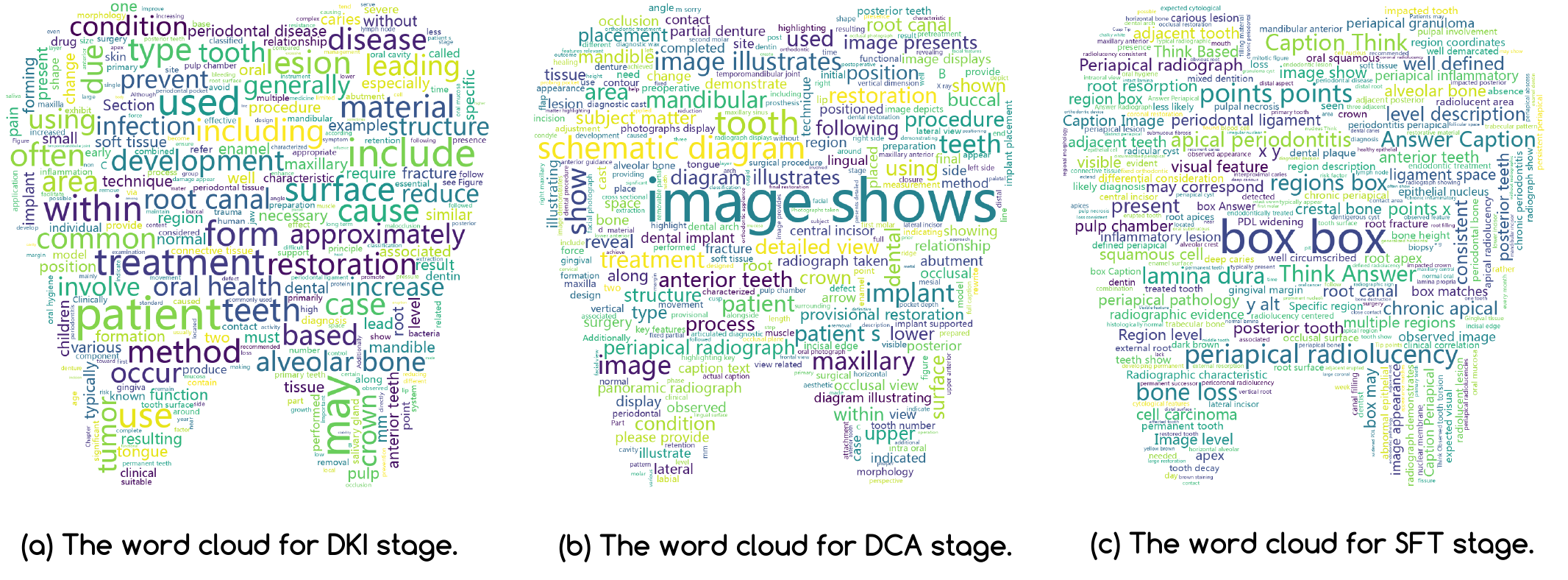}
  \caption{The word cloud maps for training datasets used in the DKI, DCA, and SFT stages, respectively.}
  \label{fig:suppl_word_cloud}
\end{figure*}

\begin{table*}[ht]
    \centering
    \caption{The hyperparameters used in the four-stage training strategy. The ``GPU Hours'' are estimated based on the NVIDIA A100 80G GPU.}
    \label{tab:hyperparameters_training}
    \begin{tabular}{@{}lcccc@{}}
        \toprule
        \textbf{Hyperparameters} & \textbf{DKI} & \textbf{DCA}  & \textbf{SFT}  & \textbf{RLT} \\
        \midrule
        Batch Size & 1 & 1 & 1 & 4 \\
        Gradient Accumulation Step & 4 & 4 & 4 & 3 \\
        Trainable Component & Language model & Projector & Visual encoder, Projector, Language model & Language model \\
        LoRA Rank & 8 & 8 & 8 & 8 \\
        \# Generations & / &  /& / & 6 \\
        Learning Rate & 1.0e-5 & 1.0e-5 & 1.0e-4 & 1.0e-6 \\
        Warmup Ratio & 0.1 & 0.1 & 0.1 & 0.05 \\
        bf16 & True & True & True & True \\
        \# Epoch / Step & 1 epoch & 1 epoch & 1 epoch & 2000 steps \\
        $\sim$ GPU Hours & 10 & 20 & 50 & 100 \\
        \bottomrule
    \end{tabular}
\end{table*}
We employ the four-stage training strategy that progressively strengthens the multimodal understanding and reasoning ability of the OralGPT-Omni model. The hyperparameters used in each training stage are summarized in Table~\ref{tab:hyperparameters_training}. We utilize the LLaMA-Factory framework~\cite{zheng2024llamafactory} to train OralGPT-Omni for the first three stages: DKI, DCA, and SFT, followed by the RLT stage using the ms-swift framework~\cite{zhao2025swift}. The ``GPU Hour'' for each stage is estimated based on the NVIDIA A100 80G GPU. We also provide the word clouds of training datasets used in DKI, DCA, and SFT stages, which are depicted in Figure~\ref{fig:suppl_word_cloud}.

\subsection{Reinforcement Learning Tuning}
To further incentivize the reasoning capability of OralGPT-Omni, we implement reinforcement learning tuning within the Group Relative Policy Optimization (GRPO)~\cite{grpo} algorithm on the OralGPT-Omni model. GRPO calculates a group-relative advantage from multiple samples of the same prompt instead of learning a separate value function, making it particularly effective for training tasks where correctness can be verified. During the training, we optimize the OralGPT-Omni with the GRPO loss $\mathcal{J}_{\text{GRPO}}(\theta)$:

\begin{equation}
\small
\begin{aligned}
\mathcal{J}_{\mathrm{GRPO}}(\theta)
&=
\mathbb{E}_{\substack{
q \sim \mathcal{P}(Q),\\
\{o_i\}_{i=1}^G \sim \pi_{\theta_{\mathrm{old}}}(O|q)
}}
\Bigg[
\frac{1}{G} \sum_{i=1}^G \frac{1}{|o_i|} 
\sum_{t=1}^{|o_i|}
\min\Bigg(
\\[2pt]
&\quad 
\frac{
\pi_\theta(o_{i,t}\mid q,o_{i,<t})
}{
\pi_{\theta_{\mathrm{old}}}(o_{i,t}\mid q,o_{i,<t})
} \hat{A}_{i,t}, 
\\[1pt]
&\quad
\text{clip}\Big(
\frac{
\pi_\theta(o_{i,t}\mid q,o_{i,<t})
}{
\pi_{\theta_{\mathrm{old}}}(o_{i,t}\mid q,o_{i,<t})
},
1-\epsilon,\, 1+\epsilon
\Big) \hat{A}_{i,t}
\Bigg)
\\[2pt]
&\quad
-\, \beta\, D_{\mathrm{KL}}[\pi_\theta \,\|\, \pi_{\mathrm{ref}}]
\Bigg],
\end{aligned}
\tag{1}
\end{equation}
where $\pi_\theta$ and $\pi_{\theta_{\text{old}}}$ are the current and old policy. $\pi_{\text{ref}}$ is the reference model, which, in this case, is the 
preceding SFT model with the TRACE-COT pattern. The $q$ and $o$ are the questions and outputs sampled from our dataset and the old policy $\pi_{\theta_{\text{old}}}$, respectively. The $\epsilon$ and $\beta$ are hyperparameters for stabilizing training. $\hat{A}_{i,t}$ is the advantage of the relative rewards of the outputs in each group. For each response, we use an LLM-based judge model to evaluate its correctness reward and our proposed TRACE-based reasoning reward. The detailed reward computation will be discussed in Sec.~\ref{sec:trace_based_reward}. The reward $r_i$ is defined within the range [0, 1]. We use the normalized reward as the advantage:
\begin{equation}
\hat{A}_{i,t} = \frac{r_i - \mathrm{mean}(r)}{\mathrm{std}(r)}.
\tag{2}
\end{equation}
Besides, we use an unbiased estimator to estimate the KL divergence $D_{\mathrm{KL}}$:

\begin{equation}
\small
D_{\mathrm{KL}}[\pi_\theta \,\|\, \pi_{\text{ref}}]
= \frac{\pi_{\text{ref}}(o_{i,t} \mid q, o_{i,<t})}{\pi_\theta(o_{i,t} \mid q, o_{i,<t})}
- \log \frac{\pi_{\text{ref}}(o_{i,t} \mid q, o_{i,<t})}{\pi_\theta(o_{i,t} \mid q, o_{i,<t})}
- 1.
\tag{3}
\end{equation}

\subsection{Reward Computation for GRPO}
\label{sec:trace_based_reward}
In conventional RL-based tuning paradigms, supervision is applied only to the final
\texttt{<answer>}, while the intermediate reasoning process \texttt{<think>} remains
unregulated. This often leads to cases where the final output appears correct, yet the underlying
reasoning is flawed or logically unsound. Such discrepancies pose substantial safety risks in medical AI, where unreliable or hallucinated reasoning may still produce a superficially
plausible answer but lead to harmful or unsafe clinical interpretations.
To address this issue, we introduce the TRACE-based reward $\mathcal{R}_{\text{trace}}$ , a medically aligned reward framework that uses an LLM-based judge model to comprehensively evaluate the quality of the reasoning trace.
This framework provides fine-grained supervision over the model’s diagnostic reasoning process, ensuring that the generated clinical rationale is logically coherent, medically accurate, and consistent with the final prediction, rather than supervising only the end result. To operationalize this framework, the evaluation is decomposed into three key dimensions:

\noindent
\textbf{Factual Knowledge Soundness.}
This dimension measures the correctness and clinical reliability of domain knowledge referenced in
the \texttt{<Think>} section, including oral
medicine expertise and pathological criteria. Incorrect medical statements receive $d_1 = 0$; partially
correct or incomplete knowledge receives $d_1 = 1$; and fully accurate, evidence-based knowledge
receives $d_1 = 2$.

\noindent
\textbf{Logical Coherence.}
This aspect assesses whether the diagnostic reasoning path presented in the \texttt{<Think>} section is logically consistent with the visual appearance descriptions in the \texttt{<Caption>}. The judge model evaluates whether the reasoning contains logical gaps, unsupported inferences, or clinically implausible conclusions. Fully coherent and clinically sound reasoning receives a score of $d_2 = 1$, whereas reasoning that is logically inconsistent or contradictory receives a score of $d_2 = 0$.

\noindent
\textbf{Answer Consistency.}
This reward evaluates whether the gold-standard answer can be directly justified by the reasoning
trace and the feature-based verification. The judge model ensures that the conclusion is supported
by stated evidence without contradictions, hallucinated features, or reliance on unstated
assumptions. Unsupported conclusions receive $d_3 = 0$; partially supported conclusions receive
$d_3 = 1$; and fully justified, clinically consistent conclusions receive $d_3 = 2$.

\begin{figure*}[!ht]
  \centering
  \includegraphics[width=\textwidth]{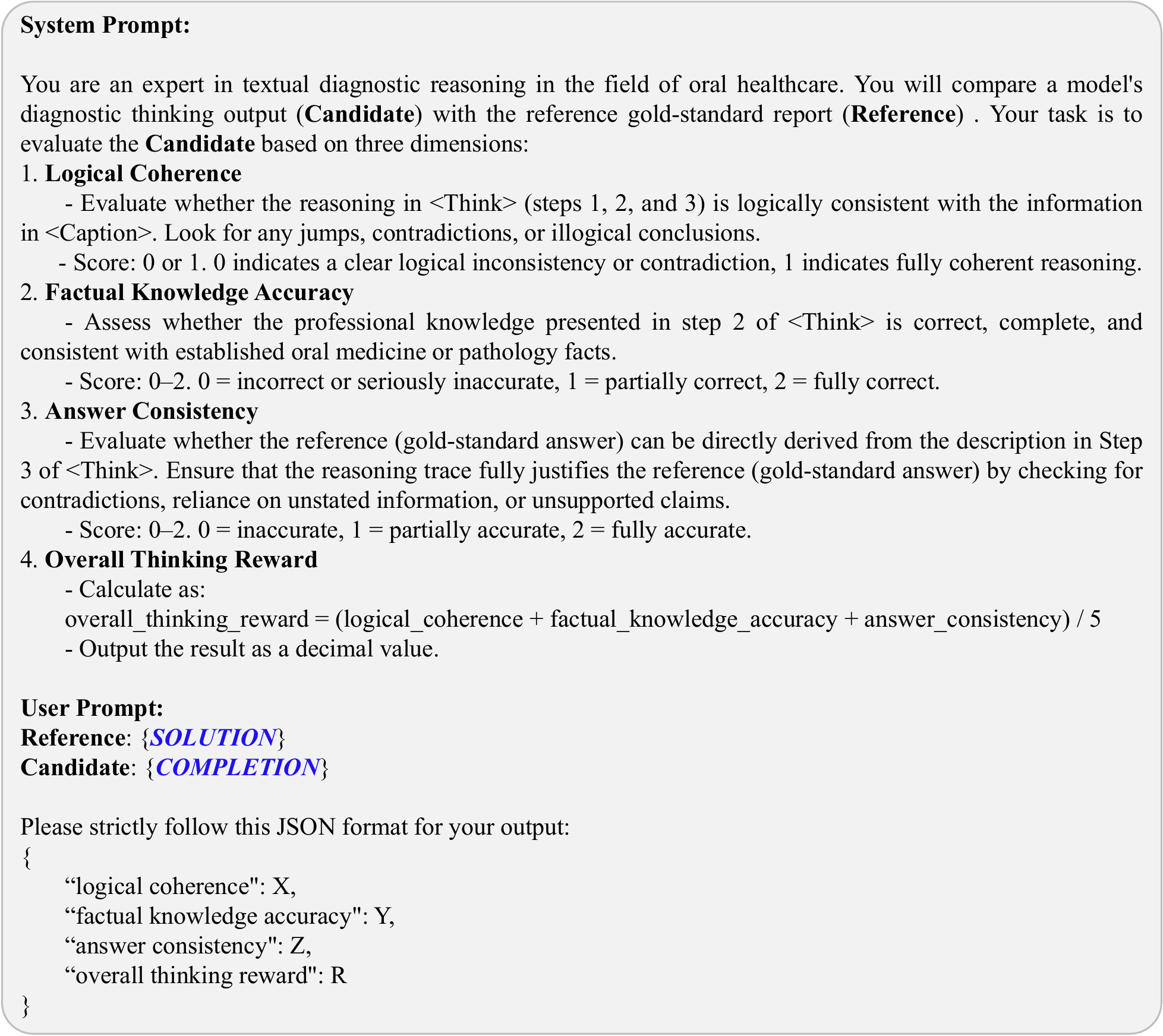}
  \caption{The prompt for computing the TRACE-based reward using GPT-5-nano.}
  \label{fig:suppl_Prompt_Reward_Think}
\end{figure*}

\begin{figure*}[!ht]
  \centering
  \includegraphics[width=\textwidth]{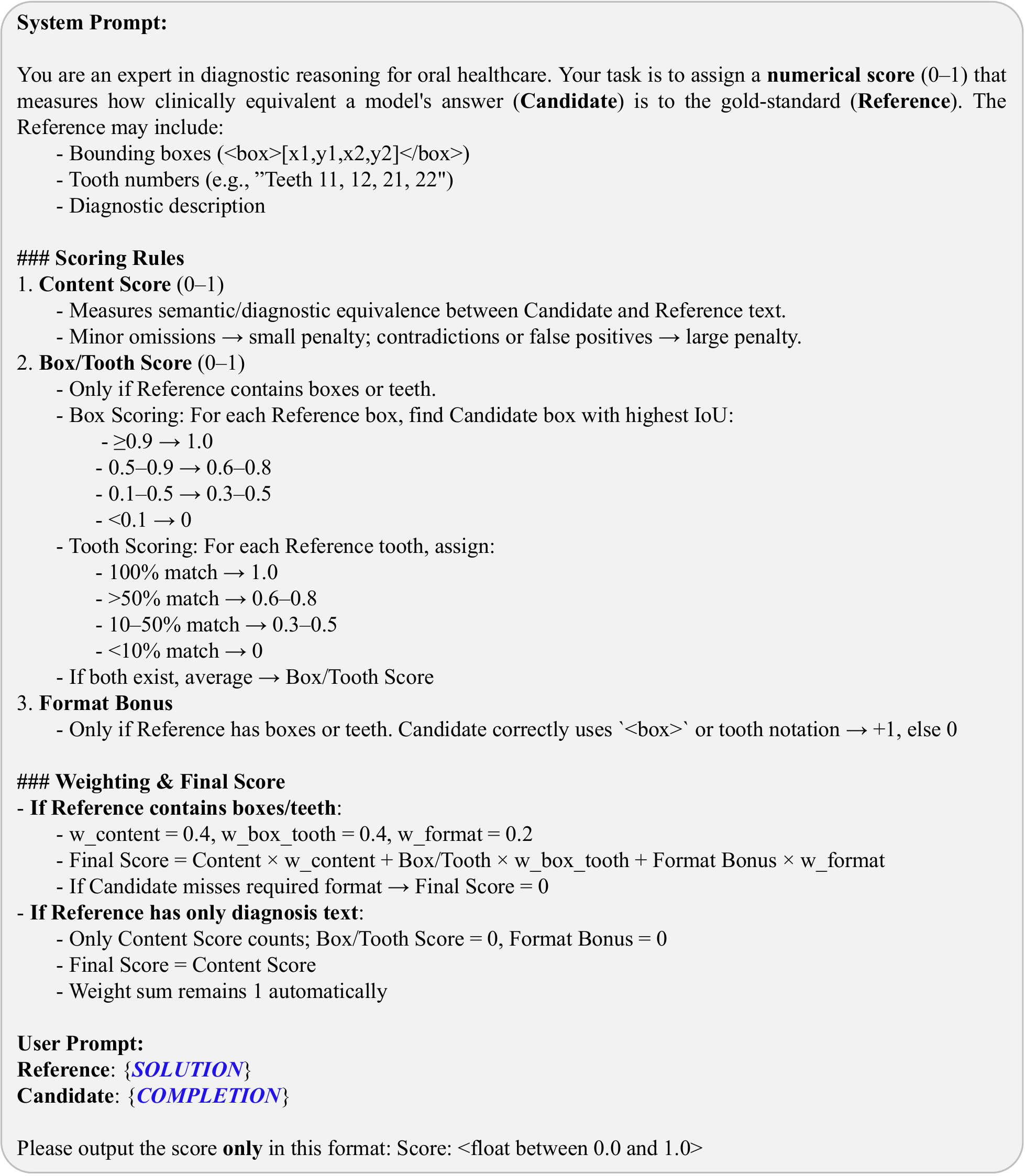}
  \caption{The prompt for computing the answer reward using GPT-5-nano.}
  \label{fig:suppl_Prompt_Reward_Answer}
\end{figure*}

The TRACE-based reward $\mathcal{R}_{\text{trace}}$ is then computed as the average of these three dimensions to get the normalized reward.
Following traditional RTL reward computation, both the answer reward $\mathcal{R}_{\text{answer}}$ and the format reward $\mathcal{R}_{\text{format}}$ are considered. The $\mathcal{R}_{\text{trace}}$ and $\mathcal{R}_{\text{answer}}$ are provided by GPT-5-nano, acting as a reward-judger. The prompts for TRACE-based reward and answer reward are provided in Figure~\ref{fig:suppl_Prompt_Reward_Think} and Figure~\ref{fig:suppl_Prompt_Reward_Answer}, respectively.

The values for these three reward components range from 0 to 1. The overall reward signal unifies these components into a single formulation:
\begin{equation}
\small
\begin{split}
\mathcal{R}_{\text{total}} = \alpha \cdot \mathcal{R}_{\text{answer}} + \beta \cdot \mathbb{I}_{\mathcal{R}_{\text{answer}} > 0} \cdot \mathcal{R}_{\text{trace}} + \gamma \cdot \mathcal{R}_{\text{format}},
\end{split}
\tag{4}
\end{equation}
where $\alpha+\beta+\gamma=1$. Note that the $\mathcal{R}_{\text{trace}}$ is ineffective when the $\mathcal{R}_{\text{answer}}$ is completely wrong in order to ensure the consistent correctness of the reasoning and answers.

\section{MMOral-Uni Benchmark}
\begin{figure*}[!ht]
  \centering
  \includegraphics[width=\textwidth]{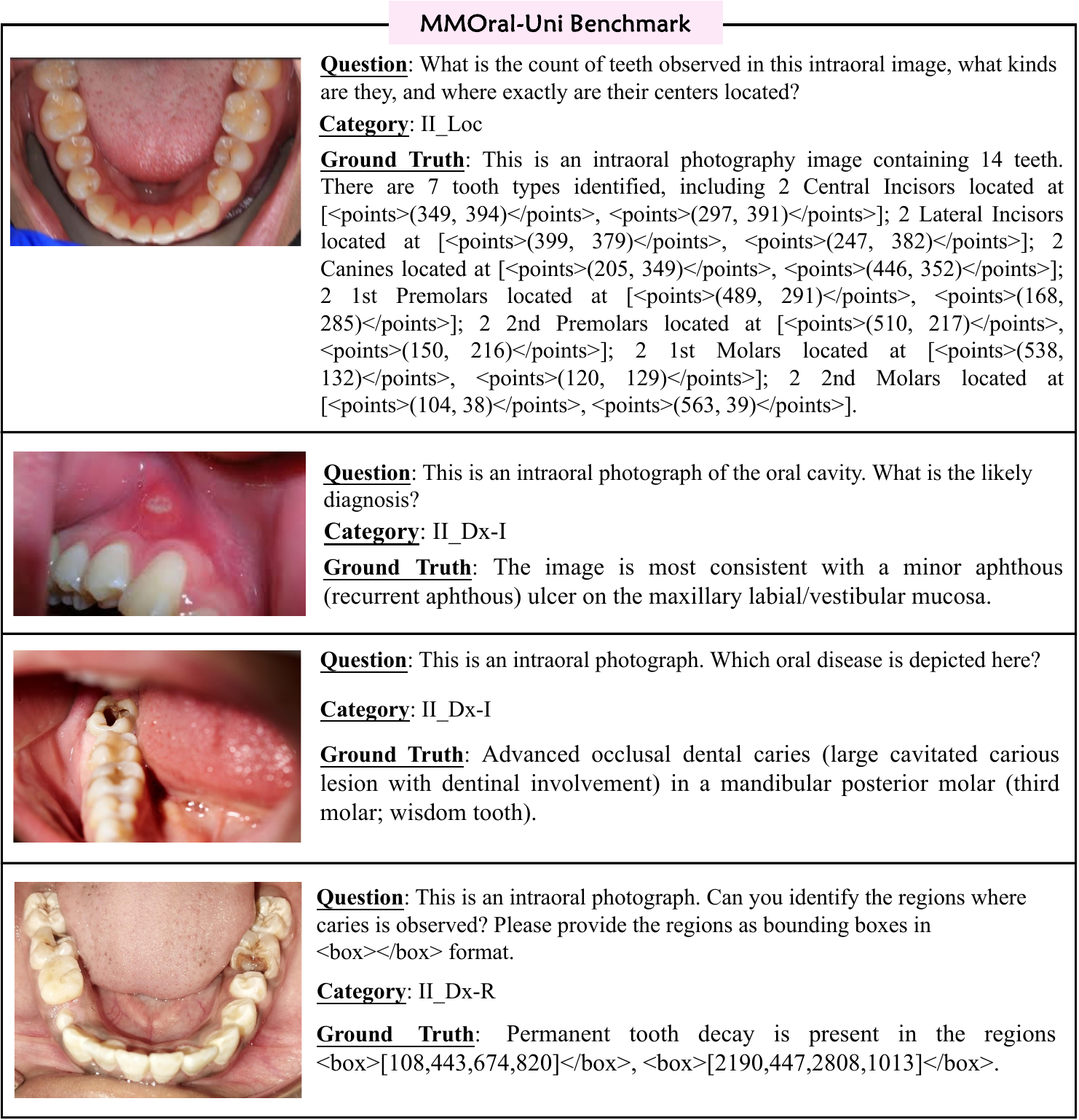}
  \caption{Some examples in our MMOral-Uni benchmark. Each example contains the image, question, category, and corresponding ground truth validated by experienced dentists.}
  \label{fig:suppl_benchmark_case_1}
\end{figure*}

\begin{figure*}[!ht]
  \centering
  \includegraphics[width=\textwidth]{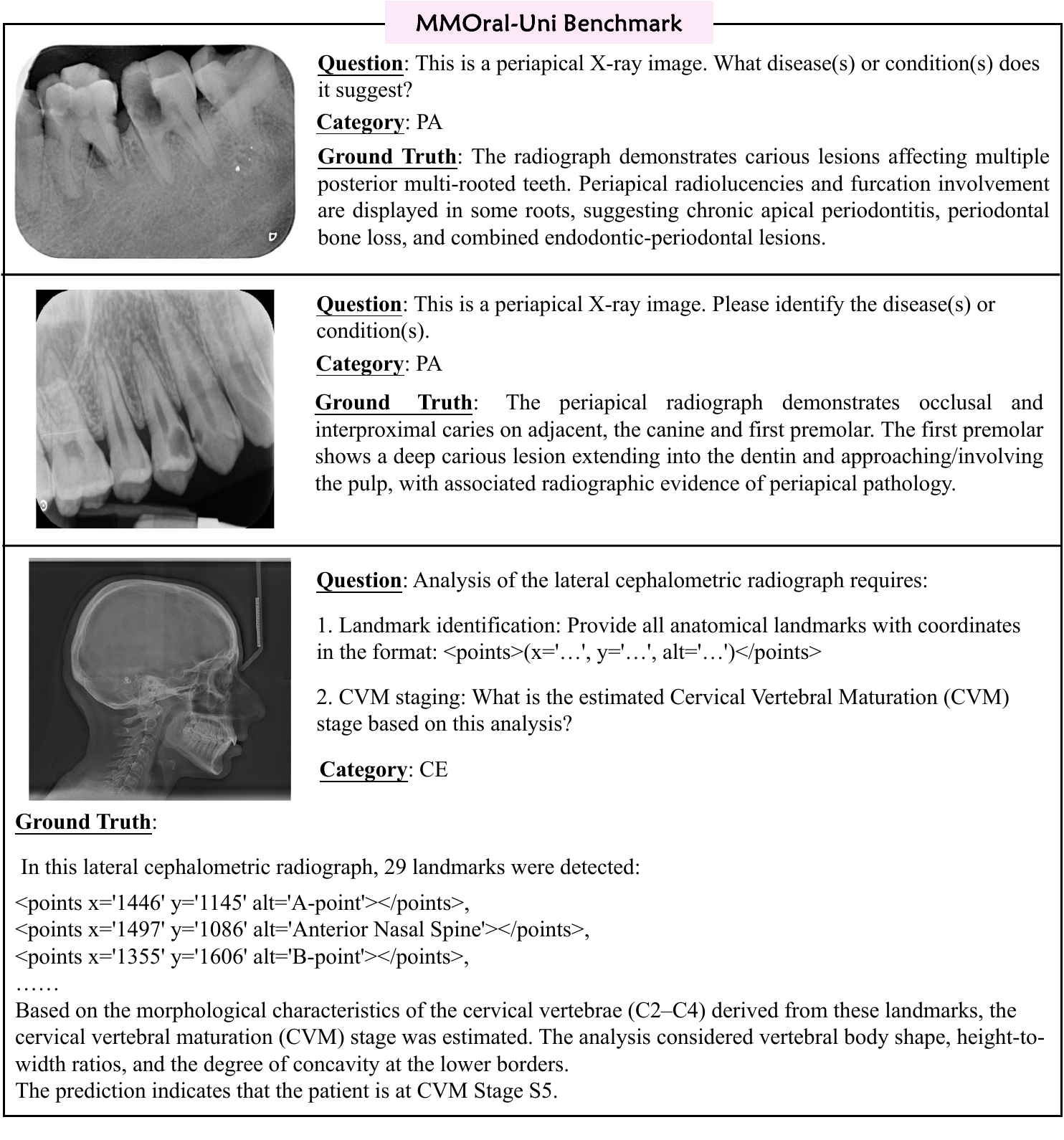}
  \caption{Some examples in our MMOral-Uni benchmark. Each example contains the image, question, category, and corresponding ground truth validated by experienced dentists.}
  \label{fig:suppl_benchmark_case_2}
\end{figure*}

\begin{figure*}[!ht]
  \centering
  \includegraphics[width=\textwidth]{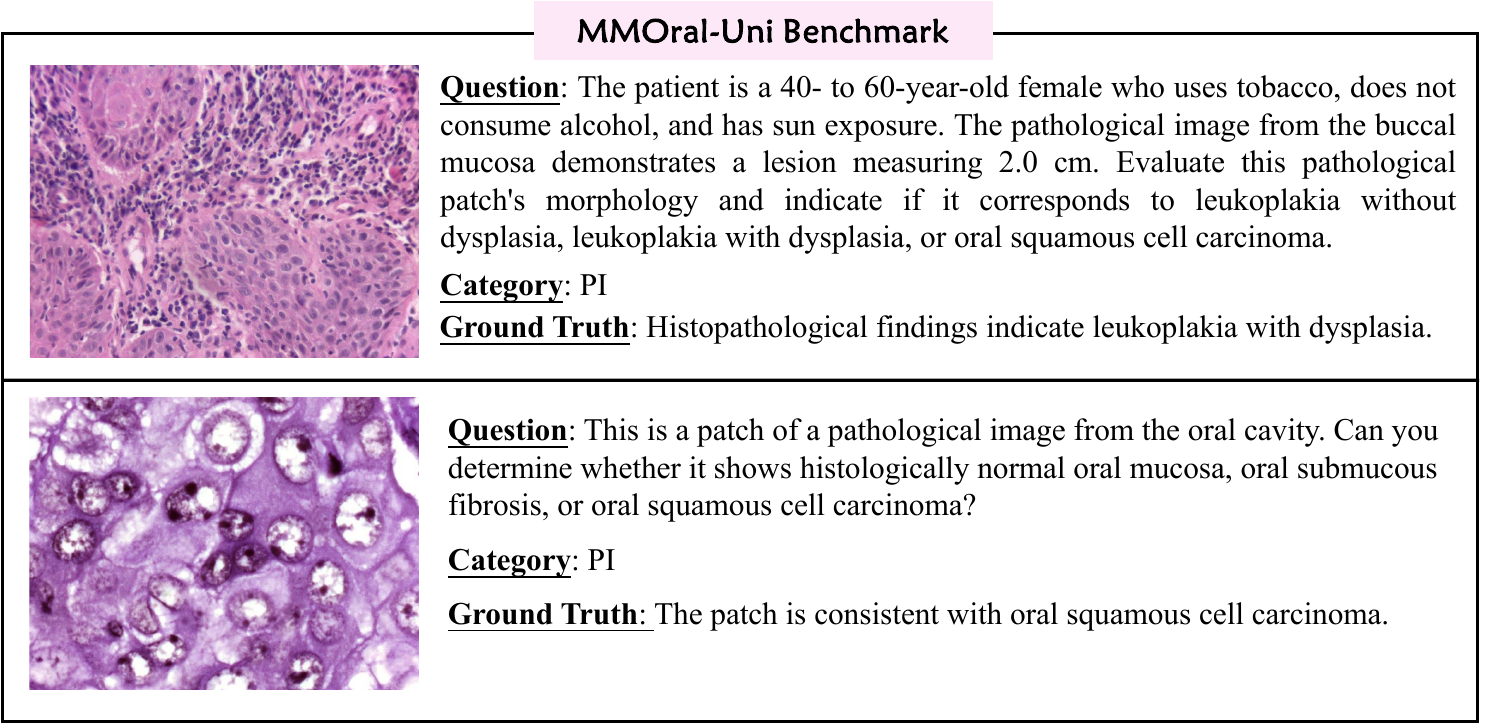}
  \caption{Some examples in our MMOral-Uni benchmark. Each example contains the image, question, category, and corresponding ground truth validated by experienced dentists.}
  \label{fig:suppl_benchmark_case_3}
\end{figure*}

\begin{figure*}[!ht]
  \centering
  \includegraphics[width=\textwidth]{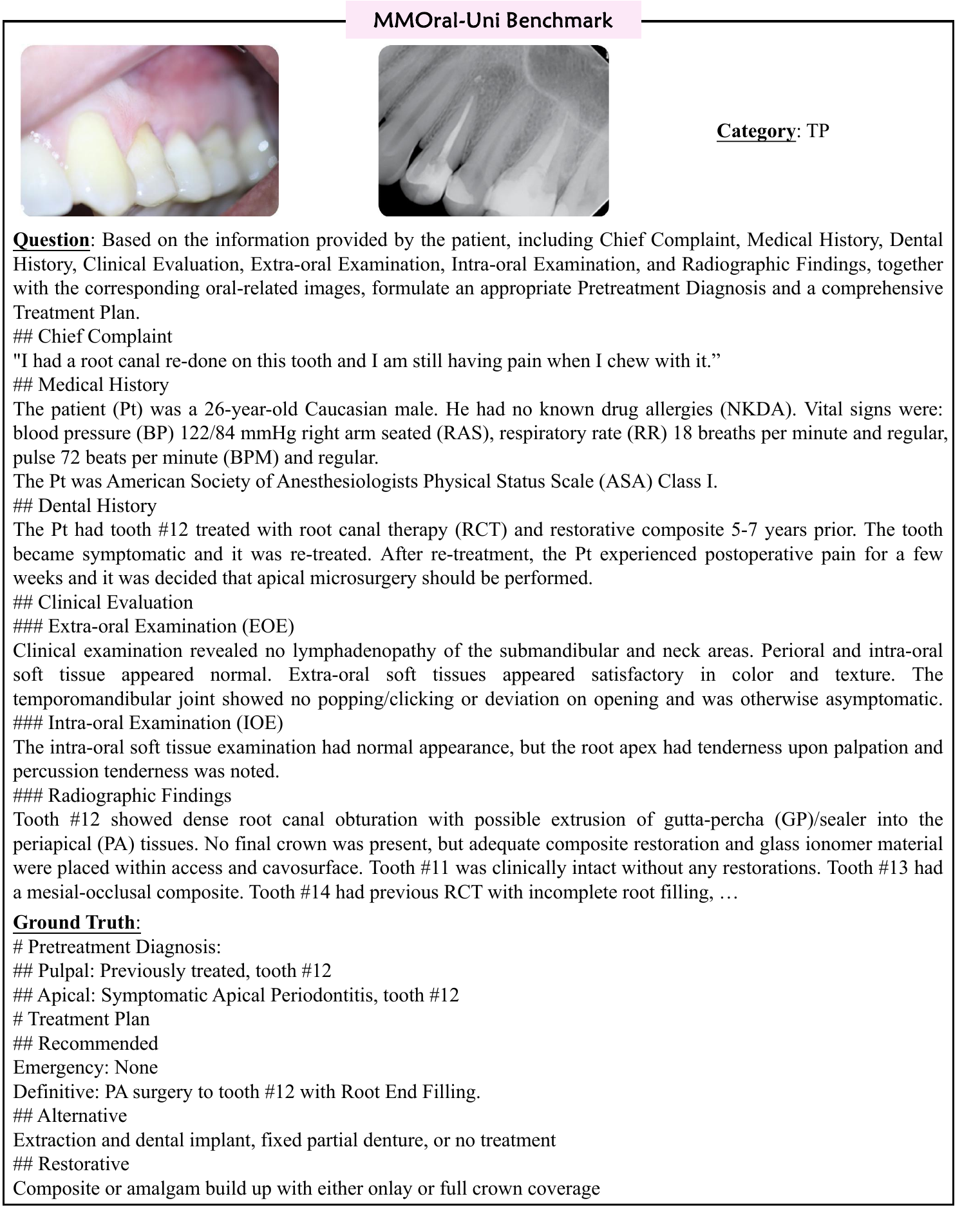}
  \caption{Some examples in our MMOral-Uni benchmark. Each example contains the image, question, category, and corresponding ground truth validated by experienced dentists.}
  \label{fig:suppl_benchmark_case_4}
\end{figure*}

\begin{figure*}[!ht]
  \centering
  \includegraphics[width=\textwidth]{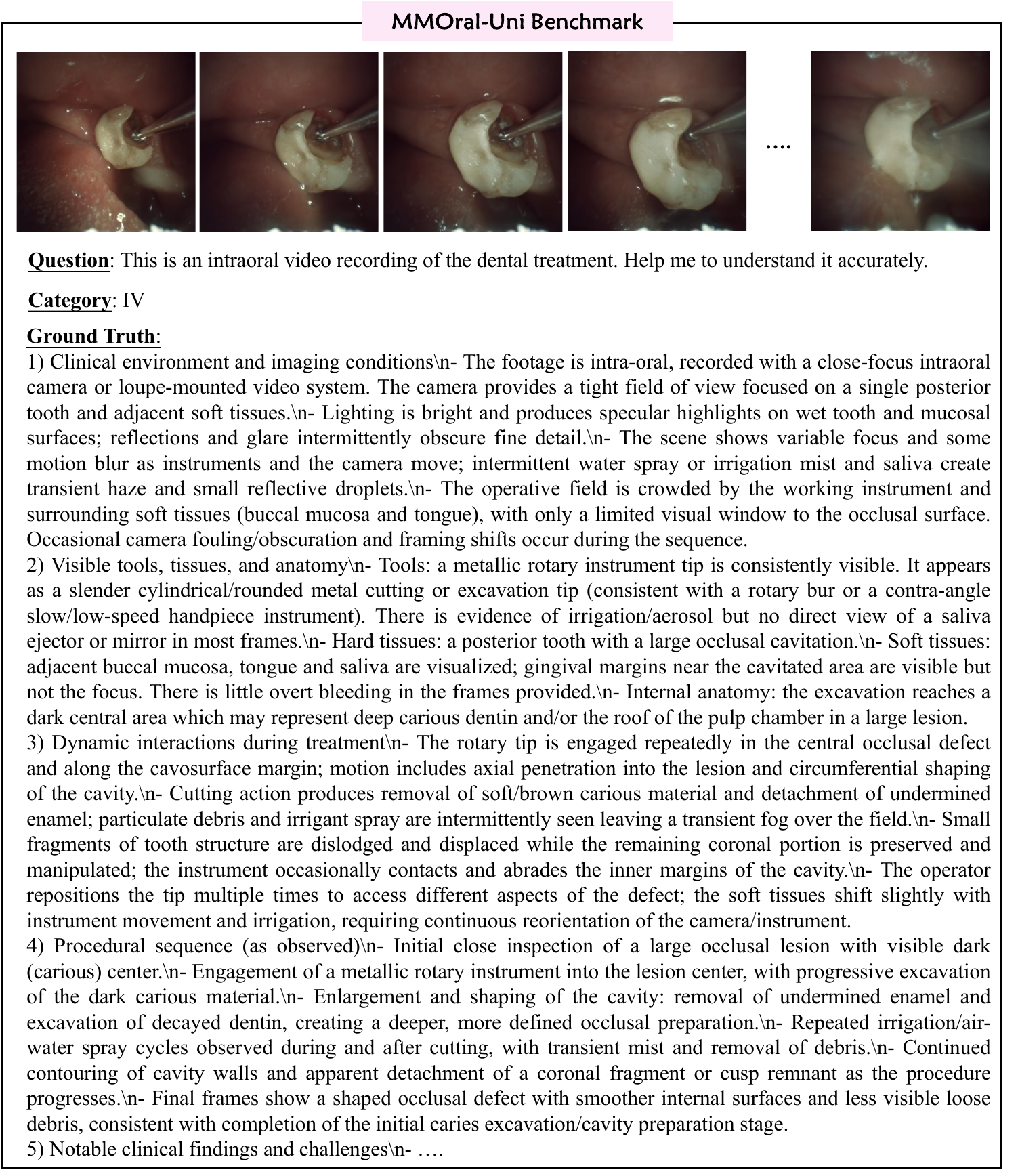}
  \caption{Some examples in our MMOral-Uni benchmark. Each example contains the image, question, category, and corresponding ground truth validated by experienced dentists.}
  \label{fig:suppl_benchmark_case_5}
\end{figure*}

Our MMOral-Uni benchmark is the first unified benchmark for dental multimodal imaging analysis, spanning five modalities and five tasks. It includes 2,809 open-ended QA pairs to simulate realistic user interactions. The modalities cover intraoral photographs, periapical and cephalometric radiographs, pathological images, intraoral videos, and interleaved image–text inputs. The tasks include abnormality diagnosis, cervical vertebral maturation (CVM) stage prediction, treatment planning, tooth localization and counting, and dental treatment video comprehension. The specific quantities for each task and modality are presented in Table~\ref{tab:benchmark_composition}. Regarding the abnormality diagnosis, the MMOral-Uni benchmark includes 40 categories of abnormalities, which are summarized in Table~\ref{tab:list_abnormality}. We also provide some examples in the MMOral-Uni benchmark for an intuitive understanding of this benchmark, which can be found in Figure~\ref{fig:suppl_benchmark_case_1} - Figure~\ref{fig:suppl_benchmark_case_5}.

\section{LLM as the judges for MMOral-Omni: A Feasibility Analysis}
\noindent
\textbf{Effectiveness.}
To verify the effectiveness of LLM-based evaluation for the MMOral-Omni benchmark, we invite two professional dentists to objectively score the outputs of different LVLMs. We calculate the absolute difference between the evaluators' scores and the human-annotated scores. Specifically, the few-shot prompts designed for LLM-based evaluation are presented to the dentists to guide the evaluation criteria. The two dentists then independently scored the predictions of GPT-5 and Lingshu-7B on 300 cases from the MMOral-Omni benchmark based on these criteria. The absolute differences between human scores and evaluators' scores are shown in Table~\ref{tab:Reliability}, represented as $\Delta$.

Overall, the absolute differences of the ``Overall'' metric given by dentists fluctuate by approximately 2 points in comparison to the LLM-based evaluations for the predictions of both LVLMs (GPT-5 and Lingshu-7B). This observation indicates that human scoring preferences generally align with the trends observed in LLM-based evaluations. However, it also suggests that there are subjective differences in the dentists' interpretations of the evaluation criteria outlined in the few-shot prompts. For each subcategory, Dentist A shows smaller differences in scores compared to the LLM-based evaluation for questions in the ``II-Loc'', ``II-Dx-I'', ``II-Dx-R'', ``PA'', ``CE'', and ``PI'', whereas the differences are larger for the ``TP'' and ``IV'' categories. Although Dentist B exhibits slightly larger differences with LLM-based scoring across all subcategories, their ``overall'' score difference is only 2.43 points. This indicates that LLM-based scoring aligns well with human preferences in reflecting the overall performance of LVLMs on the MMOral-Omni benchmark. At the same time, we speculate that the score fluctuations in each subcategory are strongly associated with the subjective perceptions of human evaluators.

\begin{figure*}[!ht]
  \centering
  \includegraphics[width=\textwidth]{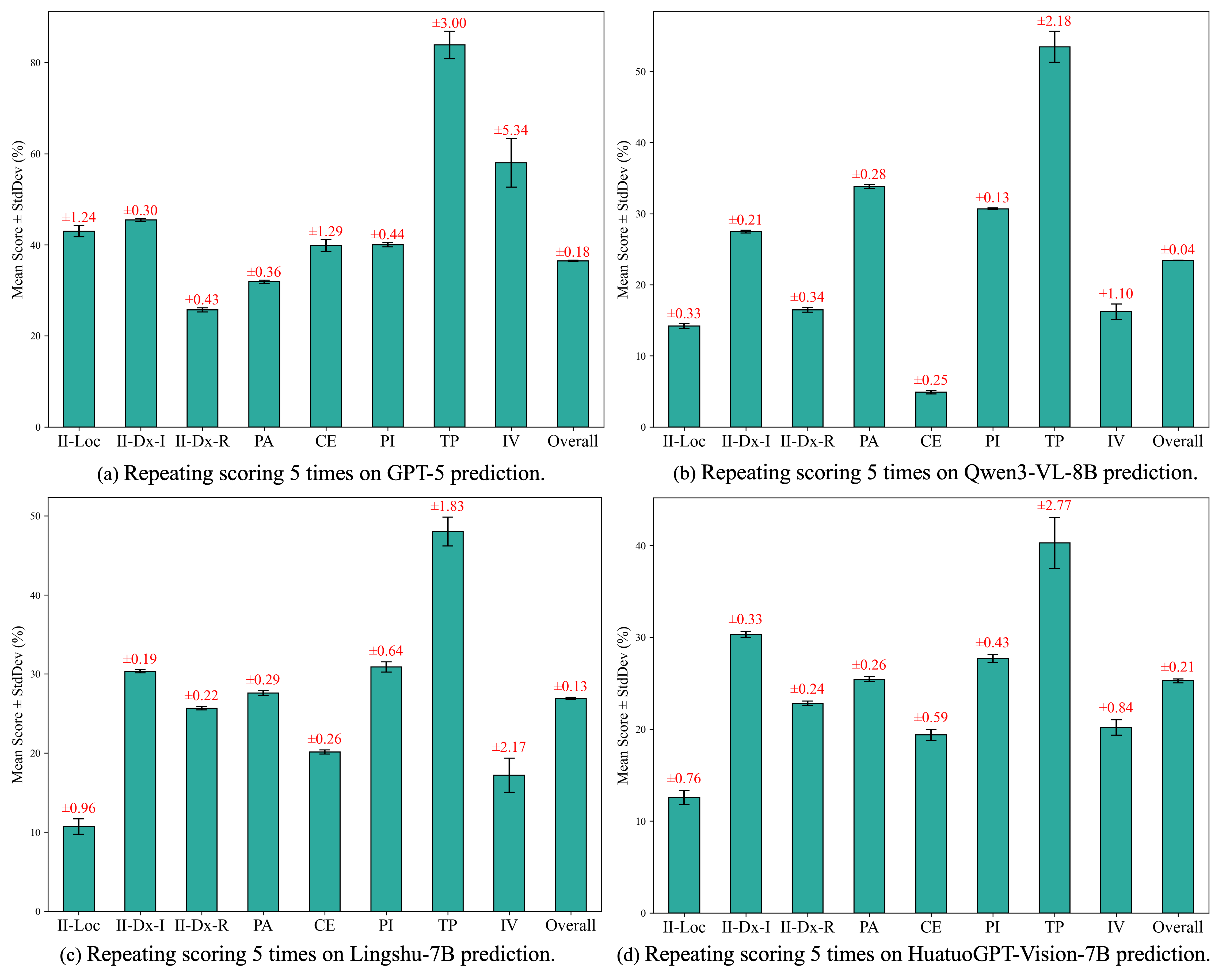}
  \vspace{-0.6cm}
  \caption{The means and standard deviations of each category on 5 repeated evaluations across four LVLMs' predictions.}
  \vspace{-0.45cm}
  \label{fig:suppl_llm-as-judge-stability}
\end{figure*}

\noindent
\textbf{Stability.}
Since using LLMs as judges inevitably introduces randomness, even with the temperature hyperparameter set to 0, we conduct multiple repeated experiments to verify the stability of LLMs as judges. Specifically, we evaluate the prediction results of GPT-5~\cite{gpt5}, Qwen3-VL-8B~\cite{yang2025qwen3}, Lingshu-7B~\cite{xu2025lingshu}, and HuatuoGPT-Vision-7B~\cite{chen2024huatuogpt} on the MMOral-Omni benchmark using GPT-5-mini~\cite{gpt5} with the same prompt five times. The obtained mean, standard deviation, and coefficient of variation (CV) of the metric ``overall'' are shown in Table~\ref{tab:Stability}. For proprietary models, medical-specific models, and general-purpose LVLMs, the standard deviation of the metric "overall" is no more than 0.212 when evaluated 5 times using GPT-5-mini with our designed few-shot prompt. Specifically, for the prediction results of GPT-5, the standard deviation of the scores is 0.179, while for Qwen3-VL-8B, the standard deviation is as low as 0.039. Meanwhile, CV (Coefficient of Variation), as a standardized measure of dispersion of a probability distribution, can be used to assess the stability of scores across multiple experiments. The CV values for the prediction results of these four models, after being scored 5 times, are all around 0.5\%, which demonstrates the evaluation stability of using LLMs as evaluators. The detailed results across each specific category are demonstrated in Figure~\ref{fig:suppl_llm-as-judge-stability}.

\section{Case Study and Clinical Validity of OralGPT-Omni}
\begin{figure*}[!ht]
  \centering
  \includegraphics[width=\textwidth]{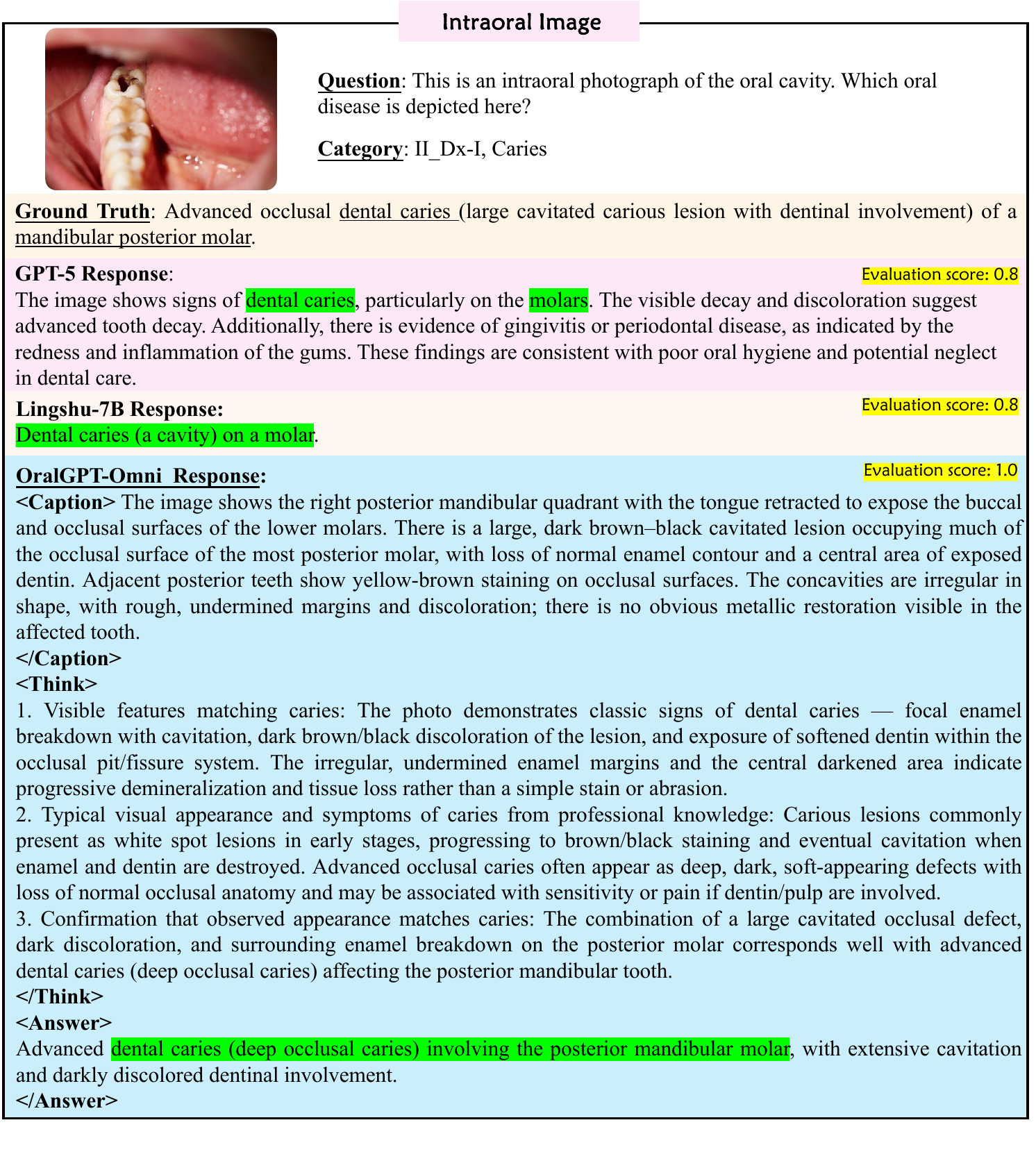}
  \caption{Case study on the intraoral image modality, with correct responses highlighted in \colorbox{green}{\textcolor{black}{green}} and obvious incorrect responses highlighted in \textcolor{red}{red}.}
  \label{fig:suppl_comparative_case_1}
\end{figure*}

\begin{figure*}[!ht]
  \centering
  \includegraphics[width=\textwidth]{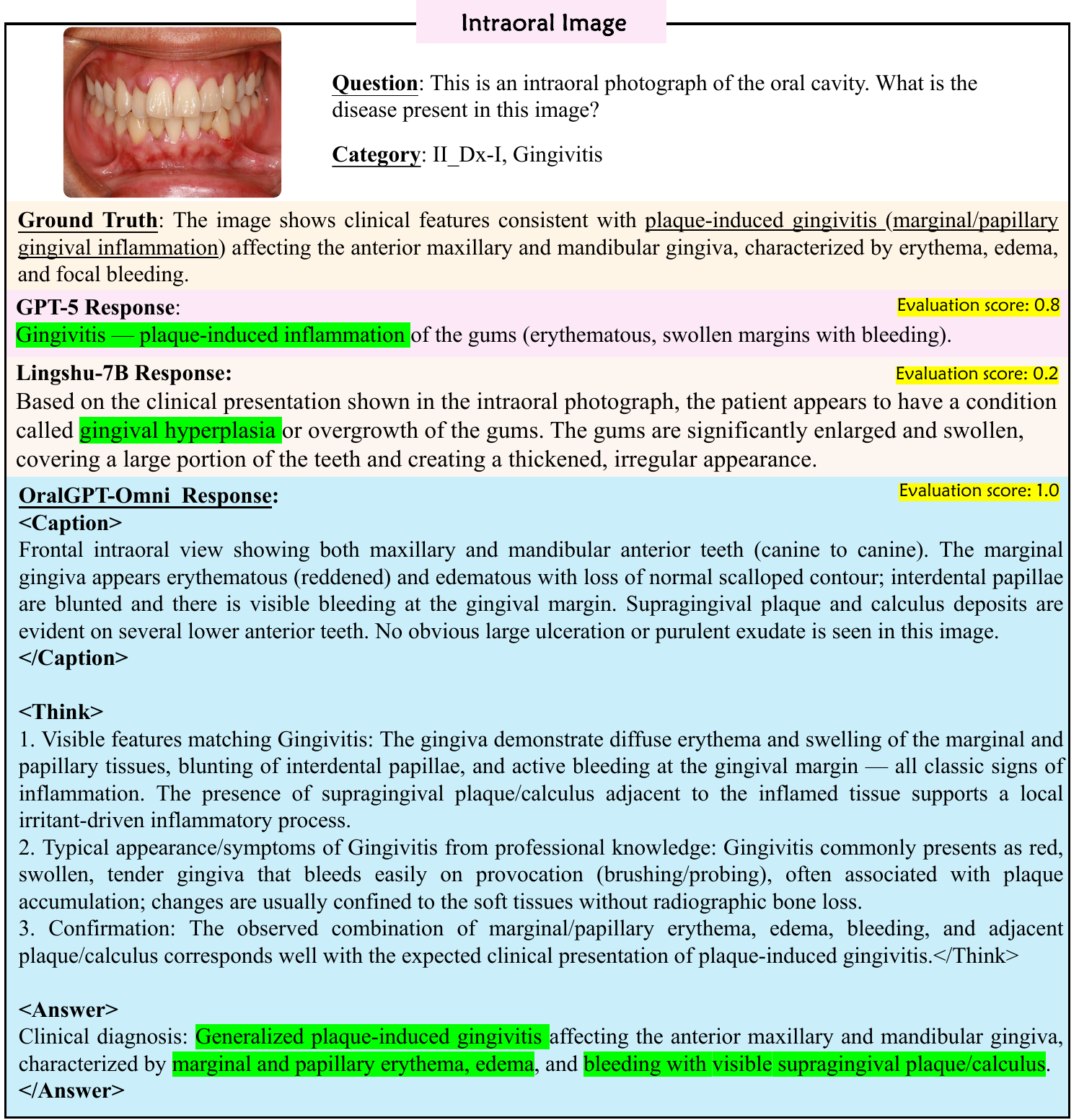}
  \caption{Case study on the intraoral image modality, with correct responses highlighted in \colorbox{green}{\textcolor{black}{green}} and obvious incorrect responses highlighted in \textcolor{red}{red}.}
  \label{fig:suppl_comparative_case_2}
\end{figure*}

\begin{figure*}[!ht]
  \centering
  \includegraphics[width=\textwidth]{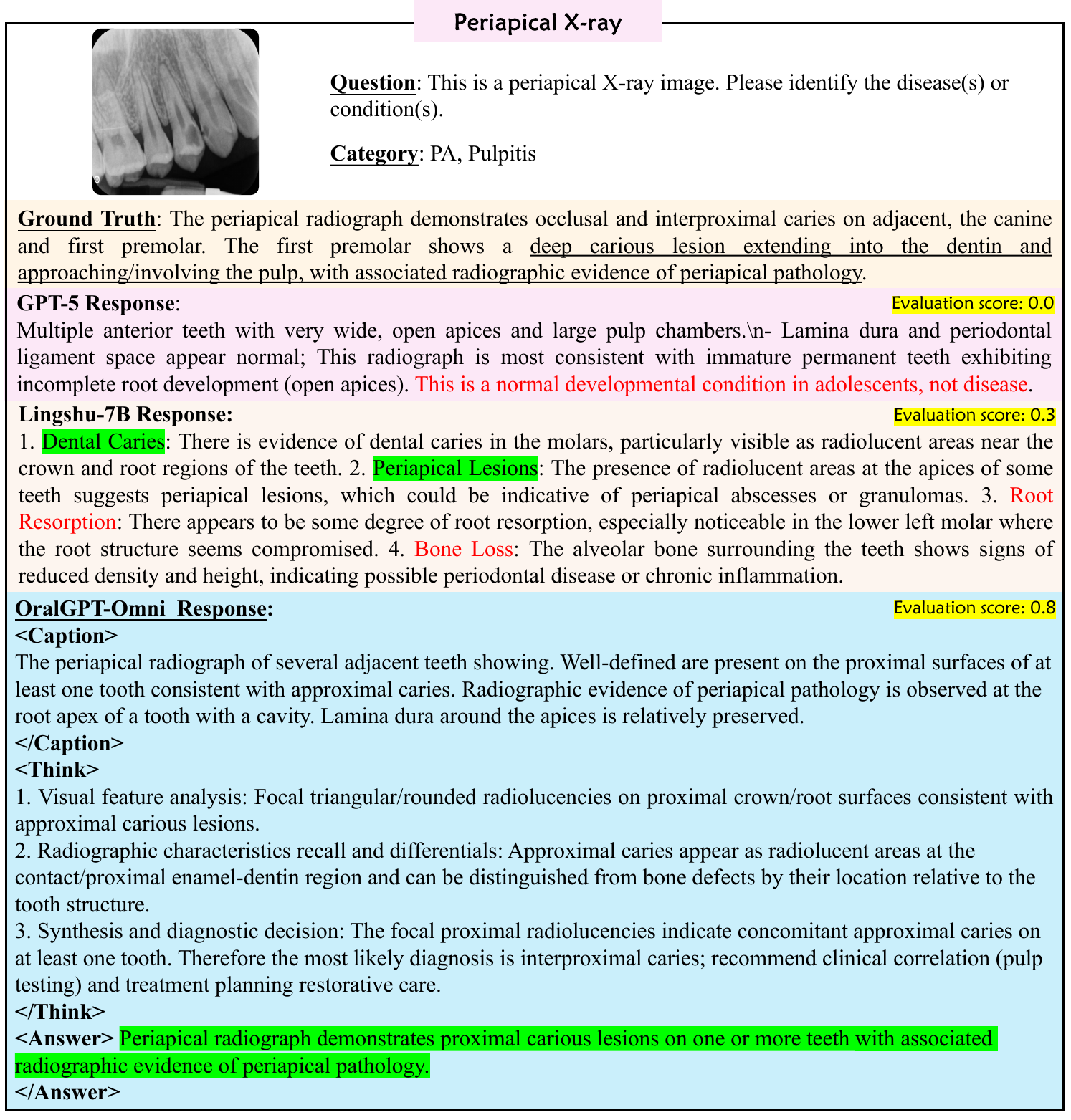}
  \caption{Case study on the periapical X-ray modality, with correct responses highlighted in \colorbox{green}{\textcolor{black}{green}} and obvious incorrect responses highlighted in \textcolor{red}{red}.}
  \label{fig:suppl_comparative_case_3}
\end{figure*}

\begin{figure*}[!ht]
  \centering
  \includegraphics[width=\textwidth]{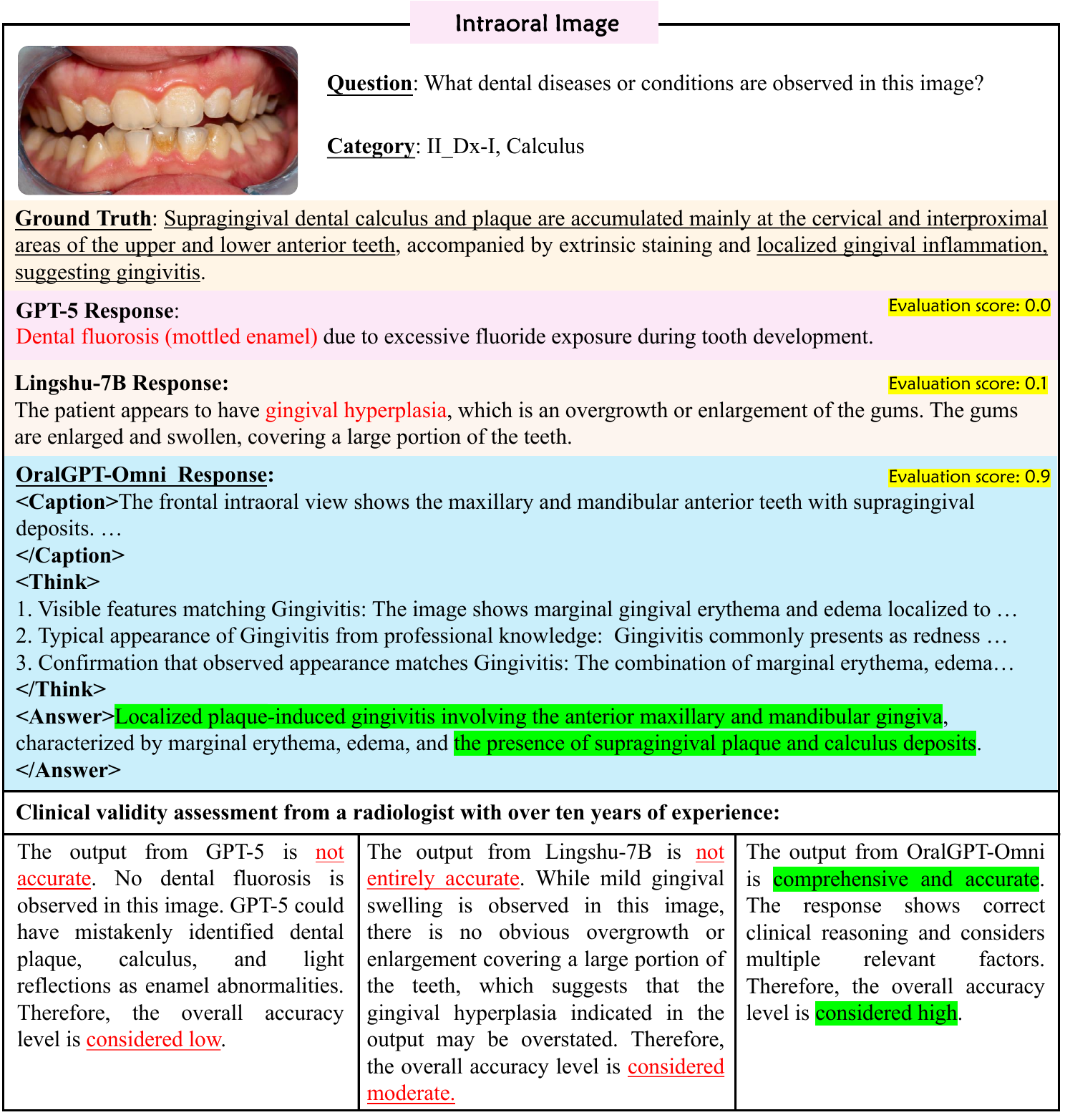}
  \caption{Case study and clinical validity on the intraoral image modality, with correct responses highlighted in \colorbox{green}{\textcolor{black}{green}} and obvious incorrect responses highlighted in \textcolor{red}{red}.}
  \label{fig:suppl_clinical_validity_1}
\end{figure*}

\begin{figure*}[!ht]
  \centering
  \includegraphics[width=\textwidth]{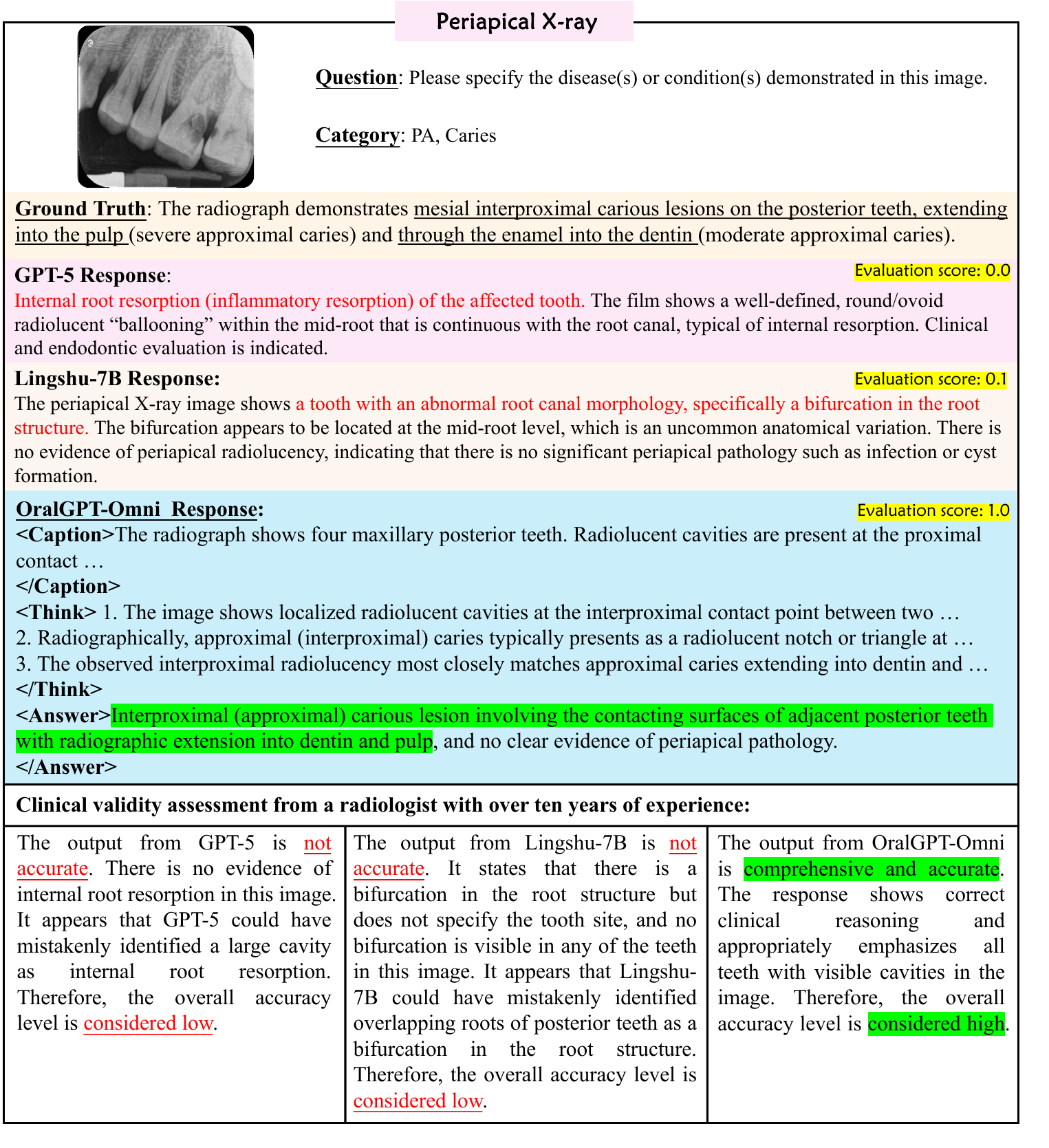}
  \caption{Case study and clinical validity on the periapical X-ray modality, with correct responses highlighted in \colorbox{green}{\textcolor{black}{green}} and obvious incorrect responses highlighted in \textcolor{red}{red}.}
  \label{fig:suppl_clinical_validity_2}
\end{figure*}

\begin{figure*}[!ht]
  \centering
  \includegraphics[width=\textwidth]{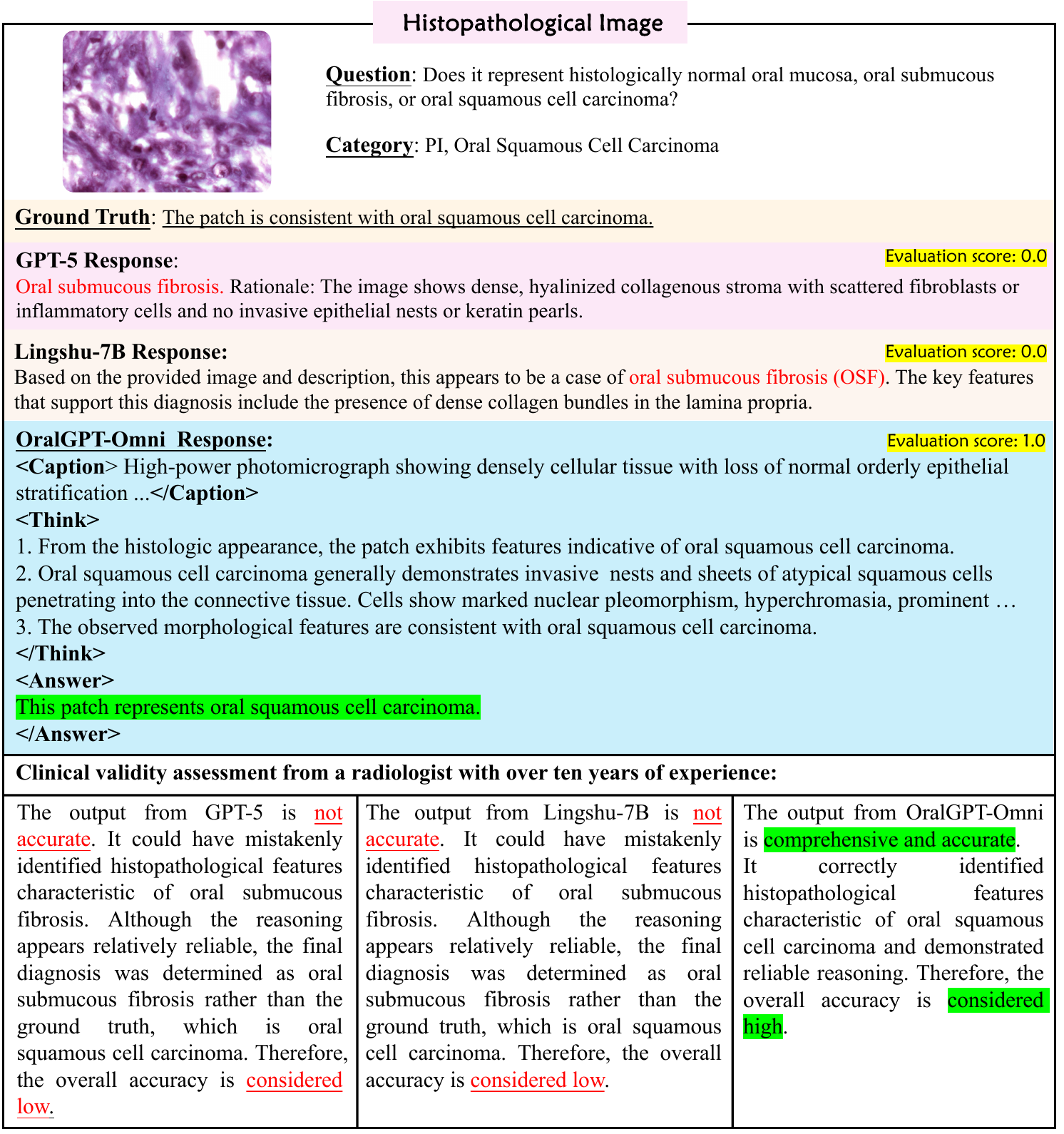}
  \caption{Case study and clinical validity on the histopathological image modality, with correct responses highlighted in \colorbox{green}{\textcolor{black}{green}} and obvious incorrect responses highlighted in \textcolor{red}{red}.}
  \label{fig:suppl_clinical_validity_3}
\end{figure*}

\begin{figure*}[!ht]
  \centering
  \includegraphics[width=\textwidth]{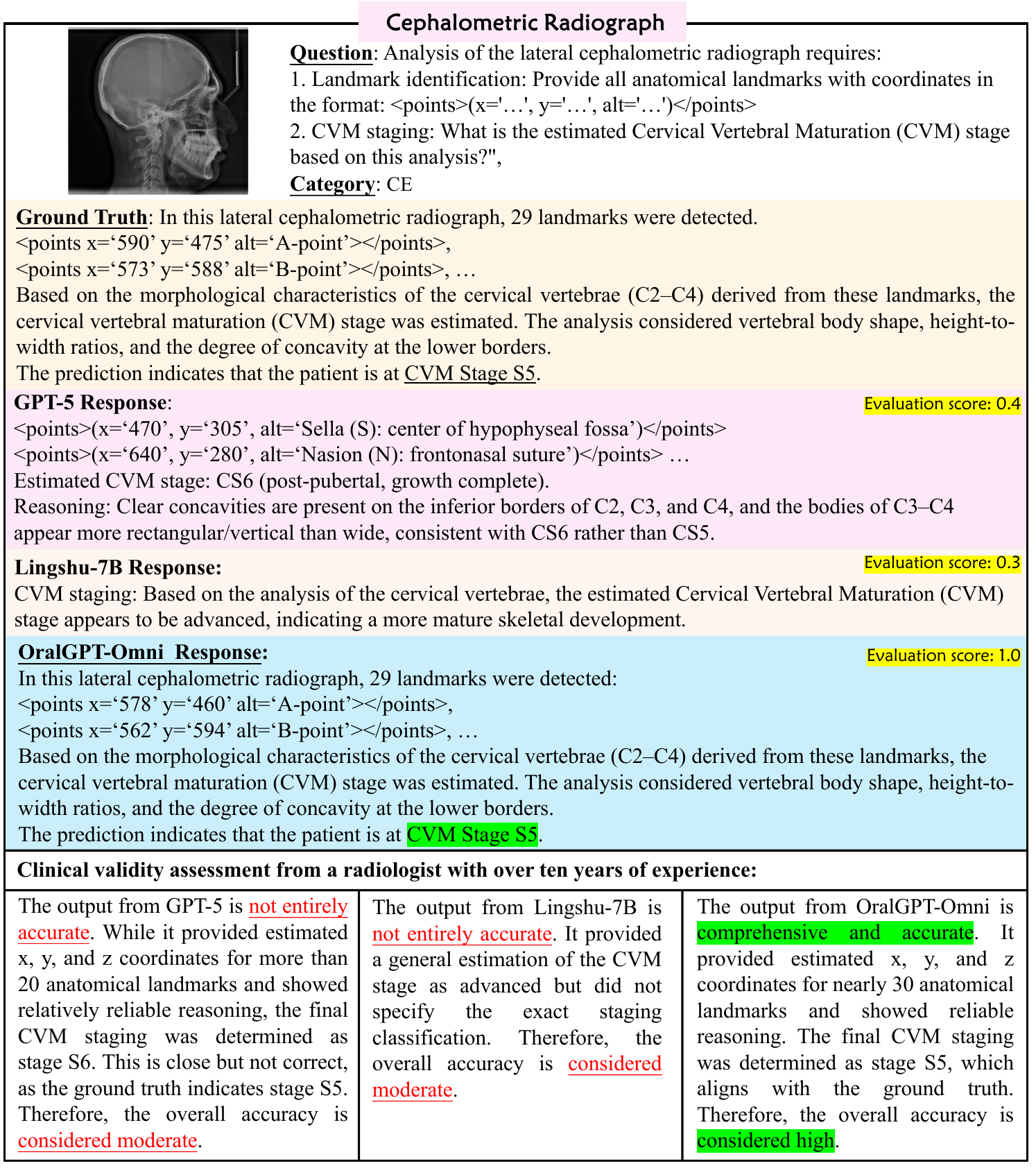}
  \caption{Case study and clinical validity on the cephalometric radiograph modality, with correct responses highlighted in \colorbox{green}{\textcolor{black}{green}} and obvious incorrect responses highlighted in \textcolor{red}{red}.}
  \label{fig:suppl_clinical_validity_4}
\end{figure*}

We provide more case studies to demonstrate the superiority of our OralGPT-Omni, as illustrated in Figure~\ref{fig:suppl_comparative_case_1}, Figure~\ref{fig:suppl_comparative_case_2}, and Figure~\ref{fig:suppl_comparative_case_3}. From a clinical perspective, the ability to accurately identify common dental diseases and conditions is a fundamental prerequisite for diagnosis and treatment planning.  To explore the clinical validity of the OralGPT-Omni, we invite a professoriate faculty member in oral-maxillofacial radiology with over ten years of experience in diagnostic imaging studies to conduct in-depth clinical assessments on the outputs of the top-performing models (GPT-5, Lingshu-7B, and our OralGPT-Omni). The clinical validation of the model outputs on four representative modalities is demonstrated in Figure~\ref{fig:suppl_clinical_validity_1}, Figure~\ref{fig:suppl_clinical_validity_2}, Figure~\ref{fig:suppl_clinical_validity_3}, and Figure~\ref{fig:suppl_clinical_validity_4}. The cases and clinical feedback effectively demonstrate the superior performance and potential clinical value of our OralGPT-Omni. 
\begin{table}[!t]
\centering
\caption{Stability verification of using LLMs as judges: Standard deviation and coefficient of variation (CV) are reported across four LVLMs from five repeated evaluations.}
\label{tab:Stability}
\setlength{\tabcolsep}{5pt}
\small
\begin{tabular}{p{3.5cm}|>{\centering}p{1.1cm}|>{\centering}p{1.1cm}|>{\centering\arraybackslash}p{1.1cm}}
\toprule
\textbf{Model} & \textbf{Mean} & \textbf{StdDev} & \textbf{CV \%}  \\ 
\midrule
GPT-5~\cite{gpt5} & 36.446  & 0.179 &  0.490  \\ 
\midrule
Qwen3-VL-8B~\cite{yang2025qwen3} & 23.436  & 0.039  & 0.164 \\  \midrule
Lingshu-7B~\cite{xu2025lingshu} & 26.910  & 0.127 & 0.472 \\  \midrule
HuatuoGPT-Vision-7B~\cite{chen2024huatuogpt} & 25.258 & 0.212 & 0.838 \\ 
\bottomrule
\end{tabular}
\end{table}

\onecolumn

\begin{longtable}[h]{@{}>{\centering\arraybackslash}m{1.5cm} p{2.5cm} >{\raggedright\arraybackslash}p{5.8cm} p{7cm}@{}}
        \caption{The category, visual appearance, and relevant dental knowledge utilized in prompts for the TRACE-CoT data construction pipeline across multiple datasets.} 
        \label{tab:dental_knowledge} \\
        
        \toprule
        Dataset & Category & Appearence &  Dental Knowledge \\
        \midrule
        \endfirsthead
    
        \toprule
        Dataset & Category & Appearence &  Dental Knowledge \\
        \midrule
        \endhead
    
        \bottomrule
        \endfoot
        \bottomrule
        \endlastfoot
        
        \multirow{9}{*}{\makecell[l]{AlphaDent \\ \href{https://huggingface.co/datasets/ZFTurbo/AlphaDent/tree/main}{\includegraphics[width=0.015\textwidth]{figs/link.png} URL}}}
        & Abrasion & Tooth with mechanical wear of hard tissues & Abrasion typically appears as flat, smooth, and well-defined areas of hard tissue loss, often wedge-shaped at the cervical margins, with a shiny polished surface. \\ 
        & Filling & Filling & A filling is visually recognized as a distinct restorative material embedded in the tooth, usually with a regular geometric outline and a surface texture or color different from natural enamel or dentin. \\ 
        & Crown & Installed crown & A crown presents as a full-coverage artificial cap over the tooth, with smooth and uniform surfaces, clear margins near the gingiva, and contours that may look slightly different from natural tooth anatomy depending on the material. \\ 
        & Caries 1 class & Caries in fissures and blind pits of teeth (occlusal surfaces of molars and premolars, buccal surfaces of molars, lingual surfaces of upper incisors) & Caries are seen as irregular areas of demineralization or cavitation, with rough surfaces and discoloration ranging from chalky white to brown or black, often with undermined enamel edges. \\
        & Caries 2 class & Caries of the contact surfaces of molars and premolars. &  \\ 
        & Caries 3 class & Caries of the contact surfaces of incisors and canines without damage to the cutting edge. &  \\ 
        & Caries 4 class & Caries of the contact surfaces of incisors and canines with damage to the cutting edge. &  \\ 
        & Caries 5 class & Cervical caries of the vestibular and lingual surfaces. &  \\ 
        & Caries 6 class & Caries of the cutting edges of the front teeth and the cusps of the chewing teeth. &  \\ 
        \midrule
        \multirow{2}{*}{\makecell[l]{ \\ \\ \\ \\ \\ Dental Caries \\ Detection \\ \href{https://zenodo.org/records/14827784}{\includegraphics[width=0.015\textwidth]{figs/link.png} URL}}
        } & Primary tooth decay & Occlusal surface of a primary molar shows a small, dark brown spot in the central pit, with surrounding enamel appearing slightly chalky. Minor enamel surface irregularities are visible along the grooves. & The affected primary tooth shows a localized brownish or dark spot on the occlusal or smooth surfaces, with enamel appearing softened or chalky. The child may exhibit mild sensitivity to sweet or cold foods, occasional discomfort when chewing, and sometimes visible plaque accumulation near the lesion. Gingival tissue surrounding the tooth may appear slightly inflamed. \\ 
         & Permanent tooth decay & Occlusal surface of a permanent molar shows a large, dark brown cavitated lesion extending into the central pit and fissures. Enamel margins appear undermined, and surface texture is roughened. & Permanent teeth with decay display larger cavitated lesions, often dark brown or black in color, with enamel margins appearing undermined. Patients may complain of prolonged sensitivity to cold, sweet, or acidic foods, intermittent pain while chewing, and visible plaque or calculus accumulation at the affected site. \\
         \midrule
         
        \multirow{1}{*}{\makecell[l]{ \\ \\  FDTooth \\ \href{https://zenodo.org/records/14827784}{\includegraphics[width=0.015\textwidth]{figs/link.png} URL}}
        } & Anterior teeth with fenestration and dehiscence & Localized gingival recession with partially exposed root surfaces, a thin labial mucosa allowing clear visualization of root contours, cementoenamel junction, and subtle convexities, and small fenestration windows or linear dehiscence defects present in the alveolar bone beneath. & Anterior teeth with fenestration and dehiscence exhibit localized gingival recession with exposed root surfaces. The thin labial mucosa allows visualization of the underlying root contour, and alveolar bone may be partially or completely deficient in the affected regions. Clinically, patients are often asymptomatic or may report mild sensitivity at the exposed sites. \\ 
        \midrule
        \multirow{1}{*}{\makecell[l]{ \\ \\  GINGIVITIS \\ \href{https://zenodo.org/records/14827784}{\includegraphics[width=0.015\textwidth]{figs/link.png} URL}}
        } & Gingivitis & The region shows marked redness, edema, and/or marginal gingival hypertrophy of the unit or spontaneous bleeding, papillary, congestion, or ulceration. & Severe inflammation of gingivitis presents with intense redness and deep erythema of the gingival margins and interdental papillae, pronounced swelling, a shiny and moist surface, spontaneous bleeding in some areas, hypertrophy and deformation of papillae, and occasional ulcerations. \\ 
        \multirow{1}{*}{\makecell[l]{ \\ \\  Dental Plaque \\ \href{https://zenodo.org/records/14827784}{\includegraphics[width=0.015\textwidth]{figs/link.png} URL}}
        } & Dental plaque & A soft, adherent film on the tooth surface, typically pale yellow to light brown, often along the gingival margin or in pits and fissures. & Dental plaque appears as a soft, adherent film on tooth surfaces, typically pale yellow to light brown, often along the gingival margin or in pits and fissures. Clinically, patients are usually asymptomatic or may report mild sensitivity to sweet foods, with slight gingival redness and minimal inflammation. \\ 
        \midrule
        \multirow{9}{*}{\makecell[l]{ \\ \\  OMNI\_COCO \\ \href{https://zenodo.org/records/14827784}{\includegraphics[width=0.015\textwidth]{figs/link.png} URL}}
        } & Tooth torsion & A condition where a tooth is rotated or twisted around its longitudinal axis, resulting in an abnormal orientation within the dental arch. & Tooth torsion results in a tooth noticeably rotated around its longitudinal axis, causing misalignment within the dental arch; this rotation creates uneven spacing with adjacent teeth, sometimes overlapping or crowding neighboring teeth, which can lead to difficulty in chewing, localized food impaction, accumulation of plaque, and mild inflammation of the surrounding gingival tissue. \\ 

         & Deep overjet & A dental malocclusion where the upper front teeth (maxillary incisors) horizontally overlap the lower front teeth (mandibular incisors) to an excessive degree. & Deep overjet occurs when the upper front teeth are prominently protruded over the lower front teeth, producing a pronounced horizontal overlap; this may impair lip closure, cause difficulty in biting or chewing, increase the risk of trauma to the anterior teeth or lips, and create aesthetic concerns due to facial profile changes. \\
         & Invisible orthodontic attachment & These attachments, part of clear aligner or aesthetic orthodontic treatments, are tooth-colored or transparent devices bonded to teeth to aid in controlled tooth movement without the visibility of traditional braces. & Invisible orthodontic attachment involves small, tooth-colored or transparent attachments bonded to the teeth to guide clear aligner therapy, blending almost seamlessly with the natural tooth surface; patients may experience minor gum irritation, localized plaque accumulation, or slight sensitivity during initial treatment stages. \\
         & Tooth emergence & The process by which a tooth moves through the alveolar bone and soft tissue to appear in the oral cavity. This is a natural phase of dental development. & Tooth emergence occurs when a tooth is erupting through the gingival tissue, causing visible swelling and redness around the emerging crown; this process may result in mild tenderness, discomfort during chewing, and localized inflammation, sometimes accompanied by increased salivation or irritability. \\
         & Cast fixed orthodontic appliance & Custom-made orthodontic devices, such as fixed space maintainers or lingual braces, which are fabricated using dental impressions (casts) of the patient’s teeth and are permanently attached to the teeth to correct or maintain dental alignment. & Cast fixed orthodontic appliance consists of custom-made fixed devices bonded to teeth, precisely matching the dental arch; these appliances may be metallic or tooth-colored, producing initial soreness, localized gum irritation, and requiring meticulous oral hygiene to prevent plaque accumulation. \\
         & Tooth misalignment & Improper positioning of teeth within the dental arch, leading to malocclusion. This includes various forms of malpositions such as crowding, spacing, and deviations from the ideal arch form. & Tooth misalignment occurs when teeth are irregularly positioned in the dental arch, showing crowding, spacing, or tilting; this misalignment can interfere with normal bite function, create difficulty in chewing, increase plaque accumulation, and elevate the risk of gingival inflammation or caries. \\
         & Mandibular retrusion & A skeletal condition characterized by the posterior positioning of the mandible relative to the maxilla, often contributing to the class II malocclusion. & Mandibular retrusion presents with the lower jaw positioned posteriorly relative to the upper jaw, visible in profile view; this skeletal discrepancy can cause Class II malocclusion, impaired occlusion, difficulty chewing, and altered facial aesthetics with a receding chin. \\
         & Orthodontic brace & A comprehensive orthodontic appliance consisting of brackets, wires, and sometimes bands, which is fixed to the teeth. This device applies continuous pressure to move teeth into the desired position. & Orthodontic brace consists of fixed braces with metal or ceramic brackets bonded to teeth with connecting archwires, sometimes including colored elastics; these appliances apply continuous force to correct alignment, causing mild soreness, gum irritation, and requiring careful oral hygiene. \\
         & Fixed orthodontic device & Any orthodontic apparatus that is bonded to the teeth and remains in place throughout the treatment period. These devices, which include traditional braces and certain types of retainers, are used to correct dental and skeletal discrepancies by exerting controlled forces over time. & Fixed orthodontic device remains permanently bonded to teeth throughout treatment, including traditional braces or fixed retainers; they may cause mild soreness, gum irritation, difficulty in maintaining oral hygiene, and localized plaque accumulation. \\        
\end{longtable}

\twocolumn

\begin{table*}[!t]
\centering
\caption{Categories of analysis for abnormal conditions and diseases.}
\label{tab:list_abnormality}
\begin{tabular}{|c|c|c|c|}
\hline
\multicolumn{4}{|c|}{\textbf{Categories of analysis for abnormal conditions and diseases}} \\
\hline
Caries & Gingivitis & Ulcer & Tooth discoloration \\
\hline
Defective dentition & Cancer & Normality & Orthodontics \\
\hline
Abrasion & Filling & Crown & Caries \\
\hline
Fenestration and dehiscence & Tooth torsion & Deep overjet & Invisible orthodontic attachment \\
\hline
Tooth emergence & Case fixed orthodontic appliances & Tooth misalignment & Mandibular retrusion \\
\hline
Orthodontic brace & Fixed orthodontic device & Dental plaque & Impacted tooth \\
\hline
Pulpitis & Periodontitis & Apical periodontitis & Bone loss \\
\hline
Root canal treatment & Restoration & Mixed dentition & Leukoplakia without dysplasia \\
\hline
Leukoplakia with dysplasia & Oral squamous cell carcinoma & Oral submucous fibrosis & healthy epithelial nucleus \\
\hline
Abnormal epithelial nucleus & Blood cell nucleus & Reactive cell nucleus & Dividing nucleus \\
\hline
\end{tabular}
\end{table*}

\begin{table*}[!h]
\centering
\caption{Average absolute differences ($\Delta$) between the evaluation scores of the LLM-based evaluator and the dentist-annotated scores on the MMOral-Omni benchmark.}
\label{tab:Reliability}
\setlength{\tabcolsep}{5pt}
\small
\begin{tabular}{p{2cm}|>{\centering}p{1.7cm}|>{\centering}p{0.9cm}|>{\centering}p{1.1cm}|>{\centering}p{1.1cm}|>{\centering}p{0.8cm}|>{\centering}p{0.7cm}|>{\centering}p{1cm}|>{\centering}p{1cm}|>{\centering}p{1cm}|>{\centering\arraybackslash}p{1cm}}
\toprule
\textbf{Model} & \textbf{Evaluators} & \textbf{II-Loc} & \textbf{II-Dx-I} & \textbf{II-Dx-R} & \textbf{PA} & \textbf{CE} & \textbf{PI} & \textbf{TP} & \textbf{IV} & \textbf{Overall}\\ 
\midrule
& GPT-5-mini & 44.60 & 45.24 & 25.16 & 31.43 & 41.27 & 40.52 & 80.67 & 56.00 & 36.42 \\
GPT-5~\cite{gpt5} & Dentist A & 45.82 & 43.24 & 27.58 & 35.30 & 39.50 & 38.17 & 60.56 & 49.80  & 34.94 \\ 
& $\Delta (\downarrow)$ & +1.22 & -2.00 & +2.42 & +3.87 & -1.77 & -2.35 & -20.11 & -6.20  & -1.48 \\ 

\midrule
& GPT-5-mini & 12.00 & 30.58 & 25.77 & 27.48 & 20.50 & 30.94 & 48.00  & 20.00 & 27.08 \\
Lingshu-7B~\cite{xu2025lingshu} & Dentist B & 18.20 & 32.96 & 28.95 & 23.34& 26.83 & 28.26 & 42.24 & 14.50  & 29.51 \\ 
& $\Delta (\downarrow)$ & +6.20 & +2.38 & +3.18 & -4.14 & +6.33 & -2.68 & -5.76 & -5.50  & +2.43 \\ 
\bottomrule
\end{tabular}
\end{table*}


\begin{table*}[ht]
    \centering
    \caption{Detailed compositions of tasks and modalities in the MMOral-Omni benchmark.}
    \label{tab:benchmark_composition}
    \begin{tabular}{@{}llc@{}}
        \toprule
        \textbf{Task} & \textbf{Modality} & \textbf{Number} \\ \midrule
        & Intraoral Image & 1462 \\
        Diagnosis of Abnormality & Periapical X-ray & 539 \\
        & Pathological Image & 383 \\ \midrule
        CVM Stage Prediction & Cephalometric X-ray & 300 \\ \midrule
        Tooth Location \& Counting & Intraoral Image & 100 \\ \midrule
        Treatment Planning & Interleaved Image-Text Data & 15 \\ \midrule
        Dental Treatment Video Comprehension & Intraoral Video & 10 \\ 
        \bottomrule
    \end{tabular}
\end{table*}

\end{document}